\pdfoutput=1 
\documentclass{article} 

\usepackage[dblblindworkshop,preprint]{neurips_2026} 
\workshoptitle{VLM4RWD}

\usepackage[utf8]{inputenc}
\usepackage[T1]{fontenc}
\usepackage[hidelinks]{hyperref}
\usepackage{url}
\usepackage{booktabs}
\usepackage{amsfonts}
\usepackage{amsmath}
\usepackage{graphicx}
\usepackage{nicefrac}
\usepackage{microtype}
\usepackage[table]{xcolor}
\usepackage{float}
\usepackage{placeins}
\usepackage{makecell}
\usepackage{multirow}
\usepackage{listings}
\usepackage[most]{tcolorbox}
\tcbuselibrary{listings,breakable,skins}
\usepackage{tikz}
\usetikzlibrary{arrows.meta,positioning,fit,backgrounds,calc,shapes.geometric}

\newtcblisting{promptbox}{%
  breakable,
  enhanced,
  colback=black!3,
  colframe=black!55,
  boxrule=0.5pt,
  arc=1.5pt,
  left=5pt,
  right=5pt,
  top=4pt,
  bottom=4pt,
  boxsep=1pt,
  before skip=0.6em,
  after skip=0.8em,
  listing only,
  listing options={%
    basicstyle=\ttfamily\scriptsize,
    breaklines=true,
    breakatwhitespace=false,
    columns=fullflexible,
    keepspaces=true,
    showstringspaces=false,
    aboveskip=0pt,
    belowskip=0pt,
  },
}

\definecolor{FigInk}{HTML}{17212B}
\definecolor{FigMuted}{HTML}{65717E}
\definecolor{FigRule}{HTML}{CBD3DA}
\definecolor{FigBad}{HTML}{C94C4C}
\definecolor{FigBadWash}{HTML}{FFF3F1}
\definecolor{FigGood}{HTML}{187A6B}
\definecolor{FigGoodWash}{HTML}{EFF8F5}
\definecolor{FigBlue}{HTML}{3568A8}
\definecolor{FigBlueWash}{HTML}{EFF4FA}
\definecolor{FigPurple}{HTML}{6E56CF}
\definecolor{FigPurpleWash}{HTML}{F5F2FF}
\definecolor{GainLight}{HTML}{E5F4F1}
\definecolor{GainMid}{HTML}{BFE4DD}
\definecolor{GainStrong}{HTML}{86CEC2}
\definecolor{GainText}{HTML}{005F56}

\newcommand{\gainband}[2]{%
  \begingroup
  \setlength{\fboxsep}{1.5pt}%
  \colorbox{#1}{%
    \makebox[2.55cm][c]{%
      {\color{GainText}\bfseries\boldmath\ensuremath{#2}}%
    }%
  }%
  \endgroup
}
\newcommand{\gainlo}[1]{\gainband{GainLight}{#1}}
\newcommand{\gainmid}[1]{\gainband{GainMid}{#1}}
\newcommand{\gainhi}[1]{\gainband{GainStrong}{#1}}

\definecolor{ReductionLight}{HTML}{E5F4F1}
\definecolor{ReductionMid}{HTML}{BFE4DD}
\definecolor{ReductionStrong}{HTML}{86CEC2}
\definecolor{ReductionText}{HTML}{005F56}
\definecolor{WorseBackground}{HTML}{F6D7D2}
\definecolor{WorseText}{HTML}{8B2F26}
\newcommand{\changeband}[3]{%
  \begingroup
  \setlength{\fboxsep}{1.5pt}%
  \colorbox{#1}{%
    \makebox[2.50cm][c]{%
      {\color{#2}\bfseries\boldmath\ensuremath{#3}}%
    }%
  }%
  \endgroup
}
\newcommand{\reductionlo}[1]{\changeband{ReductionLight}{ReductionText}{\downarrow\,#1\ \mathrm{pp}}}
\newcommand{\reductionmid}[1]{\changeband{ReductionMid}{ReductionText}{\downarrow\,#1\ \mathrm{pp}}}
\newcommand{\reductionhi}[1]{\changeband{ReductionStrong}{ReductionText}{\downarrow\,#1\ \mathrm{pp}}}
\newcommand{\worsening}[1]{\changeband{WorseBackground}{WorseText}{\uparrow\,#1\ \mathrm{pp}}}

\definecolor{DropLight}{HTML}{F6D7D2}
\definecolor{DropMid}{HTML}{E8A79C}
\definecolor{DropStrong}{HTML}{D47A6E}
\newcommand{\ncband}[2]{%
  \begingroup
  \setlength{\fboxsep}{1.5pt}%
  \colorbox{#1}{%
    \makebox[2.85cm][c]{%
      {\color{WorseText}\bfseries\boldmath\ensuremath{#2}}%
    }%
  }%
  \endgroup
}
\newcommand{\ncflat}[1]{%
  \begingroup
  \setlength{\fboxsep}{1.5pt}%
  \colorbox{GainLight}{%
    \makebox[2.85cm][c]{%
      {\color{GainText}\bfseries\boldmath\ensuremath{#1}}%
    }%
  }%
  \endgroup
}
\newcommand{\ncdroplo}[1]{\ncband{DropLight}{#1}}
\newcommand{\ncdropmid}[1]{\ncband{DropMid}{#1}}
\newcommand{\ncdrophi}[1]{\ncband{DropStrong}{#1}}

\graphicspath{{./figures/}}

\newcommand{\oursagentname}{Graph-Grounded Harness (Ours)}
\newcommand{\oursagent}{\makecell{Graph-Grounded\\Harness (Ours)}}
\newcommand{\ourspipeline}{P\&ID Graph Recovery Stack (Ours)}
\newcommand{\topopid}{TopoPID-VQA}

\title{Grounded and Faithful P\&ID Reasoning:\\
Constraining Vision-Language Models with Recovered Evidence Graphs}

\author{%
\makebox[0pt][c]{%
\begin{minipage}{\textwidth}
\centering
Prathamesh Gadekar\textsuperscript{*}
\hspace{0.65em}
Sagar Sakhinana Srinivas
\hspace{0.65em}
Venkataramana Runkana
\\[0.45em]
{\normalfont\mdseries Tata Research Development and Design Center, Pune, India 411057}
\end{minipage}%
}%
}

\begin{document}

\maketitle

\begin{abstract}
Piping and Instrumentation Diagrams (P\&IDs) are the authoritative maps of
process plants: isolation, maintenance, and HAZOP decisions depend on
\emph{what connects to what}. Vision-language models describe these sheets
fluently, yet they often invent or miss process connections---and an invented
or missed link can reverse an isolation or reachability call, so a plant
decision cannot trust a fluent answer that was never checked against the
linework. We instead recover an explicit graph of the
drawing---its symbols, the process connections between them, and the tags
that name them---and then require the model to answer only by querying that
graph through seven read-only operators, so a topology claim is returned only
when it cites the query results that support it. On \topopid{}, a new suite
of 3000 topology questions over these sheets, \oursagentname{} raises
exact-match accuracy from \textbf{36.7--41.3\%} under image-only prompting to
\textbf{74.3--76.0\%} for Qwen3-VL-4B, Qwen3-VL-8B, and Gemma-4-E4B. It does
so on an imperfect substrate: on Digitize-PID dataset the
recovered graph scores $F_1$ \textbf{0.742} on exact process connections, and
\textbf{0.801} once symbols and tags are pooled in. The residual errors track
that gap---grounding pays off where the recovered graph is right, and
perception error still breaks topology questions where it is not.
\end{abstract}

\section{Introduction}
\label{sec:introduction}

Piping and Instrumentation Diagrams (P\&IDs) are the working drawings of
process plants: they record which equipment, valves, and instruments share
process lines and control loops
\citep{paliwal2021digitize,rahul2019pid}.
Oil and gas, chemicals, power generation, pharmaceuticals, water treatment,
and aerospace propulsion or ground-test facilities all keep P\&IDs as the
authoritative map for isolation, maintenance, HAZOP, and as-built checks.
Those sheets are how a plant is licensed, modified, and shut down safely;
an invented or missed connection is an operational error, not a caption
glitch. Engineers query them for isolation planning, tag tracing, routing
checks, and other calls that depend on \emph{what connects to what}, not on
a free-form description of the drawing.

Most plant archives still hold P\&IDs as rasters. Reading topology from those
pixels is slow for people and unreliable for vision-language models (VLMs):
a model may name symbols fluently yet invent a path between mark~A and
mark~B with no check against the linework. On a P\&ID that invented edge can
reverse an isolation or reachability call.

A useful P\&ID assistant therefore needs two things. First, an explicit process
graph recovered from the sheet so connectivity is a stored fact rather than a
guessed relation. Second, a way to answer English questions that forces every
topology claim to cite those graph facts. We ask how much accuracy that
grounding buys on topology questions over these sheets, and how much remains
limited by the quality of the recovered graph itself.

Real sheets make the difficulty concrete: hundreds of near-identical symbols,
long pipe runs, fly-over crossings, and tags that look alike. Pixel-only reading
is a poor fit; process topology is what the decisions depend on.

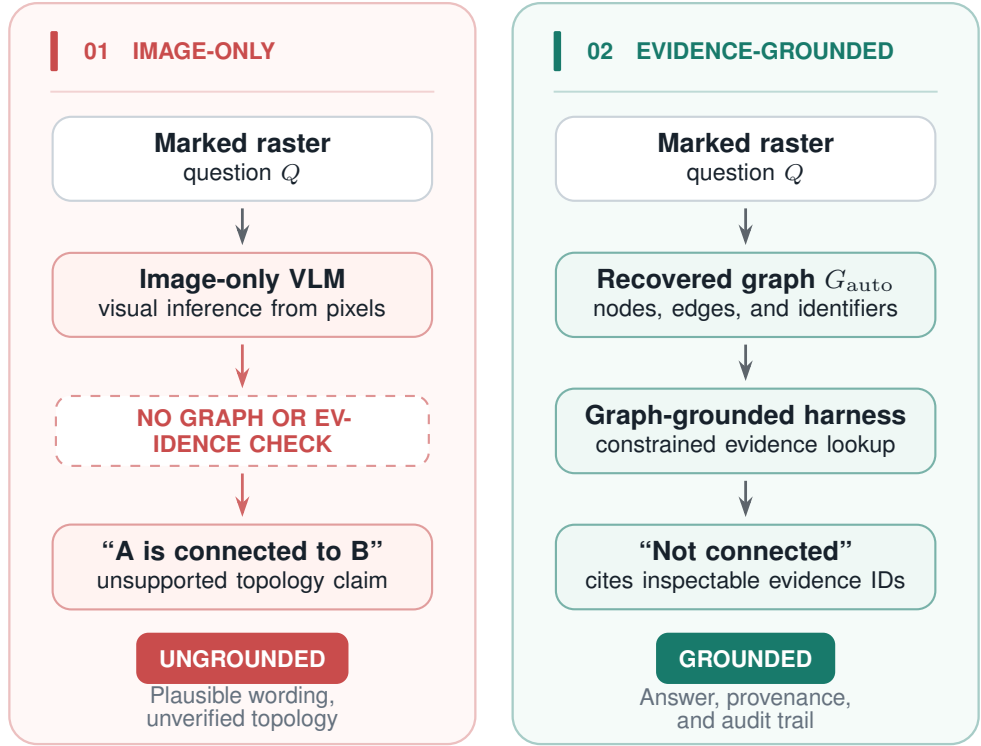
\begin{figure}[t]
\centering
\resizebox{0.92\linewidth}{!}{%
\begin{tikzpicture}[
  font=\sffamily\footnotesize,
  stage/.style={
    draw=FigRule,
    fill=white,
    text=FigInk,
    rounded corners=2mm,
    line width=0.65pt,
    align=center,
    text width=3.72cm,
    minimum height=8.5mm,
    inner sep=5pt,
    font=\sffamily\footnotesize
  },
  failure/.style={
    stage,
    draw=FigBad!55,
    fill=FigBadWash
  },
  evidence/.style={
    stage,
    draw=FigGood!58,
    fill=FigGoodWash
  },
  missing/.style={
    draw=FigBad!60,
    dashed,
    fill=white,
    text=FigBad,
    rounded corners=1.5mm,
    line width=0.65pt,
    align=center,
    text width=3.72cm,
    minimum height=7.5mm,
    inner sep=4pt,
    font=\sffamily\bfseries\scriptsize
  },
  flow/.style={
    -{Stealth[length=2.2mm,width=1.5mm]},
    draw=FigInk!70,
    line width=0.85pt,
    shorten <=2pt,
    shorten >=2pt
  },
  badflow/.style={
    flow,
    draw=FigBad!85
  },
  badge/.style={
    rounded corners=1.3mm,
    minimum height=5.5mm,
    inner xsep=7pt,
    inner ysep=2pt,
    font=\sffamily\bfseries\scriptsize,
    text=white
  }
]

\path[
  draw=FigBad!35,
  fill=FigBadWash!55,
  rounded corners=3mm,
  line width=0.7pt
]
  (-5.20,0.85) rectangle (-0.20,-7.10);

\path[
  draw=FigGood!38,
  fill=FigGoodWash!60,
  rounded corners=3mm,
  line width=0.7pt
]
  (0.20,0.85) rectangle (5.20,-7.10);

\fill[
  FigBad,
  rounded corners=0.5pt
]
  (-4.76,0.55) rectangle (-4.68,0.12);

\node[
  anchor=west,
  font=\sffamily\bfseries\scriptsize,
  text=FigBad
] at (-4.52,0.34)
  {01\quad IMAGE-ONLY};

\fill[
  FigGood,
  rounded corners=0.5pt
]
  (0.64,0.55) rectangle (0.72,0.12);

\node[
  anchor=west,
  font=\sffamily\bfseries\scriptsize,
  text=FigGood
] at (0.88,0.34)
  {02\quad EVIDENCE-GROUNDED};

\draw[
  FigBad!25,
  line width=0.6pt
]
  (-4.76,-0.10) -- (-0.64,-0.10);

\draw[
  FigGood!28,
  line width=0.6pt
]
  (0.64,-0.10) -- (4.76,-0.10);

\node[stage] (lin) at (-2.70,-0.82)
  {\textbf{Marked raster}\\[-1pt]
   {\scriptsize question $Q$}};

\node[failure] (lvlm) at (-2.70,-2.28)
  {\textbf{Image-only VLM}\\[-1pt]
   {\scriptsize visual inference from pixels}};

\node[missing] (lmissing) at (-2.70,-3.74)
  {NO GRAPH OR EVIDENCE CHECK};

\node[failure] (lout) at (-2.70,-5.20)
  {\textbf{``A is connected to B''}\\[-1pt]
   {\scriptsize unsupported topology claim}};

\draw[flow]
  (lin.south) -- (lvlm.north);

\draw[badflow]
  (lvlm.south) -- (lmissing.north);

\draw[badflow]
  (lmissing.south) -- (lout.north);

\node[stage] (rin) at (2.70,-0.82)
  {\textbf{Marked raster}\\[-1pt]
   {\scriptsize question $Q$}};

\node[evidence] (rgraph) at (2.70,-2.28)
  {\textbf{Recovered graph} $G_{\mathrm{auto}}$\\[-1pt]
   {\scriptsize nodes, edges, and identifiers}};

\node[evidence] (rharness) at (2.70,-3.74)
  {\textbf{Graph-grounded harness}\\[-1pt]
   {\scriptsize constrained evidence lookup}};

\node[evidence] (rout) at (2.70,-5.20)
  {\textbf{``Not connected''}\\[-1pt]
   {\scriptsize cites inspectable evidence IDs}};

\draw[flow]
  (rin.south) -- (rgraph.north);

\draw[flow]
  (rgraph.south) -- (rharness.north);

\draw[flow]
  (rharness.south) -- (rout.north);

\node[
  badge,
  fill=FigBad
] at (-2.70,-6.18)
  {UNGROUNDED};

\node[
  align=center,
  text=FigMuted,
  font=\sffamily\scriptsize
] at (-2.70,-6.72)
  {Plausible wording,\\[-1pt]
   unverified topology};

\node[
  badge,
  fill=FigGood
] at (2.70,-6.18)
  {GROUNDED};

\node[
  align=center,
  text=FigMuted,
  font=\sffamily\scriptsize
] at (2.70,-6.72)
  {Answer, provenance,\\[-1pt]
   and audit trail};

\end{tikzpicture}%
}

\caption{
\textbf{Grounding converts a plausible prediction into an auditable answer.}
An image-only VLM may invent a process connection without checking the
underlying linework (left). A graph-grounded harness conditions the answer on
recovered topology and returns inspectable evidence identifiers (right).
}
\label{fig:faithful}
\end{figure}

Our system has two stages. Stage~1 recovers a grounding substrate: finetuned
perception finds symbols, lines, and text; deterministic rules (intersection
noding, sheet-adaptive junction snap, first-symbol linker walk, tag association)
build the typed process graph $G_{\mathrm{auto}}$. An optional
\emph{signal-edge layer} can attach instrument links on dashed polylines; we
report its recovery separately, and \topopid{} questions use process edges
only. Stage~2 is \oursagentname{}: a VLM plans over seven fact-returning graph
operators, grounds marks and tags on the sheet, and returns a structured
answer that cites the supporting graph entities. Answers without cited
evidence are rejected. Figure~\ref{fig:faithful} contrasts the two regimes.

\paragraph{Contributions.}
\begin{itemize}
  \item \textbf{Grounding mechanism.} \oursagentname{} couples a recovered
    evidence graph with seven generic operators and an answer contract
    (tool call required; named entities must be queried; typed answers that cite
    supporting entities).
  \item \textbf{Topology benchmark.} \topopid{} is a 3000-question
    topology suite (easy/medium/hard) that scores answers under graph grounding on
    the recovered process graph $G_{\mathrm{auto}}$.
  \item \textbf{Deployment-facing measurement.} Across Qwen3-VL-4B,
    Qwen3-VL-8B, and Gemma-4-E4B, graph grounding lifts pooled accuracy by
    \textbf{33.8--37.6} percentage points over image-only inference, while exact
    edge $F_1$ \textbf{0.742} on $G_{\mathrm{auto}}$ states how much substrate error remains.
\end{itemize}

For P\&ID workflows, the practical claim is narrow: graph grounding raises
answer reliability on topology questions, while exact edge $F_1$ on $G_{\mathrm{auto}}$ states how
much sheet error still propagates into those answers.
Section~\ref{sec:related-work} places the work; Section~\ref{sec:overall-framework}
details the two stages; Section~\ref{sec:experiments} leads with \topopid{} results, then
substrate fidelity.

We evaluate open VLMs as the reasoning front-end on \topopid{}, and we also score
cloud OCR and object-localization APIs plus zero-shot line segmenters as
perception controls on the same sheets. Those baselines show that stock vision
services are a weak substitute for a P\&ID-tuned substrate
(Appendix~\ref{sec:app-cloud}), which is why Stage~1 is finetuned perception
plus deterministic graph construction rather than an off-the-shelf detector
stack alone.

\section{Related Work}
\label{sec:related-work}

\paragraph{Grounding and faithful multimodal reasoning.}
Document and chart VQA stress reading and layout
\citep{mathew2021docvqa,masry2022chartqa}; compositional suites isolate
relational reasoning under clean structure
\citep{johnson2017clevr,hudson2019gqa}. VLMs still invent relations that the
pixels never support, especially on dense graphics. Tool-using and multimodal
chain methods reduce free-form guessing by calling external actions or
retrievers \citep{yao2023react,schick2024toolformer,lu2022learnprompt}. We take
the structural route: recover a typed graph from the raster, then require every
topology claim to come from a tool result rather than an unverifiable
rationale.

\paragraph{Graph- and tool-augmented agents.}
Agentic loops that interleave reasoning with external calls
\citep{yao2023react,schick2024toolformer} and graph-query front-ends that
translate questions into executable queries \citep{gupta2025pidqa,alimin2026chatpid}
both assume a clean, given graph or database.
Our setting differs: the agent operates on $G_{\mathrm{auto}}$ built from predicted perception,
and we score both answer accuracy and how far $G_{\mathrm{auto}}$ drifts from $G_{\mathrm{oracle}}$. That couples
perception error to reasoning error instead of assuming a clean,
hand-corrected knowledge base.

\paragraph{Engineering-diagram understanding.}
Digitize-PID and earlier extraction pipelines recover symbols, text, and pipes
before linking them into a graph \citep{paliwal2021digitize,rahul2019pid}.
PIDQA converts P\&ID entities into a labeled property graph and scores
LLM-generated Cypher queries against it, so QA accuracy is reported on a
graph that is taken as the knowledge base rather than as a measured
approximation of the sheet \citep{gupta2025pidqa}. ChatP\&ID~\citep{alimin2026chatpid}
applies GraphRAG to DEXPI-style smart P\&ID models, where node and edge
grounding is largely supplied by the source format. Multi-agent
retrieval systems for process schematics likewise target open-domain question
answering over document context rather than the fidelity of a recovered
topology \citep{sakhinana2024humanlevel}. The gap we target is end-to-end:
raster $\to$ recovered evidence graph $\to$ grounded agent, with graph
imperfection measured rather than assumed away. We also keep dashed instrument links as a separate signal-edge layer, which the
Digitize ground truth connectivity does not model on our partition.

\section{Problem Formulation}
\label{sec:problem-formulation}

Each instance is a marked sheet $I$ and English question $Q$. Gold answer
$a^\star$ is computed from the oracle process graph $G_{\mathrm{oracle}}$ (ground-truth geometry
plus Digitize ground truth connectivity) and is never shown at inference. The deployed system
receives only $(I,Q)$ plus the recovered process graph $G_{\mathrm{auto}}$. An image-only baseline sees
$(I,Q)$; \oursagentname{} sees ($I,Q,G_{\mathrm{auto}}$) and may call a fixed tool
library. We report exact-match accuracy on \topopid{} and graph $F_1$ against $G_{\mathrm{oracle}}$ so
perception and reasoning failures can be separated.

\section{Evidence-Grounded VLM Framework}
\label{sec:overall-framework}

\begin{figure*}[t]
  \centering
  
  \resizebox{0.74\linewidth}{!}{%
  \begin{tikzpicture}[
    font=\sffamily\scriptsize,
    module/.style={
      draw=FigBlue!52,
      fill=FigBlueWash,
      text=FigInk,
      rounded corners=1.6mm,
      line width=0.55pt,
      align=center,
      text width=3.20cm,
      minimum height=6.8mm,
      inner sep=3.5pt,
      font=\sffamily\scriptsize
    },
    inputbox/.style={
      module,
      draw=FigRule,
      fill=white
    },
    detector/.style={
      module,
      text width=2.00cm,
      minimum height=7.2mm
    },
    graphbox/.style={
      module,
      text width=6.70cm,
      minimum height=7.5mm
    },
    fonebox/.style={
      module,
      draw=FigGood!54,
      fill=FigGoodWash,
      text width=2.20cm,
      minimum height=7.2mm
    },
    optionalbox/.style={
      module,
      draw=FigPurple!56,
      fill=FigPurpleWash,
      dashed,
      text width=2.60cm,
      minimum height=7.2mm
    },
    agentbox/.style={
      module,
      draw=FigGood!58,
      fill=FigGoodWash,
      text width=5.30cm,
      minimum height=7.8mm
    },
    answerbox/.style={
      module,
      draw=FigGood!64,
      fill=FigGood!11,
      text width=3.40cm,
      minimum height=7.2mm
    },
    flow/.style={
      -{Stealth[length=1.7mm,width=1.15mm]},
      draw=FigInk!68,
      line width=0.65pt,
      shorten <=1.2pt,
      shorten >=1.2pt
    },
    verticalflow/.style={
      -{Stealth[length=1.7mm,width=1.15mm]},
      draw=FigInk!68,
      line width=0.65pt,
      shorten <=0.6pt,
      shorten >=0.6pt
    },
    optionalflow/.style={
      -{Stealth[length=1.7mm,width=1.15mm]},
      draw=FigPurple!74,
      dashed,
      line width=0.65pt,
      shorten <=1.2pt,
      shorten >=1.2pt
    },
    stagebadge/.style={
      rounded corners=1mm,
      inner xsep=6pt,
      inner ysep=2.5pt,
      font=\sffamily\bfseries\scriptsize,
      text=white
    }
  ]
%
%
  \path[
    draw=FigBlue!30,
    fill=FigBlueWash!42,
    rounded corners=2.5mm,
    line width=0.60pt
  ]
    (-4.35,0.68) rectangle (4.35,-6.10);
%
  \path[
    draw=FigGood!32,
    fill=FigGoodWash!48,
    rounded corners=2.5mm,
    line width=0.60pt
  ]
    (-4.35,-6.38) rectangle (4.35,-10.36);
%
%
  \node[
    stagebadge,
    fill=FigBlue,
    anchor=west
  ] at (-4.00,0.27)
    {STAGE 1\quad GRAPH RECOVERY};
  \node[
    stagebadge,
    fill=FigGood,
    anchor=west
  ] at (-4.00,-6.76)
    {STAGE 2\quad GROUNDED QA};
%
%
  \node[inputbox] (input) at (0,-0.48)
    {P\&ID raster};
  \node[module] (perception) at (0,-1.53)
    {{\mdseries\ourspipeline}\\[-1pt]
     {\color{FigMuted}multimodal perception}};
%
  \node[detector] (symbols) at (-2.85,-2.82)
    {YOLO11s\\[-1pt]
     {\color{FigMuted}symbol detection}};
  \node[detector] (lines) at (0,-2.82)
    {U-Net\\[-1pt]
     {\color{FigMuted}line segmentation}};
  \node[detector] (text) at (2.85,-2.82)
    {YOLO-text + TrOCR\\[-1pt]
     {\color{FigMuted}text recognition}};
%
  \draw[verticalflow]
    (input.south) -- (perception.north);
%
  \draw[flow]
    ([xshift=-1.05cm]perception.south)
    -- (symbols.north);
  \draw[verticalflow]
    (perception.south)
    -- (lines.north);
  \draw[flow]
    ([xshift=1.05cm]perception.south)
    -- (text.north);
%
  \node[graphbox] (graph) at (0,-4.05)
    {Typed graph construction\\[-1pt]
     {\color{FigMuted}
      vectorize $\rightarrow$ node $\rightarrow$
      snap $\rightarrow$ attach tags}};
%
  \draw[flow]
    (symbols.south)
    -- ([xshift=-2.35cm]graph.north);
  \draw[verticalflow]
    (lines.south)
    -- (graph.north);
  \draw[flow]
    (text.south)
    -- ([xshift=2.35cm]graph.north);
%
  \node[fonebox] (fone) at (-1.80,-5.42)
    {$G_{\mathrm{auto}}$ process graph\\[-1pt]
     {\color{FigMuted}typed evidence}};
  \node[optionalbox] (vseven) at (1.80,-5.42)
    {Signal-edge layer\\[-1pt]
     {\color{FigMuted}dashed instrument links}};
%
  \draw[verticalflow]
    ([xshift=-1.80cm]graph.south)
    -- (fone.north);
%
  \draw[optionalflow]
    (fone.east) -- (vseven.west);
%
%
  \node[fonebox] (graphinput) at (-2.00,-7.52)
    {$G_{\mathrm{auto}}$ evidence graph\\[-1pt]
     {\color{FigMuted}recovered topology}};
  \node[inputbox] (question) at (2.00,-7.52)
    {Marked sheet + question\\[-1pt]
     {\color{FigMuted}query $Q$}};
  \node[agentbox] (agent) at (0,-8.71)
    {{\mdseries\oursagentname}\\[-1pt]
     {\color{FigMuted}reasoning loop + graph-tool harness}};
%
  \draw[flow]
    (graphinput.south)
    -- ([xshift=-1.30cm]agent.north);
  \draw[flow]
    (question.south)
    -- ([xshift=1.30cm]agent.north);
  \node[answerbox] (answer) at (0,-9.89)
    {Structured answer\\[-1pt]
     {\color{FigMuted}claim + cited evidence IDs}};
%
  \draw[verticalflow]
    (agent.south) -- (answer.north);
  \end{tikzpicture}%
  }
  
  \caption{
  \textbf{Two-stage graph-grounded reasoning pipeline.}
  Stage~1 converts a P\&ID raster into the typed evidence graph
  $G_{\mathrm{auto}}$ (optional signal-edge layer on dashed instrument
  lines). Stage~2 combines $G_{\mathrm{auto}}$ with the marked sheet and
  question and produces a structured answer through a graph-tool harness.
  The reference graph $G_{\mathrm{oracle}}$ is used only for scoring and
  \topopid{} supervision; it is never provided as agent input.
  }
  
  \label{fig:pipeline}
  \end{figure*}
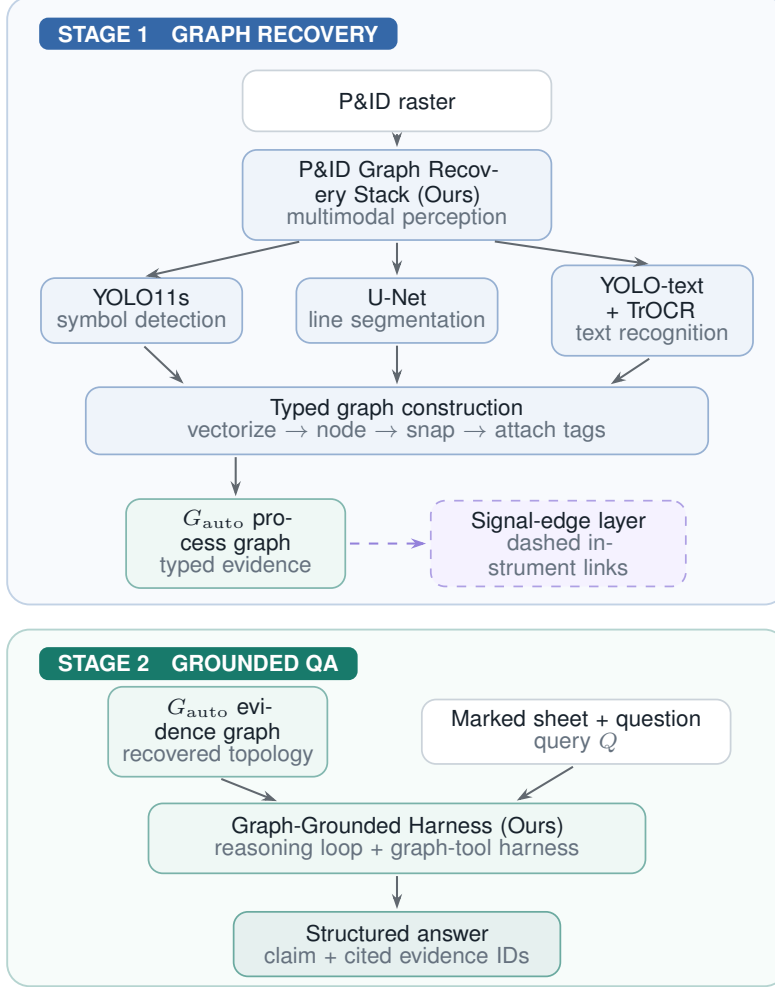

Figure~\ref{fig:pipeline} follows the evidence path from raster to structured
answer. Stage~1 recovers a grounding substrate: multimodal perception finds
symbols, line ink, and text, then deterministic construction turns those
detections into the typed process graph $G_{\mathrm{auto}}$. Stage~2 runs
\oursagentname{} under an answer contract so topology claims must come
from tool results rather than free-form guessing on pixels. Keeping recovery
and QA in separate stages makes it possible to pin a failure on the graph or
on the planner that used it.

\paragraph{Answer contract.}
(1)~No final answer without at least one graph-tool call.
(2)~Every named entity in $Q$ must be queried through a tool before answering.
(3)~For closed counting questions, the number of cited supporting entities must
match the reported count.

\paragraph{Stage 1: grounding-substrate recovery (\ourspipeline{} $\to$ $G_{\mathrm{auto}}$).}
The perception stack runs in fixed order on every sheet: tiled YOLO11s
\citep{jocher2023yolo} symbol
detection (test-set detection $F_1$ \textbf{0.983}), EfficientNet-B0 U-Net
\citep{ronneberger2015unet}
masks for solid and dashed ink (Dice micro \textbf{0.982}), and YOLO-text with
finetuned TrOCR \citep{liu2021trocr} for diagram OCR (detection $F_1$ \textbf{0.833}). Those heads
only locate ink. Topology is built afterward by deterministic geometry on the
predicted solid layer.

Skeletonization turns solid masks into polylines. \emph{Intersection noding}
splits or trims runs that cross at tees and corners so branches share exact
endpoints instead of overshooting past a junction. Endpoints within a
per-sheet snap radius merge into junctions; that radius is set from median
segment length and tip gaps rather than a global pixel threshold (fixed
10\,px snap collapses exact edge $F_1$ to \textbf{0.193} on predicted solids).
Each detected symbol attaches to the nearest solid segment within 120\,px of
its box center. A \emph{first-symbol} walk then traverses empty scaffold
junctions along the segment graph and stops at the first occupied junction,
emitting undirected process-connection edges to the symbols there; symbols
that share a junction form a clique. Two symbols become adjacent only under
this walk---page proximity or a fly-over crossing without a shared junction
never creates an edge. Geometric rules then bind OCR words to symbols and
lines by containment, overlap, and proximity, with line-seed propagation
along the segment graph. The product is $G_{\mathrm{auto}}$: typed symbol
nodes (class and optional tag), undirected process edges, and line-tag
attachments usable by later tools.

\begin{figure*}[t]
\centering
\vspace{-4pt}
\begin{minipage}[t]{0.32\textwidth}
\centering
\includegraphics[width=\linewidth,height=0.16\textheight,keepaspectratio]{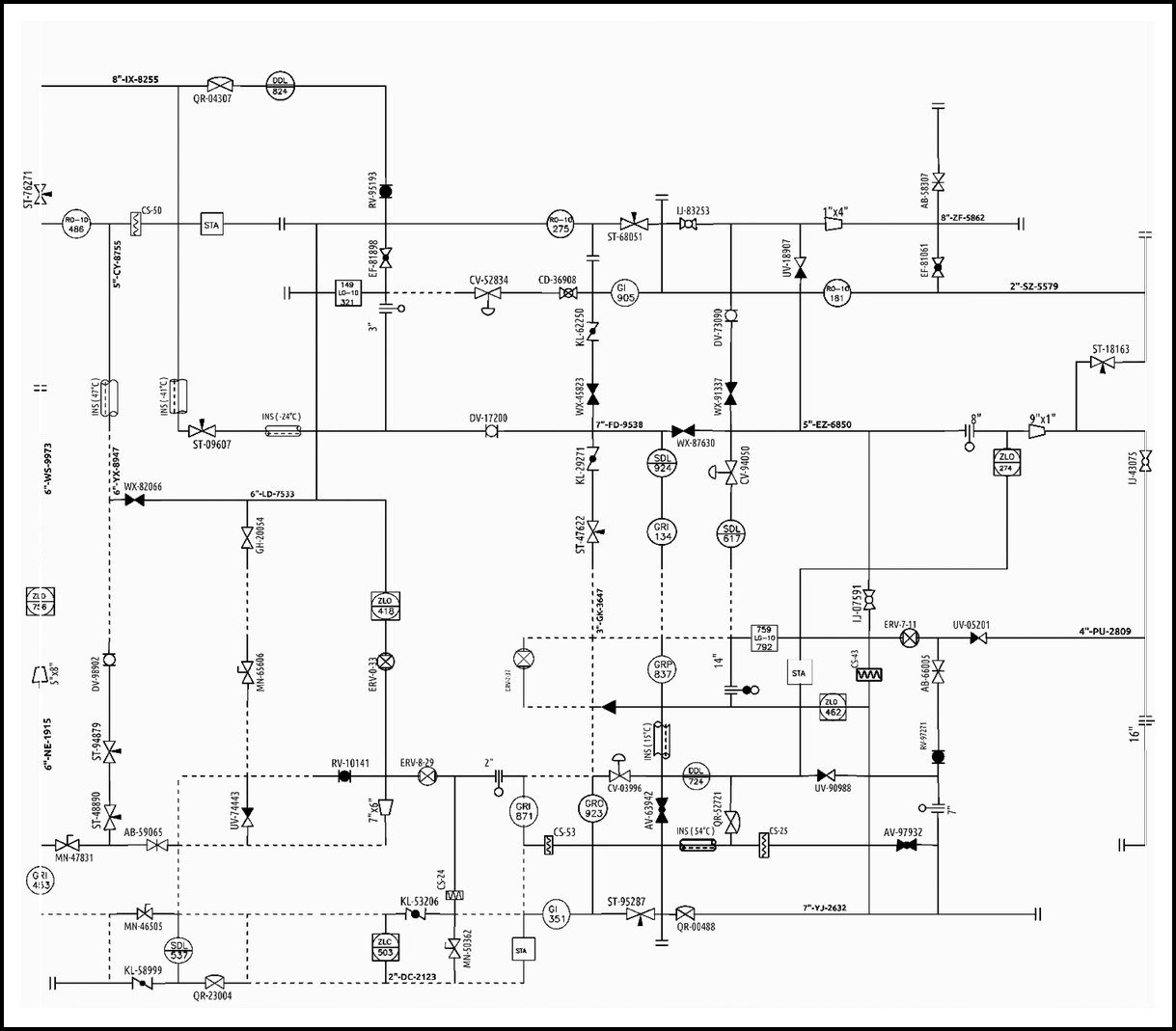}\\[1pt]
{\sffamily\scriptsize\textcolor{FigMuted}{Raw P\&ID raster}}
\end{minipage}\hfill
\begin{minipage}[t]{0.32\textwidth}
\centering
\includegraphics[width=\linewidth,height=0.16\textheight,keepaspectratio]{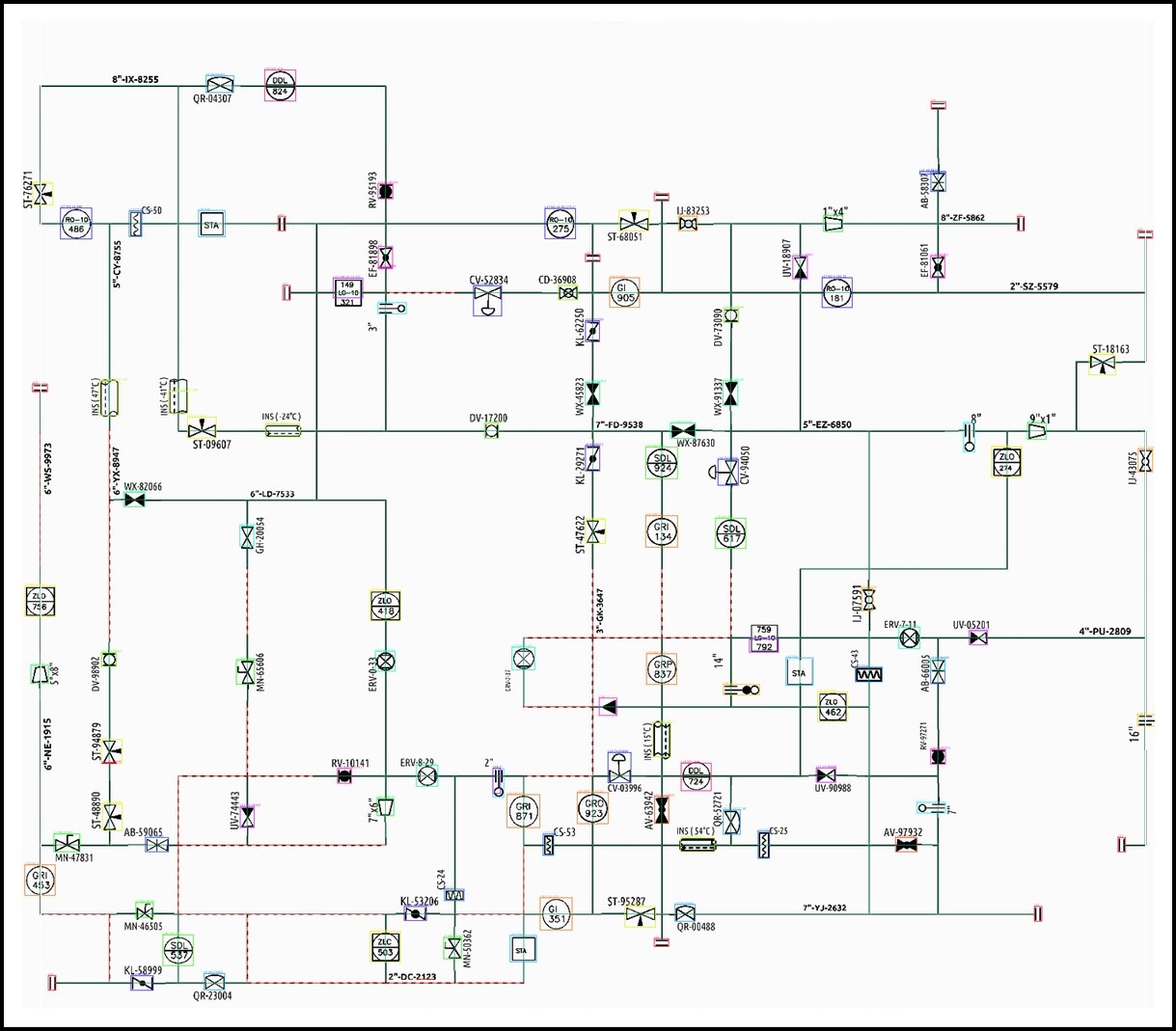}\\[1pt]
{\sffamily\scriptsize\textcolor{FigMuted}{Detected lines, symbols, and text}}
\end{minipage}\hfill
\begin{minipage}[t]{0.32\textwidth}
\centering
\includegraphics[width=\linewidth,height=0.16\textheight,keepaspectratio]{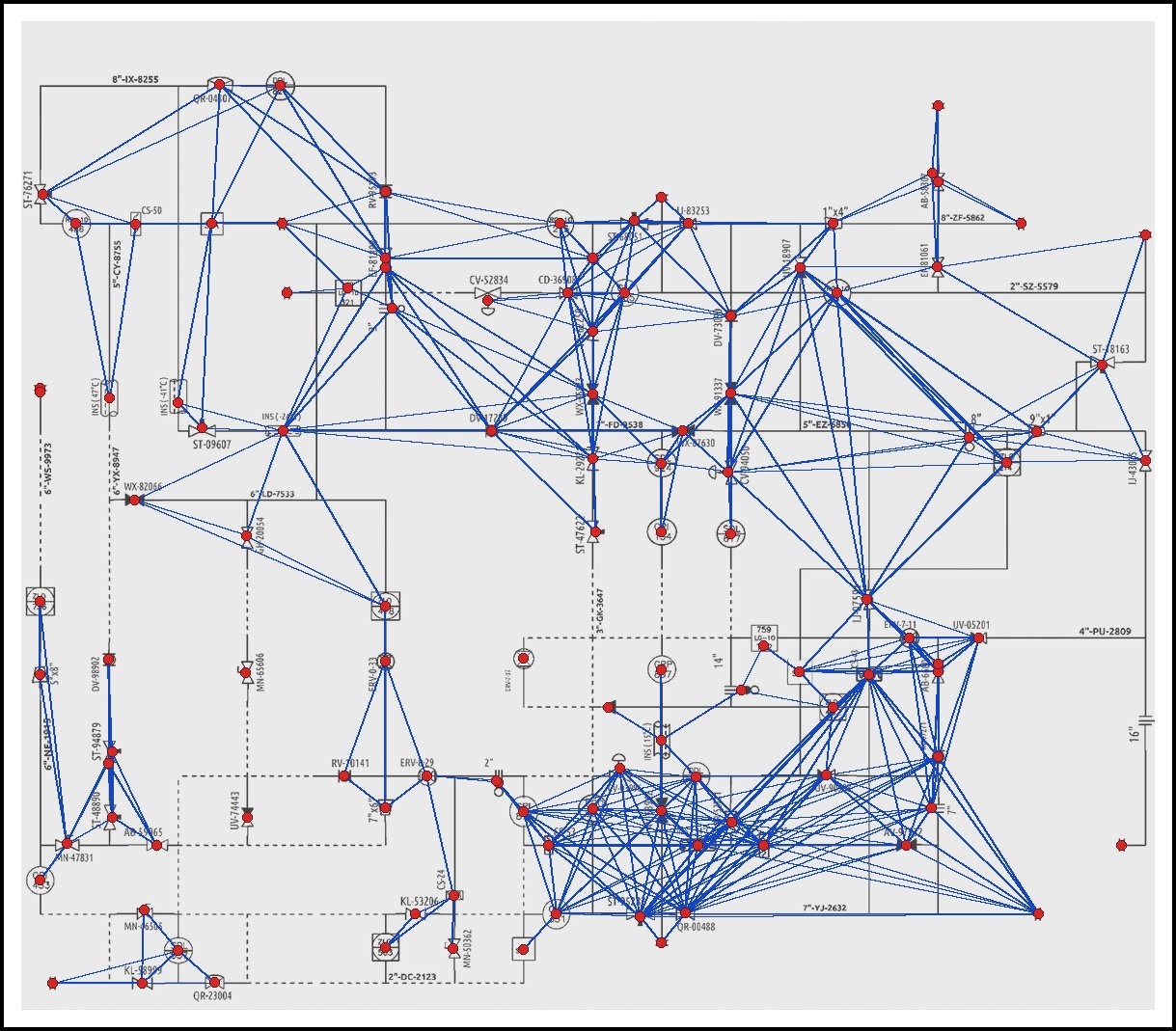}\\[1pt]
{\sffamily\scriptsize\textcolor{FigMuted}{Recovered process graph}}
\end{minipage}
\vspace{-2pt}
\caption{Raster to evidence graph on one Digitize-PID sheet (notes and title
block removed). Left: input drawing. Center: perception overlays for lines,
symbols, and text. Right: process-connection graph recovered from those
detections (nodes at symbol centers; edges from the first-symbol walk).}
\label{fig:raster-to-graph}
\end{figure*}

As an optional extension, a \emph{signal-edge layer} runs the same walk on
dashed instrument polylines with dash-specific adaptive snap (Digitize does
not attach symbols to dashes on our partition). We score that layer as
recovered evidence only; \topopid{} questions use process connections.
Neural nets find ink; deterministic rules build topology---the graph builder
does not invent edges with a learned model.
Figure~\ref{fig:raster-to-graph} shows one Digitize-PID sheet through this
stage: raw raster, perception overlays, and the recovered process graph.

\paragraph{Stage 2: tool-mediated reasoning on $G_{\mathrm{auto}}$.}
Each \topopid{} item is a marked sheet $I$, an English question $Q$, and the
frozen graph $G_{\mathrm{auto}}$ for that sheet. An image-only baseline sees
only $(I,Q)$. \oursagentname{} receives $(I,Q,G_{\mathrm{auto}})$ and answers
through a planner loop capped at six turns. Before the first tool call,
colored mark letters on $I$ (A/B/C) are detected and snapped to nearby
$G_{\mathrm{auto}}$ nodes, and any tag strings named in $Q$ are linked through
OCR tags (with an optional local crop when several nodes share a string).
That seed step is a helper outside the seven graph tools.

The planner may then call node lookup, neighbor listing, node search,
breadth-first traversal, shortest path, edge check, and set operations.
Each call returns structured facts---ids, types, tags, hop lists,
distances, or membership---never a yes/no verdict on $Q$. Counting,
comparison, thresholding, and routing stay in the model's reasoning over
those returns. A typical reachability item resolves marks to node ids, asks
for a breadth-first frontier or shortest-path distance, and only then
states true/false or a bucket label; a tag-choice item searches by type or
prefix and cites the returned ids in the answer schema. The interface is
type-agnostic: the checkpoint is not told whether $Q$ is a hop-count,
tag-choice, or negation item, and it never receives a dumped adjacency
list. Every topology claim must be fetched through a tool, so the trace
records which graph facts supported the answer.

The final message must follow a fixed schema: a short interpretation,
references to the tool steps used, the supporting entity ids, and a typed
answer $\hat{a}\in\mathcal{E}$ (integers clamped to ranges parsed from $Q$;
strings snapped to enumerated options). Answers that skip the tool loop,
omit a named entity query, or fail the counting citation check are rejected
under the answer contract above. The same VLM checkpoint (Qwen3-VL-4B,
Qwen3-VL-8B, or Gemma-4-E4B) plans in text over tool traces. Pixels are
used only for mark detection and optional tag disambiguation; edges, hops,
reachability, and counters come only from $G_{\mathrm{auto}}$ tool results,
never from re-reading pipe ink.

\paragraph{Deployment profile.}
On 100 test sheets \ourspipeline{} spends roughly 1.2\,s/sheet on symbol
stitch, 4.6\,s on line inference, and 22.7\,s on text detection+OCR on paired
16\,GB GPUs; graph build and tagging are CPU-side minutes per batch. That cost
profile targets offline batch recovery followed by interactive queries, not
low-latency full-sheet chat.
\topopid{} stress-tests the published graph product: each sheet carries ten questions
whose gold depends only on $G_{\mathrm{oracle}}$, so improvements in perception, linking, or
grounded inference can be measured on the same benchmark without family-specific
prompt routing.

Detailed construction rules, tool definitions, and ablations appear in
Appendix~\ref{sec:app-graph} and Appendix~\ref{sec:app-agent}.

\section{Experiments}
\label{sec:experiments}
\vspace{-2mm}

\subsection{Experimental Methodology}
\label{sec:experimentalmethodology}
\vspace{-2mm}

We first score grounded versus ungrounded VLM answers on \topopid{}, then report how
faithful the recovered graph is against $G_{\mathrm{oracle}}$. Digitize-PID provides
500 sheets: we use a \textbf{training} / \textbf{validation} / \textbf{test}
split of \textbf{350} / \textbf{50} / \textbf{100} (seed~$0$). The publisher's
400-sheet train folder supplies training and validation; the publisher's
100-sheet val folder is our held-out \textbf{test} set. Perception weights and
graph hyperparameters are chosen on training and validation only; the test set
is never used for model selection. All reported perception, $G_{\mathrm{auto}}$,
and \topopid{} numbers use this same 100-sheet test set. \ourspipeline{}
is frozen before scoring: tiled symbol detection, U-Net line masks,
skeleton vectorization, adaptive junction snap on solid polylines, YOLO-text
with finetuned TrOCR, and geometric tag association. Cloud and zero-shot
baselines use the same test sheets and matching protocols. \topopid{} compares
Qwen3-VL-4B-Instruct, Qwen3-VL-8B-Instruct, and Gemma-4-E4B-it under image-only
inference versus \oursagentname{} over $G_{\mathrm{auto}}$. The graph, tools, and
answer contract are scored as one system (Appendix~\ref{sec:app-agent}).

\subsection{Benchmark Suites}
\label{sec:datasets}
\vspace{-2mm}

\textbf{Digitize-PID.} Five hundred synthetic full-sheet P\&ID rasters with
32 symbol classes, solid and dashed line ink, word boxes, and Digitize
ground truth connectivity. Under our oracle rule a sheet carries about 118 symbols and 338
undirected process edges on average. We train perception on the 350-sheet
\textbf{training} split, use 50 sheets for \textbf{validation} (early stopping),
and evaluate raster-to-graph recovery on the 100-sheet \textbf{test} partition.
Digitize ground truth connectivity on ground-truth geometry is the
process-graph reference ($G_{\mathrm{oracle}}$).

\textbf{\topopid{}}. Three 1000-question suites---easy, medium,
and hard---each with 100 sheets $\times$ 10 questions. Each item contains a marked
diagram, an English question, and (for graph systems) $G_{\mathrm{auto}}$
built from predicted perception. Gold answers are computed from $G_{\mathrm{oracle}}$ only and
are never shown at answer time. Easy questions stay local (booleans and small
integers within one--two hops). Medium and hard add bounded reachability, tag
families, closed string choices, negation, and counterfactual routing.
Appendix~\ref{sec:app-topopid-generation}--\ref{sec:app-topopid-categories} cover
generation, family lists, chance baselines, leakage checks, boolean metrics,
and scores by answer type and category.

\subsection{Experimental Setup}
\label{sec:experimentalsetup}
\vspace{-2mm}

\paragraph{Graph variants.}
$G_{\mathrm{oracle}}$ is the oracle process graph: ground-truth boxes, tags, solid
segments, and Digitize ground truth connectivity for symbol attach.
$G_{\mathrm{auto}}$ is the recovered process graph from predicted symbols, solid
lines, and OCR, plus intersection noding and sheet-adaptive junction snap.
An optional \emph{signal-edge layer} keeps $G_{\mathrm{auto}}$ process
edges and adds dashed instrument links on predicted dashed polylines
(Appendix~\ref{sec:app-graph}); \topopid{} does not query it.

\paragraph{Inference.}
Symbol detection uses 20\% tile overlap and NMS at IoU 0.10. Line masks
are full-sheet; polylines come from skeletonization. OCR for the end-to-end
graph uses diagram-aware crops. Cloud OCR and object-detection controls use
a legend-stripped native-resolution diagram crop (JPEG quality 90). Line
zero-shot baselines use full sheets at 1024\,px long side before resize.
Finetuned training and GPU inference used paired 16\,GB accelerators; graph
construction and scoring are CPU-only.

\paragraph{VLM conditions.}
The image-only condition receives the marked sheet and question. The grounded
condition adds $G_{\mathrm{auto}}$ and applies the seven-tool loop and answer
contract described in Section~\ref{sec:overall-framework}. We evaluate
Qwen3-VL-4B-Instruct, Qwen3-VL-8B-Instruct~\citep{bai2025qwen3vl}, and
Gemma-4-E4B-it~\citep{gemmateam2026gemma4} under the same harness; no condition
receives $G_{\mathrm{oracle}}$ or a gold question-family label at inference.
Decoding details are in Appendix~\ref{sec:app-decoding}.

\paragraph{Gold-query ceiling.}
A non-VLM gold-query ceiling remaps gold entities/queries from
$G_{\mathrm{oracle}}$ onto $G_{\mathrm{auto}}$ and answers from recovered
adjacency. It uses $G_{\mathrm{auto}}$ topology plus gold entity/family
information the harness never sees. Pooled exact-match is
$\approx$88.9\% (not 100\%) because $G_{\mathrm{auto}}$ still errs relative
to $G_{\mathrm{oracle}}$ (exact edge $F_1$ 0.742); see
Appendix~\ref{sec:app-topopid-leakage}.

\subsection{Evaluation Metrics}
\label{sec:evaluationmetrics}
\vspace{-2mm}

Unless noted, TP/FP/FN counts are micro-pooled over all 100 sheets before
precision, recall, and $F_1$:
\begin{align}
\label{eq:prf}
P &= \frac{\mathrm{TP}}{\mathrm{TP}+\mathrm{FP}}, &
R &= \frac{\mathrm{TP}}{\mathrm{TP}+\mathrm{FN}}, &
F_1 &= \frac{2PR}{P+R}.
\end{align}
Boxes and words match greedily at IoU $\ge 0.5$. Symbol detection $F_1$ is
reported in two forms: class-aware (matched class required) and
class-agnostic (geometry only). Unless noted, headline symbol scores are
class-aware; cloud object-API comparisons use the class-agnostic form.
OCR joint $F_1$ requires both a box match and equal normalized strings. \textbf{Exact edge $F_1$} scores
undirected process-connection pairs after symbol alignment.
\textbf{Connectivity-consistency} awards credit to an edge whenever its
endpoints are linked by any path in the other graph (not only by a matching
local edge); we report the mean of per-sheet scores. On dense sheets this
score can stay high even when local wiring is wrong, so exact edge $F_1$
remains the stricter topology metric.
\textbf{Attributed micro-$F_1$} pools matched nodes, exact process edges,
symbol-tags, and line-tags (dashed signal edges excluded from the headline
0.801). Line quality uses micro-pooled pixel Dice/IoU vs thickness-2 GT
polylines. \topopid{} uses exact-match accuracy; on boolean items we also report
$P$, $R$, and $F_1$ with positive class \texttt{true}. Full definitions are in
Appendix~\ref{sec:app-metrics}.

\subsection{Results and Analysis}
\label{sec:resultsanalysis}
\vspace{-2mm}

\paragraph{Grounded VLM reasoning.}
Table~\ref{tab:main-topopid-acc} is the headline comparison: image-only versus
\oursagentname{} on the same recovered $G_{\mathrm{auto}}$ for Qwen3-VL-4B,
Qwen3-VL-8B, and Gemma-4-E4B. Graph grounding improves every tier: pooled
gains of $+35.9$, $+33.8$, and $+37.6$ percentage points
($76.0\pm 1.8$ vs.\ $40.1\pm 1.6$; $75.1\pm 1.4$ vs.\ $41.3\pm 1.7$;
$74.3\pm 1.5$ vs.\ $36.7\pm 1.6$). Hard remains the weakest slice for every
checkpoint, while easy and medium close most of the gap. Boolean $P$/$R$/$F_1$,
per-category accuracy, and answer-type splits---including family-level
checkpoint differences---are in
Appendix~\ref{sec:app-topopid-bool}--\ref{sec:app-topopid-categories}. The gold-query
ceiling on $G_{\mathrm{auto}}$ (Section~\ref{sec:experimentalsetup}) reaches
about 88.9\% pooled. The grounded condition bundles graph access, tools,
grounding, and answer schemas; Table~\ref{tab:nc-main} isolates the answer
contract. A topology overclaim proxy and graph-construction ablations are in
Appendix~\ref{sec:app-hallu} and~\ref{sec:app-ablations}.

\begin{table}[t]
\centering
\caption{\topopid{} exact-match mean accuracy (\%) with sheet-clustered bootstrap
CIs ($B{=}10{,}000$). Centered band: $\Delta=\mathrm{Agent}-\mathrm{Img.}$
(paired half-width). Teal: light $\Delta{<}25$, mid $[25,40)$, strong
$\ge 40$\,pp. $^{\ast}$~=~paired 95\% CI excludes zero.
Appendix~\ref{sec:app-bootstrap}.}
\label{tab:main-topopid-acc}
\scriptsize
\setlength{\tabcolsep}{4.5pt}
\renewcommand{\arraystretch}{1.05}
\begin{tabular}{l *{3}{c c}}
\toprule
& \multicolumn{2}{c}{Qwen3-VL-4B}
& \multicolumn{2}{c}{Qwen3-VL-8B}
& \multicolumn{2}{c}{Gemma-4-E4B} \\
\cmidrule(lr){2-3}
\cmidrule(lr){4-5}
\cmidrule(lr){6-7}
Level
& Img. & Agent
& Img. & Agent
& Img. & Agent \\
\midrule
\multirow{2}{*}{Easy}
& $32.9\pm 2.5$
& $85.7\pm 2.0$
& $34.6\pm 2.4$
& $83.1\pm 2.0$
& $34.3\pm 2.6$
& $82.6\pm 2.4$ \\[-0.4ex]
& \multicolumn{2}{c}{\gainhi{\Delta^{\ast}=+52.8\pm 3.1\ \mathrm{pp}}}
& \multicolumn{2}{c}{\gainhi{\Delta^{\ast}=+48.5\pm 3.6\ \mathrm{pp}}}
& \multicolumn{2}{c}{\gainhi{\Delta^{\ast}=+48.3\pm 3.7\ \mathrm{pp}}} \\
\addlinespace[2pt]
\multirow{2}{*}{Medium}
& $44.5\pm 2.8$
& $78.5\pm 2.8$
& $45.8\pm 3.9$
& $81.2\pm 2.0$
& $38.2\pm 2.4$
& $74.6\pm 2.7$ \\[-0.4ex]
& \multicolumn{2}{c}{\gainmid{\Delta^{\ast}=+34.0\pm 4.6\ \mathrm{pp}}}
& \multicolumn{2}{c}{\gainmid{\Delta^{\ast}=+35.4\pm 4.6\ \mathrm{pp}}}
& \multicolumn{2}{c}{\gainmid{\Delta^{\ast}=+36.4\pm 3.7\ \mathrm{pp}}} \\
\addlinespace[2pt]
\multirow{2}{*}{Hard}
& $42.9\pm 3.7$
& $63.9\pm 3.0$
& $43.5\pm 3.7$
& $60.9\pm 2.7$
& $37.7\pm 2.5$
& $65.8\pm 2.6$ \\[-0.4ex]
& \multicolumn{2}{c}{\gainlo{\Delta^{\ast}=+21.0\pm 3.7\ \mathrm{pp}}}
& \multicolumn{2}{c}{\gainlo{\Delta^{\ast}=+17.4\pm 4.1\ \mathrm{pp}}}
& \multicolumn{2}{c}{\gainmid{\Delta^{\ast}=+28.1\pm 3.7\ \mathrm{pp}}} \\
\midrule
\multirow{2}{*}{Pooled}
& $40.1\pm 1.6$
& $\mathbf{76.0\pm 1.8}$
& $41.3\pm 1.7$
& $75.1\pm 1.4$
& $36.7\pm 1.6$
& $74.3\pm 1.5$ \\[-0.4ex]
& \multicolumn{2}{c}{\gainmid{\Delta^{\ast}=+35.9\pm 2.3\ \mathrm{pp}}}
& \multicolumn{2}{c}{\gainmid{\Delta^{\ast}=+33.8\pm 2.3\ \mathrm{pp}}}
& \multicolumn{2}{c}{\gainmid{\Delta^{\ast}=+37.6\pm 2.1\ \mathrm{pp}}} \\
\bottomrule
\end{tabular}
\end{table}

\paragraph{Answer-contract ablation.}
The headline harness couples graph tools with an answer contract. To measure
what the contract contributes, we rerun all 1000 questions per tier with it
switched off: FINAL no longer requires a prior tool call, empty or illegal
answers are not re-asked (soft and hard re-asks both disabled), the planner
receives no list of allowed options or integer range, and string answers are
not snapped to the enumerated options.
Prompts, tools, grounding, and $G_{\mathrm{auto}}$ are unchanged.
Table~\ref{tab:nc-main} reports accuracy and the paired change from
Table~\ref{tab:main-topopid-acc} (sheet-clustered bootstrap;
Appendix~\ref{sec:app-bootstrap}). Pooled accuracy drops by $0.3$\,pp for both
Qwen checkpoints (paired CIs include zero) and by $3.6\pm 1.0$\,pp for
Gemma-4-E4B (CI excludes zero); the harness gains over image-only in
Table~\ref{tab:main-topopid-acc} are 34--38\,pp. With the contract on, soft
rejects are re-asked up to twice (\texttt{max\_final\_retries}${}={}$2); a
forced FINAL that still returns an illegal string gets one further hard
re-ask. Empty predictions per tier (easy/medium/hard) without the contract are
10/144/27 for Gemma-4-E4B, 0/1/8 for Qwen3-VL-4B, and 6/5/16 for Qwen3-VL-8B;
with the contract they are 1/32/6, 0/1/22, and 9/3/16.

\begin{table}[t]
\centering
\caption{Answer contract switched off: exact-match accuracy (\%) on the full
\topopid{} suites ($n{=}1000$ per tier; pooled $n{=}3000$) with sheet-clustered
bootstrap 95\% CIs ($B{=}10{,}000$). Centered band: $\Delta=$
(no contract)$-$(contract) (paired half-width). Rose: drop when contract off
(light $|\Delta|{<}2$, mid $[2,5)$, strong $\ge 5$\,pp); teal: flat/small gain.
$^{\ast}$~=~paired 95\% CI excludes zero. Appendix~\ref{sec:app-bootstrap}.}
\label{tab:nc-main}
\scriptsize
\setlength{\tabcolsep}{4.5pt}
\renewcommand{\arraystretch}{1.05}
\begin{tabular}{lccc}
\toprule
Level & Qwen3-VL-4B & Qwen3-VL-8B & Gemma-4-E4B \\
\midrule
\multirow{2}{*}{Easy}
  & $\mathbf{85.9\pm 2.0}$
  & $83.4\pm 2.1$
  & $81.6\pm 2.6$ \\[-0.4ex]
  & \ncflat{\Delta=+0.2\pm 0.7}
  & \ncflat{\Delta=+0.3\pm 0.6}
  & \ncdroplo{\Delta^{\ast}=-1.0\pm 0.7} \\
\addlinespace[2pt]
\multirow{2}{*}{Medium}
  & $75.9\pm 2.8$
  & $\mathbf{80.1\pm 2.1}$
  & $64.8\pm 2.7$ \\[-0.4ex]
  & \ncdropmid{\Delta^{\ast}=-2.6\pm 1.7}
  & \ncdroplo{\Delta^{\ast}=-1.1\pm 0.9}
  & \ncdrophi{\Delta^{\ast}=-9.8\pm 2.4} \\
\addlinespace[2pt]
\multirow{2}{*}{Hard}
  & $65.4\pm 2.8$
  & $60.9\pm 2.8$
  & $\mathbf{65.9\pm 2.9}$ \\[-0.4ex]
  & \ncflat{\Delta=+1.5\pm 2.1}
  & \ncflat{\Delta=\phantom{+}0.0\pm 1.5}
  & \ncflat{\Delta=+0.1\pm 1.9} \\
\midrule
\multirow{2}{*}{Pooled}
  & $\mathbf{75.7\pm 1.7}$
  & $74.8\pm 1.4$
  & $70.8\pm 1.6$ \\[-0.4ex]
  & \ncflat{\Delta=-0.3\pm 1.1}
  & \ncflat{\Delta=-0.3\pm 0.7}
  & \ncdropmid{\Delta^{\ast}=-3.6\pm 1.0} \\
\bottomrule
\end{tabular}
\end{table}

\paragraph{Medium tier under no contract.}
Gemma's medium drop ($-9.8\pm 2.4$\,pp; 74.6\%$\to$64.8\%) is only partly
abstention. Without the contract it leaves 144 of 1000 medium answers empty
(32 with the contract), and those empties account for 144 of its 352 medium
errors. They concentrate in four families that need several tool results before
a typed FINAL---\texttt{route\_via\_tag\_family} (38 empty),
\texttt{shared\_junction\_within\_k} (33), \texttt{at\_least\_n\_types} (32),
and \texttt{exists\_type\_within\_k} (31)---where Gemma's family accuracy falls
to 32.7--54.5\%. Empty answers do not explain the whole gap: on
\texttt{count\_options\_same\_type} Gemma scores only 45.0\% with no empties,
while Qwen3-VL-8B reaches 97.0\% on the same family. Both Qwen models also
drop on medium ($-2.6\pm 1.7$ and $-1.1\pm 0.9$\,pp; CIs exclude zero), but
with far fewer empties (1 and 5). Qwen3-VL-8B stays at or above 90\% on
\texttt{count\_options\_same\_type} (97.0), \texttt{nearest\_option\_by\_hops}
(92.0), \texttt{busier\_junction} (90.1), and \texttt{reachable\_within\_k}
(96.0), and leads medium with the contract as well (81.2\% vs.\ 80.1\%), so
its medium edge is not created by the ablation. On hard, Gemma's empties fall
to 27 and accuracy is unchanged (65.9\% vs.\ 65.8\%; paired CI includes zero).
Per-family counts are in Appendix~\ref{sec:app-nc}. A LoRA-finetuned planner
on the same harness collapses to near-constant per-family answers
(Appendix~\ref{sec:app-ft}).

\paragraph{Grounding-substrate fidelity.}
Table~\ref{tab:main-graph} scores $G_{\mathrm{auto}}$ against the oracle. The
symbol node layer is near perfect (class-aware $F_1$ 0.984, recall 0.999).
Exact wiring remains harder (0.742): Digitize expects dense one-hop
neighborhoods, and vectorized polylines overshoot junctions until intersection
noding splits runs at tees and corners. Connectivity-consistency 0.958 replaces
that one-hop test with reachability by any path, so it stays high when symbols
land in the right connected component and only the local links are misplaced.
Tags sit between the two layers---0.866 on symbol tags, 0.730 on line tags,
where one OCR miss can relabel an entire polyline. Pooling the four fact layers
(nodes, exact edges, symbol tags, line tags) gives attributed micro-$F_1$
\textbf{0.801}.

\begin{table}[H]
\centering
\vspace{-4pt}
\caption{Recovered process graph ($G_{\mathrm{auto}}$) as grounding-substrate fidelity against
Digitize ground truth connectivity on 100 test sheets. Only solid
process-connection edges are included. The symbol row scores graph nodes after
class-aware matching inside the graph pipeline; the stitched detector evaluated
under the standalone detection protocol scores $F_1$ 0.983
(Table~\ref{tab:app-a2}).}
\label{tab:main-graph}
\small
\begin{tabular}{lccc}
\toprule
Layer & $P$ & $R$ & $F_1$ \\
\midrule
Symbols (class-aware) & 0.969 & 0.999 & 0.984 \\
Exact edges & 0.677 & 0.820 & 0.742 \\
Conn.\ consistency & --- & --- & 0.958 \\
Symbol tags & 0.792 & 0.954 & 0.866 \\
Line tags & 0.599 & 0.935 & 0.730 \\
Combined tags & --- & --- & 0.798 \\
Attributed (all facts) & 0.730 & 0.886 & \textbf{0.801} \\
\bottomrule
\end{tabular}
\end{table}

\paragraph{Optional signal-edge layer.}
Beyond process topology, we recover a signal-edge layer on dashed instrument
ink (Digitize does not attach symbols to dashes here). Adaptive snap raises
signal exact $F_1$ from 0.610 ($j{=}10$) to \textbf{0.824}
(Table~\ref{tab:main-signal}); typed process+signal exact $F_1$ 0.748,
attributed 0.802. \topopid{} scores process-connection questions only; signal
QA is left for follow-up.

\begin{table}[H]
\centering
\vspace{-6pt}
\caption{Process graph and optional signal-edge layer (100 sheets). Process:
$G_{\mathrm{auto}}$ vs.\ Digitize ground truth connectivity; signal: predicted vs.\ fixed geometric
gold.}
\label{tab:main-signal}
\scriptsize
\setlength{\tabcolsep}{4pt}
\begin{tabular}{lcccc}
\toprule
Layer / rule & Exact $P$ & Exact $R$ & Exact $F_1$ & Conn.\ cons. \\
\midrule
Process connections & 0.677 & 0.820 & 0.742 & 0.958 \\
Signal (pred.\ fixed $j{=}10$) & 0.828 & 0.483 & 0.610 & 0.646 \\
Signal (pred.\ adaptive) & 0.808 & 0.840 & \textbf{0.824} & \textbf{0.869} \\
Typed complete (adaptive signal) & 0.686 & 0.822 & 0.748 & --- \\
\bottomrule
\end{tabular}
\end{table}

\paragraph{Perception and off-the-shelf controls.}
Four frozen modules (Appendix~\ref{sec:app-perception}): YOLO11s det.\ $F_1$
0.983, U-Net Dice 0.982, YOLO-text+TrOCR det.\ $F_1$ 0.833, dashed coverage
$F_1$ 0.993. Same protocols: OCR joint $F_1$ 0.810 (Azure
Read~\citep{azureread2024} 0.779, Textract~\citep{aws_textract} 0.663,
GCP Vision~\citep{gcp_vision} 0.151); class-agnostic symbol det.\ $F_1$ 0.961
(Azure/GCP empty; Rekognition~\citep{aws_rekognition} 0.001); line Dice 0.982
(SAM~2~\citep{ravi2024sam2} 0.012; DeepLab~\citep{chen2018deeplabv3plus} near
zero). Details:
Table~\ref{tab:app-cloud-summary}, Appendix~\ref{sec:app-cloud}. A Gemini 3.8
Flash image-only API baseline is in Appendix~\ref{sec:app-gemini}.

\FloatBarrier
\section{Limitations}
\label{sec:limitations}

Digitize-PID is synthetic; scores do not measure shift to scanned plant
drawings, handwritten redlines, or site-specific standards.
\topopid{} tests three open VLMs under a bundled grounded condition
($G_{\mathrm{auto}}$, tools, mark/tag grounding, answer contract). We ablate
graph construction (Appendix~\ref{sec:app-ablations}) and the answer contract
(Table~\ref{tab:nc-main}); other harness factors remain bundled, so the
image-only lift is not attributed to every component. Pixels are for
marks/tag crops only; topology is tool-read
(Section~\ref{sec:overall-framework}). The signal-edge layer is scored as recovered evidence, not
as \topopid{} QA input. Exact edge $F_1$ on $G_{\mathrm{auto}}$ is 0.742, so a wrong
recovered edge can yield a faithful but incorrect tool answer. There is no
reported calibration or abstention; human review remains necessary for
safety-relevant use.

\section{Conclusion}
\label{sec:conclusion}

When topology cannot be verified from pixels alone, a recovered evidence graph
plus cited tool results turns an unreliable image reader into an inspectable
reasoner. Stage~1 freezes $G_{\mathrm{auto}}$ from predicted ink; Stage~2
answers only through tool traces on that graph. On Digitize-PID,
$G_{\mathrm{auto}}$ reaches attributed micro-$F_1$ \textbf{0.801} (exact edge
$F_1$ \textbf{0.742}; connectivity-consistency \textbf{0.958}). Grounded
access raises pooled \topopid{} accuracy by
\textbf{35.9}/\textbf{33.8}/\textbf{37.6}\,pp for Qwen3-VL-4B/8B and
Gemma-4-E4B (to \textbf{76.0\%}/\textbf{75.1\%}/\textbf{74.3\%} from
\textbf{40.1\%}/\textbf{41.3\%}/\textbf{36.7\%}). Hard-tier misses and
family-level checkpoint gaps
(Appendix~\ref{sec:app-topopid-categories}) show that answers track both
substrate fidelity and multi-step tool use. The answer-contract ablation
(Table~\ref{tab:nc-main}) leaves Qwen pooled accuracy nearly unchanged while
cutting Gemma by $3.6\pm 1.0$\,pp, mostly on medium where empty FINALs rise
from 32 to 144; family-level counts are in Appendix~\ref{sec:app-nc}. Future
work: scanned plant sheets, further harness-factor ablations beyond the answer
contract, abstention, and \topopid{} questions over the signal-edge layer.

\FloatBarrier
\clearpage
\bibliographystyle{plainnat}
\bibliography{references}

\clearpage
\appendix
\raggedbottom

\section{Technical Details and Supplementary Results}
\label{sec:technical_appendix}

\newenvironment{apptable}{%
  \begin{table}[H]%
  \vspace{2pt}%
  \setlength{\abovecaptionskip}{4pt}%
  \setlength{\belowcaptionskip}{4pt}%
}{\end{table}}
\newenvironment{appfig}{%
  \begin{figure}[H]%
  \vspace{2pt}%
  \setlength{\abovecaptionskip}{4pt}%
  \setlength{\belowcaptionskip}{4pt}%
}{\end{figure}}
\setlength{\floatsep}{10pt plus 2pt minus 2pt}
\setlength{\textfloatsep}{12pt plus 2pt minus 2pt}
\setlength{\intextsep}{10pt plus 2pt minus 2pt}

\definecolor{sysInk}{HTML}{172033}
\definecolor{sysMuted}{HTML}{667085}
\definecolor{sysPanel}{HTML}{F7F9FC}
\definecolor{sysPanelBorder}{HTML}{DDE4EE}
\definecolor{sysBlue}{HTML}{2563EB}
\definecolor{sysBlueSoft}{HTML}{EFF6FF}
\definecolor{sysTeal}{HTML}{0F8F83}
\definecolor{sysViolet}{HTML}{7C3AED}
\definecolor{sysVioletSoft}{HTML}{F5F3FF}
\definecolor{sysGreen}{HTML}{0F8F83}
\definecolor{sysGreenSoft}{HTML}{ECFDF5}
\definecolor{sysRed}{HTML}{D64545}
\definecolor{sysRedSoft}{HTML}{FFF1F1}
\definecolor{sysAmber}{HTML}{B45309}
\definecolor{sysAmberSoft}{HTML}{FFF7ED}
\definecolor{sysRedHarness}{HTML}{C2414B}

\tikzset{
  pHeader/.style={
    anchor=west,
    font=\sffamily\bfseries\scriptsize,
    text=sysInk
  },
  pHint/.style={
    anchor=east,
    font=\sffamily\scriptsize,
    text=sysMuted
  },
  pInput/.style={
    rounded corners=2mm,
    draw=sysInk,
    fill=sysInk,
    text=white,
    line width=0.8pt,
    minimum width=4cm,
    minimum height=0.95cm,
    align=center,
    font=\sffamily\bfseries\small
  },
  pModel/.style={
    rounded corners=2mm,
    fill=white,
    line width=0.8pt,
    minimum width=4.3cm,
    minimum height=1.15cm,
    align=center,
    font=\sffamily\bfseries\small
  },
  pProcess/.style={
    rounded corners=1.8mm,
    draw=sysPanelBorder,
    fill=white,
    line width=0.75pt,
    minimum width=3cm,
    minimum height=1.02cm,
    align=center,
    font=\sffamily\bfseries\footnotesize
  },
  pResult/.style={
    rounded corners=2mm,
    draw=sysInk,
    fill=sysInk,
    text=white,
    line width=0.8pt,
    minimum width=3cm,
    minimum height=1.02cm,
    align=center,
    font=\sffamily\bfseries\footnotesize
  },
  pArrow/.style={
    -{Stealth[length=2.5mm,width=1.7mm]},
    draw=sysInk,
    line width=0.95pt,
    rounded corners=1.5mm
  },
  pBranch/.style={
    -{Stealth[length=2.3mm,width=1.6mm]},
    draw=sysMuted,
    line width=0.75pt,
    rounded corners=1.5mm
  },
  pEvidence/.style={
    -{Stealth[length=2.4mm,width=1.7mm]},
    line width=0.9pt,
    densely dashed,
    rounded corners=1.5mm
  },
  pLabel/.style={
    fill=sysPanel,
    inner xsep=1.5mm,
    inner ysep=0.6mm,
    font=\sffamily\scriptsize,
    text=sysMuted
  }
}

\tikzset{figPanel/.style={
  rounded corners=3mm,
  draw=sysPanelBorder,
  fill=sysPanel,
  line width=0.7pt
}}
\tikzset{beforeBadge/.style={
  rounded corners=2mm,
  draw=sysRed!45,
  fill=sysRedSoft,
  text=sysRed,
  inner xsep=3mm,
  inner ysep=1.2mm,
  font=\sffamily\bfseries\scriptsize
}}
\tikzset{afterBadge/.style={
  rounded corners=2mm,
  draw=sysGreen!45,
  fill=sysGreenSoft,
  text=sysGreen,
  inner xsep=3mm,
  inner ysep=1.2mm,
  font=\sffamily\bfseries\scriptsize
}}
\tikzset{operationArrow/.style={
  -{Stealth[length=2.8mm,width=1.9mm]},
  draw=sysInk,
  line width=1pt
}}
\tikzset{operationLabel/.style={
  fill=white,
  text=sysMuted,
  inner xsep=1.4mm,
  inner ysep=0.7mm,
  font=\sffamily\bfseries\scriptsize
}}
\tikzset{geometryLine/.style={
  draw=sysInk,
  line width=1.25pt,
  line cap=round
}}
\tikzset{secondaryLine/.style={
  draw=sysMuted!65,
  line width=1pt,
  line cap=round
}}
\tikzset{figureNote/.style={
  rounded corners=2mm,
  draw=sysPanelBorder,
  fill=white,
  text=sysMuted,
  inner xsep=3mm,
  inner ysep=1.5mm,
  align=center,
  font=\sffamily\footnotesize
}}
\tikzset{occupiedNode/.style={
  circle,
  draw=sysBlue,
  fill=sysBlueSoft,
  line width=1pt,
  minimum size=9mm,
  inner sep=0pt,
  font=\sffamily\bfseries\small,
  text=sysInk
}}
\tikzset{emptyNode/.style={
  circle,
  draw=sysMuted,
  fill=white,
  dashed,
  line width=0.9pt,
  minimum size=9mm,
  inner sep=0pt,
  font=\sffamily\small,
  text=sysMuted
}}
\tikzset{emittedLink/.style={
  -{Stealth[length=2.8mm,width=1.9mm]},
  draw=sysGreen,
  line width=1.25pt
}}
\tikzset{emittedLabel/.style={
  rounded corners=2mm,
  draw=sysGreen!40,
  fill=sysGreenSoft,
  text=sysGreen,
  inner xsep=2mm,
  inner ysep=0.8mm,
  font=\sffamily\bfseries\scriptsize
}}

\tikzset{sectionTitle/.style={
  anchor=west,
  text=sysInk,
  font=\sffamily\bfseries\small
}}
\tikzset{sectionRule/.style={
  draw=sysPanelBorder,
  line width=0.7pt
}}
\tikzset{inputCard/.style={
  rounded corners=2.2mm,
  draw=sysBlue!55,
  fill=sysBlueSoft,
  text=sysInk,
  line width=0.9pt,
  minimum width=3.75cm,
  minimum height=1.15cm,
  align=center,
  inner xsep=4mm,
  font=\sffamily
}}
\tikzset{processCard/.style={
  rounded corners=2.2mm,
  draw=sysPanelBorder,
  fill=white,
  text=sysInk,
  line width=0.85pt,
  minimum width=3.85cm,
  minimum height=1.25cm,
  align=center,
  inner xsep=3mm,
  inner ysep=2mm,
  font=\sffamily
}}
\tikzset{groundCard/.style={
  processCard,
  draw=sysBlue!55,
  fill=sysBlueSoft
}}
\tikzset{seedCard/.style={
  processCard,
  draw=sysViolet!45,
  fill=sysVioletSoft
}}
\tikzset{plannerCard/.style={
  processCard,
  draw=sysViolet!55,
  fill=sysVioletSoft,
  minimum width=2.60cm
}}
\tikzset{operatorCard/.style={
  processCard,
  draw=sysBlue!55,
  fill=sysBlueSoft,
  minimum width=3.00cm
}}
\tikzset{decisionCard/.style={
  diamond,
  aspect=1.55,
  draw=sysAmber!70,
  fill=sysAmberSoft,
  text=sysInk,
  line width=0.9pt,
  minimum width=2.30cm,
  minimum height=1.20cm,
  align=center,
  inner xsep=1mm,
  inner ysep=0.5mm,
  font=\sffamily\bfseries\scriptsize
}}
\tikzset{contractCard/.style={
  rounded corners=2.2mm,
  draw=sysAmber!70,
  fill=sysAmberSoft,
  text=sysInk,
  line width=0.9pt,
  minimum width=3.40cm,
  minimum height=1.45cm,
  align=center,
  inner xsep=3mm,
  inner ysep=2mm,
  font=\sffamily
}}
\tikzset{finalCard/.style={
  rounded corners=2.2mm,
  draw=sysGreen!70,
  fill=sysGreenSoft,
  text=sysGreen!70!black,
  line width=1pt,
  minimum width=3.40cm,
  minimum height=1.20cm,
  align=center,
  inner xsep=3mm,
  inner ysep=2mm,
  font=\sffamily\bfseries
}}
\tikzset{flowArrow/.style={
  -{Stealth[length=2.8mm,width=1.9mm]},
  draw=sysInk,
  line width=1pt,
  rounded corners=1.5mm
}}
\tikzset{feedbackArrow/.style={
  -{Stealth[length=2.8mm,width=1.9mm]},
  draw=sysRedHarness,
  dashed,
  line width=0.95pt,
  rounded corners=1.5mm
}}
\tikzset{edgeLabel/.style={
  fill=white,
  text=sysMuted,
  rounded corners=1mm,
  inner xsep=1.5mm,
  inner ysep=0.7mm,
  font=\sffamily\bfseries\scriptsize
}}
\tikzset{yesLabel/.style={
  edgeLabel,
  text=sysGreen
}}
\tikzset{stageTag/.style={
  rounded corners=1.5mm,
  draw=sysPanelBorder,
  fill=white,
  text=sysMuted,
  inner xsep=2mm,
  inner ysep=0.8mm,
  font=\sffamily\bfseries\scriptsize
}}


\subsection{Graph construction}
\label{sec:app-graph}

Perception recovers symbols, solid/dashed line masks, and OCR tags in a
shared sheet frame. A fixed construction order then turns those outputs
into the typed evidence graph $G_{\mathrm{auto}}$ (Figure~\ref{fig:unified-perception-graph}).
Solid arrows in that figure are geometric transforms; dashed arrows show
where each perception stream may enter. Edges are not created by a learned
module alone.

\begin{appfig}
\centering
\resizebox{\linewidth}{!}{\begin{tikzpicture}[x=1cm,y=1cm]

\path[use as bounding box]
  (-0.05,0.90) rectangle (17.05,12.40);


\filldraw[
  fill=sysPanel,
  draw=sysPanelBorder,
  line width=0.7pt,
  rounded corners=3mm
]
  (0,7.55) rectangle (17,12.35);

\filldraw[
  fill=sysPanel,
  draw=sysPanelBorder,
  line width=0.7pt,
  rounded corners=3mm
]
  (0,0.95) rectangle (17,7.35);


\node[pHeader] at (0.55,11.92)
  {\textbf{I.\quad LEARNED PERCEPTION}};

\node[pHint]
  at (16.45,11.92)
  {one raster \(\rightarrow\) three aligned evidence streams};

\node[pHeader] at (0.55,6.62)
  {\textbf{II.\quad DETERMINISTIC GRAPH CONSTRUCTION}};


\node[pInput] (input)
  at (8.50,10.95)
  {Full-sheet raster};


\node[pModel,draw=sysTeal] (segmenter)
  at (2.80,9.10)
  {
    EffNet-B0 U-Net\\[-1mm]
    {\normalfont\scriptsize\color{sysMuted}
     solid / dashed masks}
  };

\node[pModel,draw=sysViolet] (reader)
  at (8.50,9.10)
  {
    YOLO-text + TrOCR\\[-1mm]
    {\normalfont\scriptsize\color{sysMuted}
     word boxes + strings}
  };

\node[pModel,draw=sysBlue] (detector)
  at (14.20,9.10)
  {
    YOLO11s tiles\\[-1mm]
    {\normalfont\scriptsize\color{sysMuted}
     32 symbol classes}
  };


\draw[pBranch]
  (input.south)
  -- ++(0,-0.42)
  -| (segmenter.north);

\draw[pBranch]
  (input.south)
  -- (reader.north);

\draw[pBranch]
  (input.south)
  -- ++(0,-0.42)
  -| (detector.north);


\node[pProcess] (noding)
  at (2.60,5.15)
  {
    Intersection noding\\[-1mm]
    {\normalfont\scriptsize\color{sysMuted}
     split at crossings}
  };

\node[pProcess] (snap)
  at (6.50,5.15)
  {
    Adaptive snap\\[-1mm]
    {\normalfont\scriptsize\color{sysMuted}
     merge local endpoints}
  };

\node[pProcess] (scaffold)
  at (10.40,5.15)
  {
    Junction scaffold\\[-1mm]
    {\normalfont\scriptsize\color{sysMuted}
     geometry fixes topology}
  };

\node[pProcess] (attach)
  at (14.30,5.15)
  {
    Symbol attach\\[-1mm]
    {\normalfont\scriptsize\color{sysMuted}
     ground detections}
  };

\node[pProcess] (walk)
  at (10.40,2.70)
  {
    First-symbol linker walk\\[-1mm]
    {\normalfont\scriptsize\color{sysMuted}
     trace from scaffold}
  };

\node[pProcess] (associate)
  at (6.50,2.70)
  {
    Tag association\\[-1mm]
    {\normalfont\scriptsize\color{sysMuted}
     bind OCR evidence}
  };

\node[pResult] (graph)
  at (2.60,2.70)
  {
    \(G_{\mathrm{auto}}\)\\[-1mm]
    {\normalfont\scriptsize complete + grounded}
  };


\draw[pEvidence,draw=sysTeal]
  (segmenter.south)
  -- (2.80,7.25)
  -- (0.35,7.25)
  -- node[
       pLabel,
       midway,
       anchor=west,
       xshift=-1.5mm,
       yshift=-1mm
     ]
     {solid-line polylines}
     (0.35,5.15)
  -- (noding.west);

\draw[pEvidence,draw=sysBlue]
  (detector.south)
  -- (detector.south |- attach.north)
  node[
    pLabel,
    pos=0.86,
    anchor=south,
    yshift=1mm
  ]
  {class + bounding boxes};

\draw[pEvidence,draw=sysViolet]
  (reader.south)
  -- (8.50,4.10)
  -- (6.50,4.10)
  -- (associate.north)
  node[
    pLabel,
    pos=0.70,
    above,
    yshift=0.8mm
  ]
  {strings + boxes};


\draw[pArrow]
  (noding) -- (snap);

\draw[pArrow]
  (snap) -- (scaffold);

\draw[pArrow]
  (scaffold) -- (attach);

\draw[pArrow]
  (attach.south)
  -- ++(0,-0.45)
  -| (walk.north);

\draw[pArrow]
  (walk) -- (associate);

\draw[pArrow]
  (associate) -- (graph);


\draw[pArrow]
  (10.00,1.35) -- (10.90,1.35);

\node[
  anchor=west,
  font=\sffamily\scriptsize,
  text=sysMuted
]
  at (11.02,1.35)
  {fixed transform};

\draw[pEvidence,draw=sysBlue]
  (13.20,1.35) -- (14.10,1.35);

\node[
  anchor=west,
  font=\sffamily\scriptsize,
  text=sysMuted
]
  at (14.22,1.35)
  {perception evidence};

\end{tikzpicture}}
\caption{Unified perception-to-graph pipeline. Learned outputs stay in sheet
coordinates and enter only at their grounding ports. The solid construction
path is deterministic: no learned module can invent an edge alone.}
\label{fig:unified-perception-graph}
\end{appfig}

A sheet becomes a graph by a deterministic \emph{first-symbol} walk on a
chosen ink layer. On the solid process layer, intersection noding
(Figure~\ref{fig:noding}) has already split crossing runs, so tee branches
enter the walk sharing exact endpoints.
\begin{enumerate}\setlength{\itemsep}{0pt}\setlength{\parsep}{0pt}\setlength{\topsep}{2pt}
\item Keep solid segments for process connections or dashed-only segments for
  signal edges.
\item Cluster segment endpoints into junctions: endpoints within $j$ pixels
  merge (single-link closure). Here $j$ is the per-sheet adaptive radius
  $j_{\mathrm{eff}}$ (Figure~\ref{fig:snap}), not a global constant, and the
  solid and dashed layers each compute their own.
\item Attach each symbol to one junction. The oracle process graph uses
  Digitize ground truth connectivity; the recovered process graph and every
  signal layer attach to the nearest segment within a fixed 120\,px of the
  symbol center.
\item Walk through \emph{empty} junctions only. When the walk first reaches an
  occupied junction, emit edges to symbols there and stop; symbols on the same
  junction form a clique.
\end{enumerate}
Two symbols are adjacent only under this walk, not from page proximity or
fly-over crossings without a shared junction
(Figure~\ref{fig:walk}).

\begin{appfig}
\centering
\resizebox{0.92\linewidth}{!}{\begin{tikzpicture}[x=1cm,y=1cm]

\path[use as bounding box]
  (-0.05,-0.05) rectangle (15.55,5.50);

\draw[figPanel]
  (0,0) rectangle (15.50,5.45);

\node[afterBadge,anchor=west]
  at (0.45,4.96)
  {FIRST-SYMBOL WALK};


\coordinate (posA)  at (1.75,3.25);
\coordinate (posJ1) at (5.00,3.25);
\coordinate (posJ2) at (8.25,3.25);
\coordinate (posC)  at (11.50,3.25);
\coordinate (posB)  at (5.00,1.65);


\draw[geometryLine]
  (posA) -- (posJ1) -- (posJ2) -- (posC);

\draw[secondaryLine,dashed]
  (posJ1) -- (posB);


\node[occupiedNode] (A)
  at (posA)
  {A};

\node[emptyNode] (J1)
  at (posJ1)
  {$J_1$};

\node[emptyNode] (J2)
  at (posJ2)
  {$J_2$};

\node[occupiedNode] (C)
  at (posC)
  {C};

\node[occupiedNode] (B)
  at (posB)
  {B};


\draw[emittedLink,bend left=20]
  (A.north)
  to node[
    emittedLabel,
    above=1mm
  ]
  {emit $(A,C)$}
  (C.north);

\node[
  font=\sffamily\bfseries\small,
  text=sysRed
]
  at (5.00,2.32)
  {$\times$};

\node[
  font=\sffamily\scriptsize,
  text=sysRed,
  anchor=west
]
  at (5.22,2.32)
  {not emitted};


\node[
  occupiedNode,
  minimum size=5.5mm,
  font=\sffamily\scriptsize
] (occLegend)
  at (13.25,4.82)
  {};

\node[
  font=\sffamily\scriptsize,
  text=sysMuted,
  anchor=west
]
  at (13.58,4.82)
  {symbol};

\node[
  emptyNode,
  minimum size=5.5mm,
  font=\sffamily\scriptsize
] (emptyLegend)
  at (13.25,4.05)
  {};

\node[
  font=\sffamily\scriptsize,
  text=sysMuted,
  anchor=west
]
  at (13.58,4.05)
  {empty};


\node[
  figureNote,
  text width=8.4cm,
  font=\sffamily\scriptsize
]
  at (10.30,0.62)
  {
    From A, the walk passes through empty junctions $J_1$ and $J_2$
    and stops at the first occupied junction C. The side branch to B
    emits no link from A.
  };

\end{tikzpicture}}
\caption{
  First-symbol linker walk. Traversal passes through empty scaffold
  junctions and emits an edge only when the first occupied junction
  on the selected continuation is reached. Spatial nearness or
  bounding-box overlap alone never creates a graph edge.
}
\label{fig:walk}
\end{appfig}

\paragraph{Solid process extras.}
\textbf{Intersection noding} splits or trims solid polylines that cross at
tees and corners so branches share exact endpoints
(Figure~\ref{fig:noding}).

\begin{appfig}
\centering
\resizebox{0.92\linewidth}{!}{\begin{tikzpicture}[x=1cm,y=1cm]

\path[use as bounding box]
  (-0.05,-0.05) rectangle (15.55,4.35);

\draw[figPanel]
  (0,0) rectangle (6.45,4.25);

\draw[figPanel]
  (9.05,0) rectangle (15.50,4.25);

\node[beforeBadge,anchor=west]
  at (0.45,3.78)
  {BEFORE};

\node[afterBadge,anchor=west]
  at (9.50,3.78)
  {AFTER NODING};

\draw[geometryLine]
  (0.90,2.20) -- (5.55,2.20);

\draw[geometryLine]
  (3.20,0.85) -- (3.20,2.48);

\draw[sysRed,line width=0.9pt]
  (3.20,2.20) circle (2.7pt);

\fill[sysRed]
  (3.20,2.48) circle (2.1pt);

\node[
  font=\sffamily\scriptsize,
  text=sysRed,
  anchor=west
]
  at (3.42,2.48)
  {stored tip};

\node[figureNote]
  at (3.22,0.43)
  {crossing has no shared graph vertex};

\draw[operationArrow]
  (6.85,2.12) -- (8.65,2.12)
  node[
    operationLabel,
    midway,
    above=1mm
  ]
  {NODE};

\draw[geometryLine]
  (9.95,2.20) -- (12.25,2.20);

\draw[geometryLine]
  (12.25,2.20) -- (14.60,2.20);

\draw[geometryLine]
  (12.25,0.85) -- (12.25,2.20);

\fill[sysGreen]
  (12.25,2.20) circle (2.8pt);

\draw[sysGreen,line width=0.8pt]
  (12.25,2.20) circle (4.2pt);

\node[
  font=\sffamily\scriptsize,
  text=sysGreen,
  anchor=west
]
  at (12.50,2.48)
  {shared junction};

\node[figureNote]
  at (12.27,0.43)
  {one coordinate, three incident segments};

\end{tikzpicture}}
\caption{
  Intersection noding inserts an exact shared coordinate at a true
  tee intersection. Crossings classified as fly-overs, and crossings
  without a nearby terminating tip, remain separate so that the
  recovered topology is not artificially connected.
}
\label{fig:noding}
\end{appfig}

\textbf{Adaptive junction snap} sets per-sheet
$j_{\mathrm{eff}}=\min(0.12\ell,\max(8,g+0.05\ell))$\,px with $\ell$ the
median solid-segment length and $g$ the median tip nearest-neighbor gap
(Figure~\ref{fig:snap}).
Fixed 10\,px snap on predicted solids collapses exact edge $F_1$ to 0.193.

\begin{appfig}
\centering
\resizebox{0.92\linewidth}{!}{\begin{tikzpicture}[x=1cm,y=1cm]

\path[use as bounding box]
  (-0.05,-0.05) rectangle (15.55,4.55);

\draw[figPanel]
  (0,0) rectangle (6.45,4.45);

\draw[figPanel]
  (9.05,0) rectangle (15.50,4.45);

\node[beforeBadge,anchor=west]
  at (0.45,3.98)
  {FRAGMENTED TIPS};

\node[afterBadge,anchor=west]
  at (9.50,3.98)
  {AFTER SNAP};

\coordinate (p1) at (3.02,2.06);
\coordinate (p2) at (3.34,2.13);
\coordinate (p3) at (3.17,2.40);
\coordinate (cluster) at (3.17,2.18);

\draw[geometryLine]
  (1.25,1.15) -- (p1);

\draw[geometryLine]
  (p2) -- (5.20,2.42);

\draw[geometryLine]
  (p3) -- (3.38,3.50);

\fill[sysRed]
  (p1) circle (2.2pt);

\fill[sysRed]
  (p2) circle (2.2pt);

\fill[sysRed]
  (p3) circle (2.2pt);

\draw[
  sysGreen,
  densely dashed,
  line width=0.9pt
]
  (cluster) circle (0.66cm);

\node[
  font=\sffamily\scriptsize,
  text=sysGreen,
  anchor=west
]
  at (3.76,1.66)
  {$j_{\mathrm{eff}}$};

\node[
  figureNote,
  font=\sffamily\scriptsize
]
  at (3.22,0.43)
  {three endpoints within $j_{\mathrm{eff}}$};

\draw[operationArrow]
  (6.85,2.22) -- (8.65,2.22)
  node[
    operationLabel,
    midway,
    above=1mm
  ]
  {SNAP};

\coordinate (merged) at (12.25,2.18);

\draw[geometryLine]
  (10.30,1.15) -- (merged);

\draw[geometryLine]
  (merged) -- (14.30,2.42);

\draw[geometryLine]
  (merged) -- (12.43,3.50);

\fill[sysGreen]
  (merged) circle (3pt);

\draw[sysGreen,line width=0.8pt]
  (merged) circle (4.4pt);

\node[
  font=\sffamily\scriptsize,
  text=sysGreen,
  anchor=west
]
  at (12.52,2.78)
  {junction $J_0$};

\node[figureNote]
  at (12.27,0.43)
  {one representative coordinate};

\end{tikzpicture}}
\caption{
  Adaptive snapping merges endpoints within the sheet-dependent
  tolerance
  \(
    j_{\mathrm{eff}}
    =
    \min\!\bigl(
      \alpha_{\mathrm{cap}}\ell,\,
      \max(
        j_{\min},\,
        g+\alpha_{\mathrm{slack}}\ell
      )
    \bigr)
  \).
  The capped tolerance limits both under-merging and over-merging.
}
\label{fig:snap}
\end{appfig}

\paragraph{Signal-edge layer.}
Gold signal edges use ground-truth dashed segments with fixed
$j{=}10$\,px. Predicted signal reuses $G_{\mathrm{auto}}$ process edges and builds dashed
edges from U-Net dashed masks. Adaptive dashed $j$ uses
$j_{\mathrm{eff}}=\min(0.12\ell,\max(10,g+0.05\ell))$\,px on predicted
dashed polylines only. On test sheets the oracle dual-layer corpus
averages 337.6 process edges and 31.4 signal edges per sheet; about 20\% of
symbols attach to dashed ink and 7.2 undirected pairs appear in both layers.
Attributed micro-$F_1$ with typed edges is 0.802 (adaptive predicted signal)
and 0.794 (predicted fixed $j{=}10$).

\paragraph{Tag association.}
Word boxes link to symbols and lines by containment, overlap, and proximity,
then line-seed propagation along the segment graph
(\texttt{same\_line\_type\_only}). Word NMS at IoU 0.35 runs before association.

\subsection{Metric definitions}
\label{sec:app-metrics}

Unless noted, TP/FP/FN counts are micro-pooled over all 100 test sheets
before Eqs.~\eqref{eq:prf}. Boxes match greedily: sort candidate pairs by
descending IoU and take a disjoint matching with
\begin{equation}
\label{eq:iou}
\mathrm{IoU}(b_p,b_g)
  =\frac{\mathrm{area}(b_p\cap b_g)}{\mathrm{area}(b_p\cup b_g)},
\qquad
\mathrm{match}\iff\mathrm{IoU}\ge 0.5.
\end{equation}

\paragraph{OCR.}
Let $M$ be the set of IoU-matched word pairs under Eq.~\eqref{eq:iou}.
Detection $F_1$ scores geometry only (matched pairs are TP; unmatched
predictions FP; unmatched GT FN). Joint $F_1$ additionally requires
normalized string equality on the pair. Among matched pairs only,
\begin{align}
\label{eq:ocr-exact-cer}
\mathrm{Exact}
  &= \frac{1}{|M|}\sum_{(p,g)\in M}
     \mathbf{1}\!\bigl[\mathrm{norm}(t_p)=\mathrm{norm}(t_g)\bigr],
\\
\mathrm{CER}
  &= \frac{1}{|M|}\sum_{(p,g)\in M}
     \frac{\mathrm{Lev}(t_p,t_g)}{\max\!\bigl(1,|t_g|\bigr)}
\end{align}
(lower CER is better). Symbol-tag and line-tag $F_1$ use the same string
test on tags linked to matched symbols or line ids; combined tag $F_1$ is
the unweighted mean of those two layer scores.

\paragraph{Symbol detection.}
Class-aware $F_1$ counts a TP only when Eq.~\eqref{eq:iou} holds \emph{and}
class labels agree; class-agnostic $F_1$ uses geometry alone. Then
Eqs.~\eqref{eq:prf} apply. Headline Digitize scores use the class-aware
form; API comparisons in Appendix~\ref{sec:app-cloud} use the
class-agnostic form. Table~\ref{tab:app-a2} reports the same
precision/recall/$F_1$ on full-sheet stitched boxes after NMS.

\paragraph{Edges.}
After class-aware symbol matching $\phi$, let $E^{\mathrm{pred}}$ and
$E^{\mathrm{gt}}$ be undirected process-connection pairs. Exact edge
scoring maps each predicted pair through $\phi$; a mapped pair is TP if it
lies in $E^{\mathrm{gt}}$ and is unused, otherwise FP; unmatched or
collapsed endpoints count as FP; $\mathrm{FN}=|E^{\mathrm{gt}}|-\mathrm{TP}$.
Counts are micro-pooled, then Eqs.~\eqref{eq:prf} yield exact edge $F_1$.

Connectivity-consistency scores reachability rather than one-hop identity.
With adjacency graphs on matched GT indices,
\begin{align}
\label{eq:conn-cons}
\mathrm{TP}_P
  &= \bigl|\{(i,j)\in E^{\mathrm{pred}}:
     \mathrm{reachable}_{G^{\mathrm{gt}}}(\phi(i),\phi(j))\}\bigr|,
\\
\mathrm{TP}_R
  &= \bigl|\{(a,b)\in E^{\mathrm{gt}}:
     \mathrm{reachable}_{G^{\mathrm{pred}}}(a,b)\}\bigr|,
\end{align}
then $P_P=\mathrm{TP}_P/|E^{\mathrm{pred}}|$,
$R_P=\mathrm{TP}_R/|E^{\mathrm{gt}}|$, and the sheet score is the
harmonic mean of $P_P$ and $R_P$. We report the mean over sheets. Any
connecting path suffices, so the score can stay high when local wiring is
wrong; exact edge $F_1$ remains the stricter topology metric.

\paragraph{Attributed micro-$F_1$.}
Each sheet contributes four fact layers: matched nodes, exact process
edges, symbol-tags, and line-tags. Summing TP/FP/FN over layers and sheets
and applying Eqs.~\eqref{eq:prf} yields the headline attributed score
(0.801). The optional signal-edge layer can extend this pool with typed
signal edges (0.802 with adaptive predicted signal).

\paragraph{Line pixels.}
GT polylines rasterize to a thin mask $G$ and the prediction to $P$. With
$I=|P\cap G|$, $N_p=|P|$, $N_g=|G|$ (counts pooled over sheets),
\begin{align}
\label{eq:dice-iou}
\mathrm{Dice}_{\mathrm{micro}}
  &= \frac{2I}{N_p+N_g},
&
\mathrm{IoU}_{\mathrm{micro}}
  &= \frac{I}{N_p+N_g-I}.
\end{align}
Channel-wise solid/dashed scores use the same formulas on each channel.

\paragraph{Dashed coverage.}
GT dashed ink is stored in short chunks; predicted vectorization emits longer
segments. Coverage $F_1$ at tolerance $\tau{=}25$\,px uses many-to-one
matching of chunks to segments (reported 0.993), then Eqs.~\eqref{eq:prf}.

\paragraph{\topopid{} booleans.}
On boolean items only, TP = pred true and gold true; FP = pred true, gold
false; FN = pred false, gold true; Eqs.~\eqref{eq:prf} follow with positive
class \texttt{true}.

\subsection{Perception module scores}
\label{sec:app-perception}

All three perception heads are trained on Digitize-PID~\citep{paliwal2021digitize}
only (500 full-sheet rasters, 32 symbol classes, solid and dashed ink, word
boxes). We use an explicit \textbf{training} / \textbf{validation} /
\textbf{test} split of \textbf{350} / \textbf{50} / \textbf{100} sheets
(seed~$0$). The publisher's 400-sheet train folder is split into the 350-sheet
training set (parameter updates) and the 50-sheet validation set (early
stopping). The publisher's 100-sheet val folder is our held-out test set: every
score in this subsection, and every $G_{\mathrm{auto}}$ and \topopid{} number in
the main paper, is reported on test only. Training and validation never see
test sheets; test is never used to pick weights or graph knobs. Training and
GPU inference used paired 16\,GB accelerators.

\paragraph{Symbols.}
The reported detector is YOLO11s, initialized from the COCO checkpoint and
trained for 50 epochs on 50\%-overlap tiles at $640{\times}640$ (batch 32,
SGD with Ultralytics' default $10^{-2}$ peak LR, weight decay $5{\times}10^{-4}$,
3-epoch warmup, AMP, patience 15, seed~$0$). Letterbox YOLO11s / YOLOv10n
and RT-DETR-L in Table~\ref{tab:app-a1} use the same split, epoch budget,
and seed; only architecture and letterbox vs.\ tiles change. At test time
the tiled YOLO11s run is stitched on full sheets with 20\% overlap; Table~\ref{tab:app-a2}
varies NMS IoU, and IoU~$0.10$ is the operating point used for $G_{\mathrm{auto}}$.

\paragraph{Lines.}
Solid and dashed masks come from a two-channel U-Net with an ImageNet
EfficientNet-B0 encoder (MobileNetV2 is the ablation row). Training uses
$768{\times}768$ tiles with 25\% overlap, batch 4, AdamW at $10^{-4}$ with
cosine decay, BCE+Dice per channel, AMP, 50 epochs, and seed~$0$. Empty
tiles are mostly dropped (minimum mask area 40). Inference is overlapping
tiles on the native-resolution sheet; vectorization is a later CPU step, not
part of this training loop. Table~\ref{tab:app-a3} reports both encoders.

\paragraph{Text.}
Word boxes are predicted by YOLO11n fine-tuned for 15 epochs on 800\,px
patches (batch 16, 25\% train overlap) from the same training/validation
split (350/50). Strings
are read by \texttt{microsoft/trocr-base-printed} fine-tuned for 15 epochs
on ground-truth word crops (batch 32, AdamW $5{\times}10^{-5}$, encoder
frozen, AMP). Table~\ref{tab:app-a4} also reports detection-only YOLO-text
and TrOCR on GT crops (pretrained vs.\ fine-tuned) so box error and
recognition error can be separated. The $G_{\mathrm{auto}}$ stack uses YOLO-text boxes with
the fine-tuned TrOCR head on diagram crops.

\begin{apptable}
\centering
\caption{Symbol detector zoo (Digitize-PID test set, 100 sheets).}
\label{tab:app-a1}
\small
\begin{tabular}{llcccc}
\toprule
Model & Mode & $P$ & $R$ & mAP50 & mAP50--95 \\
\midrule
\textbf{Ours (YOLO11s)} & \textbf{tiles} & \textbf{0.992} & \textbf{0.982} & \textbf{0.994} & \textbf{0.987} \\
YOLOv10n & tiles & 0.989 & 0.979 & 0.994 & 0.983 \\
RT-DETR-L & letterbox & 0.838 & 0.845 & 0.883 & 0.677 \\
YOLO11s & letterbox & 0.784 & 0.783 & 0.824 & 0.613 \\
YOLOv10n & letterbox & 0.291 & 0.488 & 0.281 & 0.180 \\
\bottomrule
\end{tabular}
\end{apptable}

\begin{apptable}
\centering
\caption{Ours (YOLO11s): full-sheet stitch (tiles, 20\% overlap, test set).
$P$, $R$, and Det.\ $F_1$ follow Eqs.~\eqref{eq:prf} after greedy box matching
at IoU\,$\ge 0.5$ (Appendix~\ref{sec:app-metrics}).}
\label{tab:app-a2}
\small
\begin{tabular}{lccc}
\toprule
NMS IoU & $P$ & $R$ & Det.\ $F_1$ \\
\midrule
0.50 & 0.810 & 1.000 & 0.895 \\
0.20 & 0.925 & 0.999 & 0.961 \\
\textbf{0.10} & \textbf{0.969} & \textbf{0.998} & \textbf{0.983} \\
\bottomrule
\end{tabular}
\end{apptable}

\begin{apptable}
\centering
\caption{Line U-Net pixel scores (test set, 100 sheets).}
\label{tab:app-a3}
\small
\begin{tabular}{lcccccc}
\toprule
Backbone & Dice$_{\mu}$ & IoU$_{\mu}$ & Dice$_{\mathrm{sol}}$ & Dice$_{\mathrm{dash}}$ & IoU$_{\mathrm{sol}}$ & IoU$_{\mathrm{dash}}$ \\
\midrule
\textbf{Ours (EffNet-B0)} & \textbf{0.982} & \textbf{0.966} & \textbf{0.983} & \textbf{0.971} & \textbf{0.968} & \textbf{0.946} \\
MobileNetV2 & 0.980 & 0.961 & 0.981 & 0.969 & 0.963 & 0.942 \\
\bottomrule
\end{tabular}
\end{apptable}

\begin{apptable}
\centering
\caption{Text detection and OCR (test set, 100 sheets).}
\label{tab:app-a4}
\small
\begin{tabular}{lccccc}
\toprule
Setting & $P$ & $R$ & Det.\ $F_1$ & Exact & CER$\downarrow$ \\
\midrule
YOLO-text (det.) & 0.712 & 0.999 & 0.831 & --- & --- \\
\quad + pretrained TrOCR (GT crops) & --- & --- & --- & 0.857 & 0.031 \\
Finetuned TrOCR (GT crops) & --- & --- & --- & 0.995 & 0.001 \\
\textbf{Ours (YOLO-text + finetuned TrOCR)} & \textbf{0.772} & \textbf{0.906} & \textbf{0.833} & --- & --- \\
\bottomrule
\end{tabular}
\end{apptable}

\subsection{VLM decoding and run protocol}
\label{sec:app-decoding}

All reported \topopid{} numbers use a single greedy decode per VLM call
(\texttt{do\_sample=False}; no temperature, top-$p$, or top-$k$). Each question
is evaluated once; we do not average multiple samples. Generation budgets:
96 new tokens for entity extraction, 256--512 for planner turns (medium/hard
use the higher cap), and 64 for optional vision crops. Image-only answers use
a 64-token budget. Qwen3-VL-4B-Instruct and Qwen3-VL-8B-Instruct load in 4-bit
NF4 (bitsandbytes double quant, fp16 compute) with a capped visual-token budget
($\le$1280$\times$28$\times$28 pixels) and long-side crops of 336\,px for
grounding / 1024\,px for image-only sheets. Gemma-4-E4B-it uses the public
W4A16 QAT checkpoint. Hub revisions are the instruct/it weights named in
Section~\ref{sec:experimentalsetup} as pulled at run time (no alternate
snapshots). Planner and entity system prompts are fixed strings shared across
models (Appendix~\ref{sec:app-agent}); they were not retuned on \topopid{} items.
Soft contract rejects (empty or unparsable FINAL, missing tool call,
incomplete coverage of named candidates) are re-asked up to twice
(\texttt{max\_final\_retries}${}={}$2); a forced FINAL that still returns an
illegal string gets one further hard re-ask. Transient GPU out-of-memory
triggers at most one generate retry.

\subsection{Bootstrap CI protocol}
\label{sec:app-bootstrap}

Table~\ref{tab:main-topopid-acc} reports sheet-clustered percentile bootstrap
95\% confidence intervals on \topopid{} exact-match accuracy. For each model,
difficulty tier, and condition we load per-question correct/incorrect flags
from the \topopid{} prediction dumps used for Table~\ref{tab:main-topopid-acc}
(each row includes a sheet id). Harness rows exclude gold-query-ceiling runs.
Image-only and harness answers are aligned by question id.

\begin{itemize}\setlength{\itemsep}{0.15em}
\item \textbf{Resampling.} Sheets with replacement ($n{=}100$). All questions
  on a drawn sheet are kept together (10 per tier; 30 for pooled
  easy$+$medium$+$hard). Score CIs resample each condition independently.
  Difference CIs form
  $d_q=\mathbf{1}[\mathrm{agent}_q]-\mathbf{1}[\mathrm{image}_q]$ on shared
  question ids, then average those deltas inside the same sheet multiset.
\item \textbf{Why sheets.} Questions on one sheet share $G_{\mathrm{auto}}$ graph-recovery
  error, so question-level Bernoulli resampling understates variance.
\item \textbf{Settings.} $B{=}10{,}000$ sheet resamples, seed~$0$, percentile
  two-sided 95\% CI. Tables show mean~$\pm$ half-width of that interval.
\item \textbf{Significance.} At $\alpha{=}0.05$ we claim a gain only when the
  paired CI for mean $\Delta$ excludes zero---not from non-overlapping score
  CIs alone. All easy/medium/hard/pooled paired intervals in
  Table~\ref{tab:main-topopid-acc} exclude zero.
\end{itemize}

\subsection{Topology overclaim proxy}
\label{sec:app-hallu}

As a cheap faithfulness check that needs no new VLM runs, we measure a
\textbf{fabricated-true} rate on topology-related boolean questions with gold
answer \texttt{false}: the fraction predicted \texttt{true}. Categories:
\texttt{spatial\_connections}, \texttt{structural}, \texttt{topology},
\texttt{reachability}, \texttt{isolation}, \texttt{negation},
\texttt{constrained\_routing}, \texttt{spatial\_topology}
($n_{\mathrm{gold=false}}=708$ pooled). Lower is better. Among incorrect
harness answers, the share with nonempty \texttt{evidence\_ids}
(faithful-but-wrong) is 45.8\% (Qwen3-VL-4B), 33.4\% (Qwen3-VL-8B), and
52.6\% (Gemma-4-E4B).
Hard-tier image-only fabricated-true for Qwen3-VL-4B is low (5.1\%) because
that checkpoint often answers \texttt{false} on hard booleans; the metric is
an overclaim rate, not overall boolean error. Prefer the pooled row in
Table~\ref{tab:app-hallu}.

\begin{apptable}
\centering
\caption{Fabricated-true rate (\%) on topology-related booleans with gold
\texttt{false}. Lower is better. The centered band under each Img.--Agent pair
is the absolute change: $\downarrow$ is a beneficial reduction,
$\uparrow$ a deterioration. Grounding reduces pooled overclaim for all three
models.}
\label{tab:app-hallu}
\small
\setlength{\tabcolsep}{5.5pt}
\renewcommand{\arraystretch}{1.12}
\begin{tabular}{l *{3}{c c}}
\toprule
& \multicolumn{2}{c}{Qwen3-VL-4B}
& \multicolumn{2}{c}{Qwen3-VL-8B}
& \multicolumn{2}{c}{Gemma-4-E4B} \\
\cmidrule(lr){2-3}
\cmidrule(lr){4-5}
\cmidrule(lr){6-7}
Level
& Img. & Agent
& Img. & Agent
& Img. & Agent \\
\midrule
\multirow{2}{*}{Easy}
& 49.0
& \textbf{10.0}
& 21.0
& \textbf{9.7}
& 33.3
& \textbf{9.0} \\[-0.4ex]
& \multicolumn{2}{c}{\reductionmid{39.0}}
& \multicolumn{2}{c}{\reductionlo{11.3}}
& \multicolumn{2}{c}{\reductionlo{24.3}} \\
\addlinespace[2pt]
\multirow{2}{*}{Medium}
& 66.7
& \textbf{0.0}
& 76.5
& \textbf{0.0}
& 100.0
& \textbf{9.8} \\[-0.4ex]
& \multicolumn{2}{c}{\reductionhi{66.7}}
& \multicolumn{2}{c}{\reductionhi{76.5}}
& \multicolumn{2}{c}{\reductionhi{90.2}} \\
\addlinespace[2pt]
\multirow{2}{*}{Hard}
& \textbf{5.1}
& 34.1
& 70.6
& \textbf{40.8}
& 52.2
& \textbf{42.0} \\[-0.4ex]
& \multicolumn{2}{c}{\worsening{29.0}}
& \multicolumn{2}{c}{\reductionmid{29.8}}
& \multicolumn{2}{c}{\reductionlo{10.2}} \\
\midrule
\multirow{2}{*}{Pooled}
& 37.0
& \textbf{16.5}
& 50.8
& \textbf{18.8}
& 54.5
& \textbf{21.0} \\[-0.4ex]
& \multicolumn{2}{c}{\reductionlo{20.5}}
& \multicolumn{2}{c}{\reductionmid{32.0}}
& \multicolumn{2}{c}{\reductionmid{33.5}} \\
\bottomrule
\end{tabular}

\vspace{4pt}
\begin{minipage}{0.94\linewidth}
\scriptsize
\raggedright
\textit{Notes.}
Boldface marks the lower value in each pair. Teal $\downarrow$ bands report
$\mathrm{Img.}-\mathrm{Agent}$ (improvement): light $<$25\,pp, mid
$[25,40)$, strong $\ge 40$\,pp. Coral $\uparrow$ bands report
$\mathrm{Agent}-\mathrm{Img.}$ (deterioration). Changes use the displayed
values.
\end{minipage}
\end{apptable}

\subsection{Graph ablations and compute}
\label{sec:app-ablations}

Table~\ref{tab:app-graph-ablate} summarizes construction knock-outs already
measured on the Digitize-PID test set. Adaptive junction snap is not optional:
fixed $j{=}10$ on predicted solids collapses exact edge $F_1$ from 0.742 to
0.193. Removing intersection noding costs 1.3 points (0.742$\to$0.729). With
ground-truth boxes and lines, topology exact $F_1$ is 1.000 and combined tag
$F_1$ is 0.918. With predicted boxes and ground-truth lines, connectivity-consistency is
already 0.963. With ground-truth boxes and predicted lines, exact edge $F_1$
is 0.777. Looser symbol NMS (IoU 0.20) gave 0.684 exact / 0.906 connectivity-consistency; IoU 0.10
is used in \ourspipeline{} (detection $F_1$ 0.983). Word NMS 0.35 yields
combined tag $F_1$ 0.798; IoU 0.50 drops it to 0.764.
Table~\ref{tab:app-compute} gives the wall-clock cost of \ourspipeline{}.

\begin{apptable}
\centering
\caption{Selected graph-construction ablations (existing measurements).
Exact edge $F_1$ on recovered process graphs unless noted.}
\label{tab:app-graph-ablate}
\small
\begin{tabular}{lc}
\toprule
Setting & Exact edge $F_1$ \\
\midrule
\ourspipeline{} & \textbf{0.742} \\
Without intersection noding & 0.729 \\
Fixed $j{=}10$ snap on predicted solids (no adaptive) & 0.193 \\
Predicted boxes + GT lines & 0.777 \\
GT boxes + GT lines (topology) & 1.000 \\
\midrule
Signal: pred.\ fixed $j{=}10$ & 0.610 \\
Signal: pred.\ adaptive $j$ & \textbf{0.824} \\
\bottomrule
\end{tabular}
\end{apptable}

\begin{apptable}
\centering
\caption{\ourspipeline{}: wall-clock on 100 test sheets.}
\label{tab:app-compute}
\small
\begin{tabular}{lccc}
\toprule
Stage & Device & s/sheet & 100-sheet total \\
\midrule
Symbol stitch (YOLO11s) & 2$\times$16\,GB GPU & 1.2 & $\sim$2\,min \\
Line U-Net (EffNet-B0) & 2$\times$16\,GB GPU & 4.6 & $\sim$8\,min \\
Text det.\ + OCR & 2$\times$16\,GB GPU & 22.7 & $\sim$38\,min \\
Vectorize + graph + tags & CPU & $\sim$4 & minutes \\
Oracle re-score & CPU & 0.03 & $\sim$3\,s \\
\bottomrule
\end{tabular}
\end{apptable}

\subsection{Cloud baseline detail}
\label{sec:app-cloud}

Cloud OCR and object APIs, plus zero-shot segmenters, are run on the same 100
test sheets and matching rules as \ourspipeline{}
(Tables~\ref{tab:app-cloud-summary}--\ref{tab:app-c3}). On diagram text, Azure
Read is the closest commercial control (joint $F_1$ 0.779 vs.\ our 0.810);
Textract trails on string accuracy, and GCP Vision floods the crop with
detections (385 boxes/sheet) while joint $F_1$ stays at 0.151. General object
APIs do not recover P\&ID symbols: Azure and GCP return empty lists; Rekognition
emits fewer than one box per sheet on average (class-agnostic $F_1$ 0.001)
against our tiled detector at 0.961. Off-the-shelf line segmenters likewise fail on solid/dashed
ink (SAM~2 AMG Dice 0.012; DeepLab variants near zero) while the finetuned
U-Net reaches Dice 0.982. The gap is domain fit, not a matching quirk: stock
vision services are not trained for schematic glyphs or thin process lines.

\begin{apptable}
\centering
\caption{Cloud and zero-shot controls on 100 test sheets. The table gives
the primary metric for each task.}
\label{tab:app-cloud-summary}
\small
\begin{tabular}{llc}
\toprule
Task & System & Primary score \\
\midrule
OCR & Ours (YOLO-text + TrOCR) & Joint $F_1$ \textbf{0.810} \\
OCR & Azure Read & Joint $F_1$ 0.779 \\
OCR & AWS Textract & Joint $F_1$ 0.663 \\
OCR & GCP Vision & Joint $F_1$ 0.151 \\
\midrule
Symbols & Ours (YOLO11s tiles) & Class-agnostic det.\ $F_1$ \textbf{0.961} \\
Symbols & Azure / GCP / AWS & \emph{Nil} / \emph{Nil} / 0.001 \\
\midrule
Lines & Ours (U-Net EffNet-B0) & Dice micro \textbf{0.982} \\
Lines & SAM~2 AMG (union) & Dice micro 0.012 \\
Lines & DeepLabV3 (COCO) & \emph{Nil} \\
Lines & DeepLabV3+ (VOC) & Dice micro 0.0009 \\
\bottomrule
\end{tabular}
\end{apptable}

\begin{apptable}
\centering
\caption{OCR on diagram crop (100 sheets; mean 156.1 GT words per crop).}
\label{tab:app-c1}
\small
\begin{tabular}{lcccccc}
\toprule
System & Det.\ $F_1$ & Joint $F_1$ & Exact & CER & Pred./sheet \\
\midrule
Ours (YOLO-text + TrOCR) & \textbf{0.833} & \textbf{0.810} & 0.971 & 0.008 & 183.3 \\
Azure Read & 0.818 & 0.779 & 0.954 & 0.021 & 163.1 \\
AWS Textract & 0.824 & 0.663 & 0.807 & 0.142 & 162.5 \\
GCP Vision & 0.174 & 0.151 & 0.872 & 0.055 & 385.2 \\
\bottomrule
\end{tabular}
\end{apptable}

\begin{apptable}
\centering
\caption{OCR precision and recall (diagram crop).}
\label{tab:app-c1-prf}
\small
\begin{tabular}{lcccccc}
\toprule
System & Det.\ $P$ & Det.\ $R$ & Det.\ $F_1$ & Joint $P$ & Joint $R$ & Joint $F_1$ \\
\midrule
Ours (YOLO-text + TrOCR) & 0.772 & 0.906 & 0.833 & 0.750 & 0.880 & 0.810 \\
Azure Read & 0.801 & 0.836 & 0.818 & 0.762 & 0.796 & 0.779 \\
AWS Textract & 0.808 & 0.841 & 0.824 & 0.650 & 0.677 & 0.663 \\
GCP Vision & 0.122 & 0.302 & 0.174 & 0.106 & 0.261 & 0.151 \\
\bottomrule
\end{tabular}
\end{apptable}

\begin{apptable}
\centering
\caption{Class-agnostic symbol boxes (diagram crop; mean 118.4 GT symbols per sheet).}
\label{tab:app-c2}
\small
\begin{tabular}{lccccc}
\toprule
System & Det.\ $P$ & Det.\ $R$ & Det.\ $F_1$ & Class-aware $F_1$ & Pred./sheet \\
\midrule
Ours (YOLO11s tiles) & 0.925 & 0.999 & \textbf{0.961} & 0.961 & 127.76 \\
Azure objects & \emph{Nil} & \emph{Nil} & \emph{Nil} & \emph{Nil} & \emph{Nil} \\
GCP Object Localization & \emph{Nil} & \emph{Nil} & \emph{Nil} & \emph{Nil} & \emph{Nil} \\
AWS Rekognition & 0.115 & 0.001 & 0.001 & \emph{Nil} & 0.61 \\
\bottomrule
\end{tabular}
\end{apptable}

\begin{apptable}
\centering
\caption{Line-mask Dice vs.\ thickness-2 GT polylines (full sheets). Zero-shot
rows: long side 1024\,px then resize; ours: native-resolution U-Net tiles.}
\label{tab:app-c3}
\small
\begin{tabular}{lcccc}
\toprule
System & Dice micro & IoU micro & Dice mean & $n$ \\
\midrule
Ours (U-Net EffNet-B0) & \textbf{0.9823} & 0.9658 & 0.9823 & 100 \\
SAM~2 AMG (union) & 0.0121 & 0.0061 & 0.0098 & 100 \\
SAM~2 AMG (elongated) & 0.0005 & 0.0003 & 0.0006 & 100 \\
DeepLabV3 (COCO) & \emph{Nil} & \emph{Nil} & \emph{Nil} & 100 \\
DeepLabV3+ (VOC) & 0.0009 & 0.0004 & 0.0003 & 100 \\
\bottomrule
\end{tabular}
\end{apptable}

\subsection{Grounded Inference Details}
\label{sec:app-agent}

\topopid{} Table~\ref{tab:main-topopid-acc} reports \textbf{\oursagentname{}} over the
$G_{\mathrm{auto}}$. At prediction time the system receives the marked diagram,
the English question, and $G_{\mathrm{auto}}$; it does not receive $G_{\mathrm{oracle}}$, gold mark ids, or a
question-family label. The following sections specify the planner, grounding
rules, answer contract, and graph interface.

\paragraph{VLM planner.}
A VLM planner may call seven fact-returning graph tools from a fixed library,
receives structured tool results, and iterates until it emits a final answer
or exhausts a six-turn budget (at least one tool call required before the
answer is accepted). Tools return graph facts only; counting,
comparison, thresholding, and routing logic are performed in the model's
chain-of-thought. Planning is type-agnostic: the model is not told which
question family it faces.
Marks and tags are linked once on $G_{\mathrm{auto}}$; the planner then issues operators
(at most six turns; at least one before FINAL). Unsupported candidates are
pushed back by the answer-contract check
(Figure~\ref{fig:grounded-harness-loop}).

\begin{appfig}
\centering
\resizebox{\linewidth}{!}{\begin{tikzpicture}[x=1cm,y=1cm]
  \path[use as bounding box]
    (-0.05,-0.05) rectangle (17.55,10.95);

  \draw[
    rounded corners=3mm,
    draw=sysPanelBorder,
    fill=sysPanel,
    line width=0.7pt
  ] (0,0) rectangle (4.35,10.90);

  \draw[
    rounded corners=3mm,
    draw=sysPanelBorder,
    fill=sysPanel,
    line width=0.7pt
  ] (4.65,0) rectangle (17.50,10.90);

  \node[
      text=sysInk,
      font=\sffamily\bfseries\scriptsize,
      align=center
    ] at (2.18,10.27)
      {I. QUESTION GROUNDING};
    
    \draw[sectionRule]
      (0.35,9.92) -- (4.00,9.92);

  \node[sectionTitle] at (5.10,10.27)
    {II. GROUNDED PLANNER LOOP};

  \draw[sectionRule]
    (5.10,9.92) -- (17.05,9.92);


  \node[inputCard] (input) at (2.18,8.90)
    {\textbf{Inputs}\\[-0.2mm]
     \scriptsize $I \cdot Q \cdot F_1$};

  \node[processCard] (extract) at (2.18,6.85)
    {\textbf{Entity extract}\\[-0.2mm]
     \scriptsize VLM};

  \node[groundCard] (ground) at (2.18,4.75)
    {\textbf{Mark + tag link}\\[-0.2mm]
     \scriptsize B \;$\cdot$\; grounding};

  \node[seedCard] (seed) at (2.18,2.55)
    {\textbf{Bootstrap seeds}\\[-0.2mm]
     \scriptsize \texttt{list\_seeds}};

  \draw[flowArrow] (input) -- (extract);
  \draw[flowArrow] (extract) -- (ground);
  \draw[flowArrow] (ground) -- (seed);

  \node[stageTag] at (2.18,0.70)
    {RUN ONCE PER QUESTION};


  \node[plannerCard] (planner) at (7.00,7.75)
    {\textbf{Planner}};

  \node[operatorCard] (execute) at (11.25,7.75)
    {\textbf{Operator on $G_{\mathrm{auto}}$}\\[-0.2mm]
     \scriptsize A \;$\cdot$\; 7 tools};

  \node[decisionCard] (enough) at (15.55,7.75)
    {Enough\\evidence?};

  \node[contractCard] (contract) at (15.55,4.65)
    {\textbf{Answer-contract check}\\[-0.2mm]
     \scriptsize C \;$\cdot$\; snap + score};

  \node[finalCard] (final) at (15.55,2.00)
    {FINAL};

  \draw[flowArrow]
    (seed.east) --
    (5.10,2.55) --
    (5.10,7.75) --
    (planner.west);

  \node[edgeLabel,rotate=90] at (5.10,5.25)
    {GROUNDED SEEDS};

  \draw[flowArrow]
    (planner.east) -- (execute.west)
    node[
      edgeLabel,
      midway,
      above=1mm,
      align=center
    ]
    {TOOL\\JSON};

  \draw[flowArrow]
    (execute.east) -- (enough.west)
    node[
      edgeLabel,
      midway,
      above=1mm,
      align=center
    ]
    {FACT\\JSON};

  \draw[feedbackArrow]
    (enough.north) --
    (15.55,9.20) --
    node[
      edgeLabel,
      midway,
      above=1mm,
      text=sysRed
    ]
    {NO \;$\cdot$\; NEXT TURN $\leq 6$}
    (7.00,9.20) --
    (planner.north);

  \draw[flowArrow]
    (enough.south) -- (contract.north)
    node[yesLabel,midway,right=1mm]
    {YES};

  \draw[flowArrow]
    (contract.south) -- (final.north)
    node[yesLabel,midway,right=1mm]
    {SUPPORTED};

  \draw[feedbackArrow]
    (contract.west) --
    node[
      edgeLabel,
      midway,
      above=1mm,
      text=sysRed
    ]
    {UNSUPPORTED}
    (10.45,4.65) --
    (10.45,6.35) --
    (7.00,6.35) --
    (planner.south);

  \node[stageTag] at (9.75,0.70)
    {AT LEAST ONE OPERATOR BEFORE FINAL};

\end{tikzpicture}}
\caption{Graph-grounded harness and planner loop. Question entities are
grounded once on $G_{\mathrm{auto}}$; the planner composes operators over returned facts.
Missing evidence starts another turn (cap six). Unsupported answers are
pushed back by the answer contract. FINAL requires at least one
operator call plus a passing evidence check.}
\label{fig:grounded-harness-loop}
\end{appfig}

\paragraph{Grounding and answer contract.}
\textbf{Grounding.} Colored mark letters are detected on the raster
(crimson A, blue B, green C) and snapped to nearest $G_{\mathrm{auto}}$ symbol nodes. Tag
strings mentioned in the question are linked to nodes via OCR tags on $G_{\mathrm{auto}}$, with
optional VLM crops when several nodes share a tag. \textbf{Answer contract.} The final message uses a fixed schema: a short
interpretation, references to the tool steps used, the supporting graph
entities, and a typed answer. Integers are clamped to ranges parsed from the question;
strings are snapped to enumerated options. \textbf{Role split.} Qwen3-VL-4B,
Qwen3-VL-8B, and Gemma-4-E4B plan in text over tool traces. Vision is used only
for entity extraction, tag disambiguation, and mark grounding---not for
inferring process connectivity from pipe ink.

\paragraph{Graph tool interface.}
Table~\ref{tab:app-agent-tools} lists the seven primitives with arguments and
return semantics.
All topology reads come from $G_{\mathrm{auto}}$ adjacency. Breadth-first traversal returns hop
distances (and optional target restrictions); the model interprets reachability,
within-$k$, and bucket labels. Edge check is the only tool that returns a
boolean verdict. The planner does not receive serialized subgraph dumps; every
fact is fetched through typed tools and recorded in a structured trace. Each
tool result gets an identifier that the final answer must cite. A seed-listing
helper bootstraps grounded mark/tag labels before the
first planner turn and is not counted among the seven tools.

\begin{apptable}
\centering
\caption{\oursagentname{} tool library: seven fact-returning primitives on $G_{\mathrm{auto}}$
(arguments and return semantics).}
\label{tab:app-agent-tools}
\scriptsize
\setlength{\tabcolsep}{3.5pt}
\begin{tabular}{lp{2.35cm}p{3.1cm}p{5.9cm}}
\toprule
Group & Tool & Arguments & Behavior \\
\midrule
Nodes & \texttt{node\_info} & \texttt{\{id\}} &
Class name, OCR tag, degree, and sheet coordinates for one node. \\
& \texttt{neighbors} & \texttt{\{id\}} &
One-hop neighbor ids with class, tag, degree, and coordinates. \\
\midrule
Search & \texttt{find\_nodes} & \texttt{\{attr, match, value\}} &
All node ids whose \texttt{class} or \texttt{tag} matches \texttt{value}
(\texttt{equals} or case-insensitive \texttt{prefix}), with count and a compact
attribute table. \\
\midrule
Paths & \texttt{bfs} & \texttt{\{id, ids?, exclude\_ids?, max\_hops?\}} &
Hop distances from \texttt{id}; optional \texttt{ids} restricts the returned
map. Unreachable targets receive \texttt{null}. \\
& \texttt{shortest\_path} & \texttt{\{a, b, exclude\_ids?\}} &
Shortest node path \texttt{a..b} (inclusive), or \texttt{null} if none.
Optional \texttt{exclude\_ids} are not traversed. \\
& \texttt{edge\_exists} & \texttt{\{a, b\}} &
Boolean: true iff an undirected process-connection edge is present in $G_{\mathrm{auto}}$. \\
\midrule
Sets & \texttt{set\_op} & \texttt{\{op, a[], b[]\}} &
Set algebra on id lists: \texttt{intersect}, \texttt{union}, or \texttt{diff}. \\
\bottomrule
\end{tabular}
\end{apptable}

\paragraph{Harness system prompts.}
\label{sec:app-harness-prompts}
Entity extraction and planning use the two fixed system prompts below, shared
by all three checkpoints in both the contract and no-contract runs. Both are
text-only calls; images are used only for mark and tag grounding. The entity
call receives \texttt{Question: \ldots{} JSON:}. Each planner call receives
the question, the expected answer type, the grounded mark and tag seeds, the
seed facts returned by \texttt{list\_seeds}, the tool catalog, and all earlier
tool results, and must return either one tool call or a FINAL JSON object
followed by an \texttt{Answer:} line. With the contract on, the user turn also
lists the allowed string options or the integer range and requires a fact for
every named candidate; with it off, these additions are omitted.

\paragraph{Entity-extraction system prompt.}
\begin{promptbox}
You read a question about a piping-and-instrumentation diagram (P&ID).
Extract the entities the question refers to.
Reply with ONLY one JSON object on one line. No markdown.
Schema: {"marks":["A"],"tags":["TAG-1"],"answer_type":"boolean"}
- marks: letter labels mentioned (A, B, C) -- empty list if none
- tags: EVERY equipment/instrument tag or identifier mentioned as a symbol
  name, including every option listed in a multiple-choice question --
  empty list if none
- answer_type: "boolean" if yes/no, "integer" if a count is required,
  "string" if the answer is a tag, a letter choice, or a labeled bucket
  (e.g. "3-5")
\end{promptbox}

\paragraph{Planner system prompt.}
\begin{promptbox}
You are a careful graph-reasoning agent for piping-and-instrumentation
diagrams (P&ID). You may use exactly seven tools. Every tool returns a
graph FACT. You must do the THINKING: counts, comparisons, thresholds,
negation, universal checks, buckets, and the final yes/no or choice.

Chain of thought (required):
- Before the first tool, decide which facts you need (ids, attributes,
  distances, a path, a set). Do not skip gathering.
- After each ToolResult, update what you know and what is still missing.
  Call another tool if a fact is missing. Never guess an id, hop count,
  class, or tag.
- In the FINAL interpretation, write the computation explicitly: e.g.
  'hops(A,X)=2, hops(A,Y)=5, 2<=3 so X counts, 5>3 so Y does not; count=1'.
  The answer field is only the verdict; the interpretation must show the work.

Hard rules:
- TRUST tool results. Do NOT invent node ids.
- Only these tools exist: node_info, neighbors, find_nodes, bfs,
  shortest_path, edge_exists, set_op. If you name any other tool the call
  fails.
- edge_exists is the only tool that returns a boolean. bfs/shortest_path
  return distances/paths; YOU decide reachable / closer / within-k /
  same-network.
- find_nodes returns a set. YOU count it, filter it further, or pass it
  as exclude_ids. It does not answer the question.
- bfs: pass ids=[the nodes you care about] so the hop map stays small.
  hops[t] is null if unreachable. Use exclude_ids to skip nodes (removed
  equipment, an entire class/tag-family you already collected with
  find_nodes). Use max_hops when the question states a hop budget.
- shortest_path: use when you need the route itself or length_edges.
  Same exclude_ids rule. Endpoints are never excluded.
- Grounded seeds are already in the prompt (marks A/B/C and named tags).
  Look up each named option's id from Grounded seeds before measuring it.
- Direct connection = edge_exists, not a short bfs.
- Local type/degree questions: node_info + neighbors, then YOU count how
  many neighbors share a class, how many distinct classes, or who has
  higher degree.
- Attribute lookup over the sheet (a class, a tag, or a tag prefix):
  find_nodes, then compose with bfs/set_op/shortest_path.
- Sheet regions: node_info gives xy; bootstrap includes sheet midpoint.
  YOU assign quadrants (x>=mid_x and y<mid_y is top-right if y grows down).
- Multiple named symbols: you MUST obtain a fact about EACH option
  (bfs with all option ids, or node_info/neighbors per id) before FINAL.
- Integer 'how many of these': answer 0-4 (or the stated range). Put only
  passing ids in evidence_ids. 0 is valid.
- String answers: copy one allowed option EXACTLY (tag text, 'A'/'B', or
  a bucket like '3-5'). Never a node id, never 'Mark A', never true/false.
- Boolean answers: true/false only.
- Gather tools BEFORE answering. Never answer from a guess.

Each turn reply with ONE JSON object only:
  {"tool":"<name>","args":{...}}
OR a FINAL object (interpretation MUST contain the arithmetic):
  {"interpretation":"<step-by-step>",
   "steps":[{"op":"...","result_ids":[...]}],
   "evidence_ids":[...],"answer": <bool|int|string>}
Then a line: Answer: true|false|<int>|<string>
- evidence_ids: positive witnesses only; [] if answer is false, 0, or absence.
- For integer counts, len(evidence_ids) must equal the answer.
Example tool turn:
{"tool":"bfs","args":{"id":"n12","ids":["n3","n7"],"max_hops":4}}
Example FINAL turn:
{"interpretation":"bfs hops n3=2, n7=5; 2<=3 so n3 passes; 5>3 so n7 fails;
 count=1","steps":[{"op":"bfs","result_ids":["n3"]}],
 "evidence_ids":["n3"],"answer":1}
Answer: 1
Emit no markdown fences and no prose outside the JSON / Answer line.
\end{promptbox}

\subsection{Answer-contract ablation: details}
\label{sec:app-nc}

The main-text answer contract (Section~\ref{sec:overall-framework}) is three
semantic requirements: at least one graph-tool call before FINAL, every named
entity in $Q$ queried through a tool, and cited supporting entities matching
the reported count on closed counting questions. Runtime enforcement also
clamps integers to ranges parsed from the question, snaps strings to
enumerated options, and re-asks illegal FINALs (soft rejects up to
\texttt{max\_final\_retries}${}={}$2, then one hard re-ask;
Appendix~\ref{sec:app-decoding}). The ablation in
Section~\ref{sec:resultsanalysis} turns those enforcement switches off
together with the tool-before-FINAL requirement: no prior-tool check, soft and
hard re-asks disabled, no allowed-options or integer-range prompt, and no
string snap. Prompts, tools, grounding, and $G_{\mathrm{auto}}$ stay
unchanged. Section~\ref{sec:resultsanalysis} reports the re-ask budget,
empty-prediction counts, and the medium-tier reading; here
Table~\ref{tab:app-nc-medium} gives per-family accuracy and empty counts on
medium. One Qwen3-VL-8B medium episode ended with a runtime error and is
scored incorrect. Gemma's empty medium answers concentrate in
\texttt{route\_via\_tag\_family} (38), \texttt{shared\_junction\_within\_k}
(33), \texttt{at\_least\_n\_types} (32), and \texttt{exists\_type\_within\_k}
(31).

\begin{apptable}
\centering
\caption{Answer contract off, medium tier: accuracy (\%) by question family.
Parentheses: empty predictions (shown when nonzero).}
\label{tab:app-nc-medium}
\scriptsize
\setlength{\tabcolsep}{4pt}
\begin{tabular}{lrrrr}
\toprule
Family & $n$ & Qwen3-VL-4B & Qwen3-VL-8B & Gemma-4-E4B \\
\midrule
\texttt{at\_least\_n\_types}               & 101 & 80.2 (1) & 76.2 (3) & 54.5 (32) \\
\texttt{busier\_junction}                 & 101 & 48.5     & 90.1     & 82.2 \\
\texttt{count\_options\_same\_type}        & 100 & 80.0     & 97.0     & 45.0 \\
\texttt{exists\_type\_within\_k}           & 101 & 61.4     & 68.3 (2) & 52.5 (31) \\
\texttt{nearest\_option\_by\_hops}         & 100 & 88.0     & 92.0     & 72.0 \\
\texttt{reachable\_within\_k}             & 101 & 96.0     & 96.0     & 96.0 \\
\texttt{route\_via\_tag\_family}           & 101 & 66.3     & 54.5     & 32.7 (38) \\
\texttt{same\_network}                    & 101 & 84.2     & 67.3     & 96.0 (1) \\
\texttt{same\_type\_neighbors\_at\_least\_n} &  93 & 80.6     & 86.0     & 77.4 (9) \\
\texttt{shared\_junction\_within\_k}       & 101 & 74.3     & 74.3     & 40.6 (33) \\
\midrule
All medium                                & 1000 & 75.9 (1) & 80.1 (5) & 64.8 (144) \\
\bottomrule
\end{tabular}
\end{apptable}

\subsection{LoRA-finetuned planner}
\label{sec:app-ft}

We LoRA-finetune each checkpoint as the harness planner on 200 questions per
tier and evaluate on the remaining 800; the two splits share sheets but no
questions. Table~\ref{tab:app-ft} gives accuracy. Pooled accuracy is
47.5--52.6\%, below the untuned harness in Table~\ref{tab:main-topopid-acc}
(74.3--76.0\%); the evaluation sets differ (800 held-out vs.\ 1000
questions), so we do not report a paired difference.
We adapt each planner with QLoRA (4-bit NF4; LoRA on \(q/k/v/o\) projections
with \(r{=}16\), \(\alpha{=}32\)) for 4~epochs at learning rate
\(2{\times}10^{-4}\) (batch size~1, gradient accumulation~8).
Supervision is text-only SFT on teacher tool trajectories (tool JSON through
FINAL) built with gold marks/tags and the oracle graph; sheet pixels are not
used in planner fine-tuning.

The finetuned planners mostly return one fixed answer per family. In 19
(Gemma-4-E4B), 17 (Qwen3-VL-4B), and 18 (Qwen3-VL-8B) of the 28 families, at
least 95\% of held-out predictions are identical. On easy, Qwen3-VL-4B makes
exactly one tool call on all 800 questions and Qwen3-VL-8B on 796. Differences
between checkpoints in Tables~\ref{tab:app-ft}--\ref{tab:app-ft-hard}
therefore mostly reflect which fixed answer matches the gold majority. On
\texttt{busier\_junction}, for example, both Qwen planners answer \texttt{A}
on all 81 questions (50.6\%, the share of gold \texttt{A}), while
Gemma-4-E4B returns an empty string or \texttt{None} on every question
(0.0\%). With only 200 training questions per tier, the adapters largely
collapse to near-constant per-family answers rather than improved multi-step
tool use, and held-out pooled accuracy (47.5--52.6\%) stays near the suite
majority baseline (48.1\%; Table~\ref{tab:app-topopid-maj-rand}). We therefore
report these runs as a negative adaptation check and do not treat checkpoint
ranking as evidence that LoRA improves the harness.

\begin{apptable}
\centering
\caption{LoRA-finetuned planner: exact-match accuracy (\%) on the held-out
800 questions per tier (pooled $n{=}2400$) with sheet-clustered bootstrap
95\% CIs ($B{=}10{,}000$).}
\label{tab:app-ft}
\small
\begin{tabular}{lcccc}
\toprule
Model & Easy & Medium & Hard & Pooled \\
\midrule
Qwen3-VL-4B & $68.5\pm 2.8$ & $47.5\pm 5.0$ & $41.9\pm 2.8$ & $52.6\pm 2.3$ \\
Qwen3-VL-8B & $66.9\pm 2.8$ & $43.9\pm 5.1$ & $44.9\pm 3.4$ & $51.9\pm 2.6$ \\
Gemma-4-E4B & $67.6\pm 2.9$ & $33.1\pm 4.1$ & $41.8\pm 3.7$ & $47.5\pm 2.4$ \\
\bottomrule
\end{tabular}
\end{apptable}

\begin{apptable}
\centering
\caption{LoRA-finetuned planner, easy tier: held-out accuracy (\%) by family.}
\label{tab:app-ft-easy}
\scriptsize
\setlength{\tabcolsep}{4pt}
\begin{tabular}{lrrrr}
\toprule
Family & $n$ & Qwen3-VL-4B & Qwen3-VL-8B & Gemma-4-E4B \\
\midrule
\texttt{connected\_process}           & 75  & 56.0 & 56.0 & 56.0 \\
\texttt{connection\_exists\_tag}       & 75  & 85.3 & 68.0 & 76.0 \\
\texttt{count\_common\_neighbors}      & 175 & 85.1 & 85.1 & 85.1 \\
\texttt{count\_same\_class\_neighbors} & 175 & 88.0 & 88.0 & 88.0 \\
\texttt{has\_same\_class\_neighbor}    & 75  & 44.0 & 44.0 & 44.0 \\
\texttt{is\_branch}                   & 75  & 44.0 & 44.0 & 44.0 \\
\texttt{share\_neighbor}              & 75  & 52.0 & 52.0 & 52.0 \\
\texttt{tag\_connected\_to\_mark}      & 75  & 45.3 & 45.3 & 45.3 \\
\bottomrule
\end{tabular}
\end{apptable}

\begin{apptable}
\centering
\caption{LoRA-finetuned planner, medium tier: held-out accuracy (\%) by family.}
\label{tab:app-ft-medium}
\scriptsize
\setlength{\tabcolsep}{4pt}
\begin{tabular}{lrrrr}
\toprule
Family & $n$ & Qwen3-VL-4B & Qwen3-VL-8B & Gemma-4-E4B \\
\midrule
\texttt{at\_least\_n\_types}               & 81 & 49.4 & 49.4 & 49.4 \\
\texttt{busier\_junction}                 & 81 & 50.6 & 50.6 & 0.0 \\
\texttt{count\_options\_same\_type}        & 80 & 58.8 & 26.3 & 15.0 \\
\texttt{exists\_type\_within\_k}           & 81 & 58.0 & 58.0 & 6.2 \\
\texttt{nearest\_option\_by\_hops}         & 80 & 12.5 & 11.3 & 12.5 \\
\texttt{reachable\_within\_k}             & 81 & 43.2 & 43.2 & 43.2 \\
\texttt{route\_via\_tag\_family}           & 81 & 49.4 & 45.7 & 53.1 \\
\texttt{same\_network}                    & 81 & 53.1 & 54.3 & 53.1 \\
\texttt{same\_type\_neighbors\_at\_least\_n} & 73 & 50.7 & 50.7 & 50.7 \\
\texttt{shared\_junction\_within\_k}       & 81 & 49.4 & 49.4 & 49.4 \\
\bottomrule
\end{tabular}
\end{apptable}

\begin{apptable}
\centering
\caption{LoRA-finetuned planner, hard tier: held-out accuracy (\%) by family.}
\label{tab:app-ft-hard}
\scriptsize
\setlength{\tabcolsep}{4pt}
\begin{tabular}{lrrrr}
\toprule
Family & $n$ & Qwen3-VL-4B & Qwen3-VL-8B & Gemma-4-E4B \\
\midrule
\texttt{all\_prefix\_in\_network}   & 69 & 56.5 & 55.1 & 39.1 \\
\texttt{count\_options\_within\_k}  & 81 & 13.6 & 40.7 & 16.0 \\
\texttt{counterfactual\_reconnect} & 82 & 51.2 & 50.0 & 50.0 \\
\texttt{distance\_bucket}          & 78 & 30.8 & 28.2 & 23.1 \\
\texttt{more\_options\_near\_a}     & 82 & 47.6 & 34.1 & 46.3 \\
\texttt{multi\_constraint\_option} & 82 & 23.2 & 37.8 & 41.5 \\
\texttt{no\_prefix\_neighbor}      & 81 & 46.9 & 50.6 & 46.9 \\
\texttt{path\_avoiding\_type}      & 82 & 50.0 & 50.0 & 50.0 \\
\texttt{region\_neighbor\_exists}  & 81 & 48.1 & 48.1 & 48.1 \\
\texttt{route\_comparison}         & 82 & 52.4 & 54.9 & 54.9 \\
\bottomrule
\end{tabular}
\end{apptable}

\subsection{Cloud image-only baseline (Gemini API)}
\label{sec:app-gemini}

We query Gemini 3.8 Flash with the image-only protocol of the local baseline:
the marked sheet (long side 1024) and the question, without graph tools, on all
1000 questions per tier, scored by the same exact-match rule.

Pooled accuracy is 36.9\% (Table~\ref{tab:app-gemini}), below the best local
image-only model (41.3\%) and 37.4\,pp below the lowest harness score in
Table~\ref{tab:main-topopid-acc} (74.3\%).

\begin{apptable}
\centering
\caption{Gemini 3.8 Flash image-only exact-match accuracy (\%) on the full
\topopid{} suites ($n{=}1000$ per tier; pooled $n{=}3000$) with sheet-clustered
bootstrap 95\% CIs ($B{=}10{,}000$).}
\label{tab:app-gemini}
\small
\begin{tabular}{lcccc}
\toprule
Model & Easy & Medium & Hard & Pooled \\
\midrule
Gemini 3.8 Flash & $42.2\pm 2.7$ & $34.8\pm 3.0$ & $33.7\pm 2.8$ & $36.9\pm 1.7$ \\
\bottomrule
\end{tabular}
\end{apptable}

\subsection{\topopid{} question generation}
\label{sec:app-topopid-generation}

Gold answers never pass through a language model. Construction is:

\begin{enumerate}
  \item \textbf{Sheets.} Digitize-PID test partition, seed~0 (100 sheets),
    the same split used for perception reporting.
  \item \textbf{Oracle graph $G_{\mathrm{oracle}}$.} Ground-truth boxes, Digitize ground truth connectivity, and
    ground-truth tags. Every gold answer is computed on $G_{\mathrm{oracle}}$ alone.
  \item \textbf{Family sampling.} Typed candidates are drawn from $G_{\mathrm{oracle}}$ under a
    fixed catalog of 28 question families (8 easy, 10 medium, 10 hard). After
    validation each sheet keeps 10 easy, 10 medium, and 10 hard items.
  \item \textbf{Stems.} Deterministic English stems with slots for marks and
    tags (for example, ``Are marks A and B directly connected by a process
    line?'').
  \item \textbf{Wording.} The released suite uses those stems as written
    (template wording; no paraphrase stage on the frozen set).
  \item \textbf{Freeze.} Each kept item ships
    with a marked-sheet crop for visual reference.
  \item \textbf{Inference vs.\ scoring.} $G_{\mathrm{oracle}}$, mark-to-node maps, and family labels
    are scoring-only. At answer time the harness sees the marked image, the
    English question, and recovered $G_{\mathrm{auto}}$, and must resolve entities through tools.
\end{enumerate}

\subsection{\topopid{} benchmark families}
\label{sec:app-topopid-families}

Tables~\ref{tab:app-topopid-easy}--\ref{tab:app-topopid-hard} list every
family with its answer type and question stem.

\paragraph{Easy (8 families, 100 items each).}
600 booleans + 400 integers. Categories: \texttt{spatial\_connections} (300),
\texttt{simple\_counting} (400), \texttt{structural} (300).

\begin{apptable}
\centering
\caption{Easy \topopid{} question families.}
\label{tab:app-topopid-easy}
\footnotesize
\begin{tabular}{llp{5.6cm}}
\toprule
Family & Ans. & Example \\
\midrule
\texttt{connected\_process} & bool & Are marks A and B directly connected by a process line? \\
\texttt{connection\_exists\_tag} & bool & Is symbol MN-46505 directly connected to symbol RO-10 181? \\
\texttt{tag\_connected\_to\_mark} & bool & Is symbol CS-25 directly connected to mark A? \\
\texttt{share\_neighbor} & bool & Do marks A and B share a common directly connected neighbor? \\
\texttt{has\_same\_class\_neighbor} & bool & Does mark A connect to a same-type neighbor? \\
\texttt{is\_branch} & bool & Is mark A a branch point ($\ge 3$ direct connections)? \\
\texttt{count\_same\_class\_neighbors} & int & How many neighbors of mark A share A's component type? \\
\texttt{count\_common\_neighbors} & int & How many symbols connect to both A and B? \\
\bottomrule
\end{tabular}
\end{apptable}

\paragraph{Medium (10 families).}
699 booleans, 201 closed strings, 100 integers in $[0,4]$. Categories:
\texttt{reachability}, \texttt{tag\_reasoning}, \texttt{topology},
\texttt{type\_search}.

\begin{apptable}
\centering
\caption{Medium \topopid{} question families.}
\label{tab:app-topopid-medium}
\footnotesize
\begin{tabular}{llp{5.2cm}}
\toprule
Family & Ans. & Example \\
\midrule
\texttt{reachable\_within\_k} & bool & Is B reachable from A in $\le 3$ connections? \\
\texttt{same\_network} & bool & Are A and B on the same connected network? \\
\texttt{same\_type\_neighbors\_at\_least\_n} & bool & Does A have $\ge 1$ same-type neighbor? \\
\texttt{exists\_type\_within\_k} & bool & Within 2 hops of A, any symbol of B's type? \\
\texttt{shared\_junction\_within\_k} & bool & A symbol within 2 hops of both A and B? \\
\texttt{route\_via\_tag\_family} & bool & Route A$\to$B in $\le 4$ hops through a tag family? \\
\texttt{at\_least\_n\_types} & bool & $\ge 8$ distinct component types on A's neighbors? \\
\texttt{nearest\_option\_by\_hops} & tag & Closest to A among four named symbols? \\
\texttt{busier\_junction} & A/B & Which mark has more direct connections? \\
\texttt{count\_options\_same\_type} & 0--4 & How many of four named symbols match A's type? \\
\bottomrule
\end{tabular}
\end{apptable}

\paragraph{Hard (10 families).}
699 booleans, 200 strings, 101 integers. Eight categories including
\texttt{constrained\_routing}, \texttt{isolation}, \texttt{quantifier}.

\begin{apptable}
\centering
\caption{Hard \topopid{} question families.}
\label{tab:app-topopid-hard}
\footnotesize
\begin{tabular}{llp{5.2cm}}
\toprule
Family & Ans. & Example \\
\midrule
\texttt{more\_options\_near\_a} & bool & More named symbols within 3 hops of A than of B? \\
\texttt{path\_avoiding\_type} & bool & Can A reach B while bypassing C's component class? \\
\texttt{counterfactual\_reconnect} & bool & If A were removed, would B still connect to C? \\
\texttt{route\_comparison} & bool & Is A closer to C than B is (hop count)? \\
\texttt{multi\_constraint\_option} & tag & Tag near A and same type as B (4 options)? \\
\texttt{all\_prefix\_in\_network} & bool & All \texttt{AB-*} tags on same network as A? \\
\texttt{no\_prefix\_neighbor} & bool & No \texttt{ST-*} symbol directly connected to A? \\
\texttt{count\_options\_within\_k} & 0--4 & How many of four symbols within 4 hops of A? \\
\texttt{distance\_bucket} & bucket & Shortest-route distance bucket between A and B? \\
\texttt{region\_neighbor\_exists} & bool & Any neighbor of A in the top-right region? \\
\bottomrule
\end{tabular}
\end{apptable}

\subsection{\topopid{} answer distributions and chance baselines}
\label{sec:app-topopid-baselines}

Table~\ref{tab:app-topopid-answer-dist} gives the answer-type mix and
Table~\ref{tab:app-topopid-maj-rand} the majority-class and random-choice
floors that every reported accuracy must clear.
\textbf{Majority:} always predict the most common gold label of that type
(e.g.\ always \texttt{false} for booleans, always $0$ for integers), then
micro-average across types.
\textbf{Random:} pick uniformly from the observed label set of that type
(expected accuracy $1/|\mathcal{Y}|$).

\begin{apptable}
\centering
\caption{\topopid{} answer-type counts (exact-match scoring).}
\label{tab:app-topopid-answer-dist}
\small
\begin{tabular}{lrrrr}
\toprule
Level & Boolean & Integer & String & Total \\
\midrule
Easy   & 600 & 400 & 0   & 1000 \\
Medium & 699 & 100 & 201 & 1000 \\
Hard   & 699 & 101 & 200 & 1000 \\
\midrule
Pooled & 1998 & 601 & 401 & 3000 \\
\bottomrule
\end{tabular}
\end{apptable}

\begin{apptable}
\centering
\caption{Chance baselines (\% exact match). Majority always predicts the most
frequent gold label within each answer type, then micro-averages. Random draws
uniformly from the observed label set of that answer type (expected accuracy
$1/|\mathcal{Y}_{\mathrm{type}}|$).}
\label{tab:app-topopid-maj-rand}
\small
\begin{tabular}{lrr}
\toprule
Split & Majority (by answer type) & Random (by answer type) \\
\midrule
Easy   & 64.8 & 35.0 \\
Medium & 42.4 & 37.2 \\
Hard   & 39.7 & 37.2 \\
\midrule
Pooled & 48.1 & 35.6 \\
\bottomrule
\end{tabular}
\end{apptable}

Pooled label modes: boolean majority 50.3\% (\texttt{false}), random 50.0\%
($n{=}1998$); integer majority 64.7\% (mode $0$), random 11.1\% over nine
values ($n{=}601$); string majority 12.7\% (mode \texttt{A}), random 0.5\% over
202 values ($n{=}401$). Easy majority is pulled up by sparse integer counts
(many zeros). Medium and hard sit closer to balanced booleans plus harder
string options.
Image-only pooled accuracy in Table~\ref{tab:main-topopid-acc}
($36.7$--$41.3\%$) sits below the majority floor ($48.1\%$) and near the
random floor ($35.6\%$): the ungrounded VLMs do not even exploit label
frequency, which matches the claim that pixel-only answers invent or miss
process connections. The gold-query ceiling on $G_{\mathrm{auto}}$ still
reaches about $88.9\%$ pooled
(Section~\ref{sec:experimentalsetup}), above the grounded harness
($74.3$--$76.0\%$), so the suite is not solved; the remaining gap tracks
substrate error and multi-step tool use.

\subsection{\topopid{} template diversity and duplicates}
\label{sec:app-topopid-diversity}

The suite has 28 question families (8 easy, 10 medium, 10 hard). Across 3000
items there are 792 unique question strings after whitespace and case
normalization. Many sheets reuse the same stem and the same mark letters
(A/B); variation sits in the underlying entities and graphs rather than in
lexical paraphrase. Masking mark letters and tag-like tokens leaves about 511
distinct surface forms---a rough upper bound on stem variants. Eighty-four
normalized strings appear more than once, while every question identifier is
unique (3000/3000). Repeated stems across sheets are expected under a fixed
template catalog; they are not duplicate gold instances. The frozen evaluation
set uses template stems only.

\subsection{\topopid{} prompt and leakage checklist}
\label{sec:app-topopid-leakage}

\begin{enumerate}
  \item At answer time the harness and the image-only baseline never receive
    $G_{\mathrm{oracle}}$, gold mark-to-node maps, gold graph queries, or a question-family label.
  \item System prompts and tool schemas were fixed before the final \topopid{} runs.
    No \topopid{} questions, stems, or family labels were used to tune those prompts
    for the accuracies in Table~\ref{tab:main-topopid-acc}.
  \item Gold answers are never shown in the planner or vision loop; scoring is
    offline exact match.
  \item The lab-only gold-query ceiling ($\approx$88.9\% pooled) receives gold
    entity maps and family/query structure and answers them on
    $G_{\mathrm{auto}}$ (Section~\ref{sec:experimentalsetup}). It is an upper
    bound for comparison, not a deployable method.
\end{enumerate}

\subsection{\topopid{} accuracy by answer type}
\label{sec:app-topopid-answer-type}

Category tables (Appendix~\ref{sec:app-topopid-categories}) do not show whether
gains are boolean-only. Tables~\ref{tab:app-topopid-atype-pooled}
and~\ref{tab:app-topopid-atype-level} report exact-match accuracy (\%) by answer
type on the same dumps as Table~\ref{tab:main-topopid-acc}.

\begin{apptable}
\centering
\caption{Pooled \topopid{} exact-match accuracy (\%) by answer type. Image-only vs.\
\oursagentname{}; gold-query ceiling uses gold entities and
query structure on $G_{\mathrm{auto}}$.}
\label{tab:app-topopid-atype-pooled}
\small
\begin{tabular}{llrrr}
\toprule
Model & System & Bool & Int & Str \\
\midrule
\multirow{3}{*}{Qwen3-VL-4B}
  & Image-only & 50.3 & 9.7 & 34.9 \\
  & \oursagent{} & 75.3 & 84.7 & 66.8 \\
  & Gold-query ceiling & 89.9 & 92.0 & 79.1 \\
\midrule
\multirow{3}{*}{Qwen3-VL-8B}
  & Image-only & 51.2 & 12.6 & 35.2 \\
  & \oursagent{} & 75.7 & 82.2 & 61.1 \\
  & Gold-query ceiling & 89.9 & 92.0 & 79.1 \\
\midrule
\multirow{3}{*}{Gemma-4-E4B}
  & Image-only & 51.9 & 11.0 & 0.0 \\
  & \oursagent{} & 77.8 & 74.4 & 57.1 \\
  & Gold-query ceiling & 89.9 & 92.0 & 79.1 \\
\bottomrule
\end{tabular}
\end{apptable}

\begin{apptable}
\centering
\caption{Per-level exact-match (\%) by answer type for Qwen3-VL-4B-Instruct.
Easy has no string items.}
\label{tab:app-topopid-atype-level}
\small
\begin{tabular}{llrrr}
\toprule
Level & System & Bool & Int & Str \\
\midrule
\multirow{3}{*}{Easy}
  & Image-only & 52.0 & 4.2 & --- \\
  & \oursagent{} & 87.0 & 83.8 & --- \\
  & Gold-query ceiling & 90.7 & 89.2 & --- \\
\midrule
\multirow{3}{*}{Medium}
  & Image-only & 48.8 & 25.0 & 39.3 \\
  & \oursagent{} & 77.1 & 91.0 & 77.1 \\
  & Gold-query ceiling & 92.0 & 100.0 & 82.1 \\
\midrule
\multirow{3}{*}{Hard}
  & Image-only & 50.4 & 15.8 & 30.5 \\
  & \oursagent{} & 63.4 & 82.2 & 56.5 \\
  & Gold-query ceiling & 87.1 & 95.0 & 76.0 \\
\bottomrule
\end{tabular}
\end{apptable}

Harness gains are not boolean-only: integer accuracy moves from single-digit
or low-twenties (image-only) into the 70--90 range, and string accuracy roughly
doubles for Qwen3-VL-4B. Hard booleans remain the weakest harness slice
(63.4\% for 4B), in line with negation, counterfactual, and constrained-routing
families. Gold-query ceiling strings ($\sim$79\% pooled) still miss some items under $G_{\mathrm{auto}}$
noise even with perfect queries.

\subsection{\topopid{} boolean metrics}
\label{sec:app-topopid-bool}

Table~\ref{tab:app-topopid-bool} reports boolean $P$, $R$, and $F_1$ (positive class
\texttt{true}) for the same image-only vs.\ harness runs as
Table~\ref{tab:main-topopid-acc}. Harness pooled $F_1$ is 73.3\% for Qwen3-VL-4B
(image-only 44.7\%), 74.9\% for Qwen3-VL-8B (52.6\%), and 78.2\% for
Gemma-4-E4B (53.0\%). Easy harness $F_1$ is highest for every model; hard
image-only recall for Qwen3-VL-4B collapses (4.0\%), while the harness
recovers to 59.2\%.

\begin{apptable}
\centering
\caption{\topopid{} boolean subset for Qwen3-VL-4B-Instruct, Qwen3-VL-8B-Instruct, and
Gemma-4-E4B-it: $P$, $R$, $F_1$ (\%, positive class \texttt{true}). Boolean
counts: easy 600; medium 699; hard 699; pooled 1998.}
\label{tab:app-topopid-bool}
\small
\setlength{\tabcolsep}{4pt}
\begin{tabular*}{\linewidth}{@{\extracolsep{\fill}} lrrr rrr rrr @{}}
\toprule
& \multicolumn{3}{c}{Qwen3-VL-4B} & \multicolumn{3}{c}{Qwen3-VL-8B} & \multicolumn{3}{c@{}}{Gemma-4-E4B} \\
\cmidrule(lr){2-4}\cmidrule(lr){5-7}\cmidrule(l){8-10}
Level & $P$ & $R$ & $F_1$ & $P$ & $R$ & $F_1$ & $P$ & $R$ & $F_1$ \\
\midrule
\multicolumn{10}{@{}l}{\textit{Image-only}} \\
Easy   & 52.0 & 53.0 & 52.5 & 57.7 & 28.7 & 38.3 & 53.9 & 39.0 & 45.3 \\
Medium & 48.7 & 66.2 & 56.1 & 50.0 & 69.1 & 58.0 & 50.9 & 69.9 & 59.0 \\
Hard   & 51.9 & 4.0 & 7.5 & 49.4 & 62.1 & 55.0 & 50.8 & 52.9 & 51.8 \\
Pooled & 50.1 & 40.4 & 44.7 & 50.8 & 54.4 & 52.6 & 51.5 & 54.6 & 53.0 \\
\midrule
\multicolumn{10}{@{}l}{\textit{\oursagentname{}}} \\
Easy   & 89.4 & 84.0 & 86.6 & 90.1 & 87.7 & 88.9 & 89.5 & 77.0 & 82.8 \\
Medium & 87.8 & 62.4 & 73.0 & 80.8 & 65.6 & 72.4 & 83.6 & 70.8 & 76.7 \\
Hard   & 65.0 & 59.2 & 62.0 & 66.9 & 63.2 & 65.0 & 71.6 & 81.3 & 76.2 \\
Pooled & 79.8 & 67.8 & 73.3 & 78.7 & 71.4 & 74.9 & 80.2 & 76.4 & 78.2 \\
\bottomrule
\end{tabular*}
\end{apptable}

\subsection{\topopid{} per-category accuracy}
\label{sec:app-topopid-categories}

Tables~\ref{tab:app-topopid-cat-qwen4b}--\ref{tab:app-topopid-cat-gemma} break exact-match
accuracy by category. Qwen3-VL-4B matches or slightly exceeds the 8B harness
on the pooled set and on easy ($85.7$ vs.\ $83.1$ in
Table~\ref{tab:main-topopid-acc}); Qwen3-VL-8B is strongest on medium; Gemma-4-E4B
leads on hard. Family-level gaps can be large even when pooled scores look
similar: under the hard hop-distance bucket family, Qwen3-VL-8B harness
accuracy is only 15.3\% while Qwen3-VL-4B and Gemma-4-E4B reach 71.4\%
(Tables~\ref{tab:app-topopid-cat-qwen} and~\ref{tab:app-topopid-cat-qwen4b}).

\paragraph{Categories where image-only scores higher.}
Three of the 42 category cells favor the image-only baseline, all in the hard
tier: isolation for Qwen3-VL-4B (50.0 vs.\ 48.0), constrained routing for
Qwen3-VL-8B (50.0 vs.\ 48.0), and hop-distance buckets for Qwen3-VL-8B
(24.5 vs.\ 15.3). The first two are boolean families where the image-only value
is the chance floor itself (boolean majority 50.3\%, random 50.0\%;
Appendix~\ref{sec:app-topopid-baselines}), reached by answering hard booleans
with one label---the behavior behind the low hard-tier fabricated-true rate
reported for Qwen3-VL-4B in Appendix~\ref{sec:app-hallu}. The harness instead
commits to a tool-derived verdict, and these two families are the ones most
exposed to wiring error: each turns on whether one specific path or cut exists,
which exact edge $F_1$ 0.742 does not secure even while
connectivity-consistency stays at 0.958. The 2\,pp shortfall against the
coin-flip floor matches the faithful-but-wrong case in
Section~\ref{sec:limitations}. The bucket cell is the checkpoint-specific
tool-use gap noted above (15.3\% for Qwen3-VL-8B against 71.4\% for the other
two checkpoints); both conditions score far below the string floor.

\begin{apptable}
\centering
\caption{\topopid{} exact-match accuracy by category (\%) for Qwen3-VL-4B-Instruct.}
\label{tab:app-topopid-cat-qwen4b}
\small
\begin{tabular}{llrr}
\toprule
Level & Category & Image-only & \oursagent{} \\
\midrule
\multicolumn{4}{l}{\textit{Easy}} \\
& simple\_counting & 4.2 & 83.8 \\
& spatial\_connections & 52.3 & 90.7 \\
& structural & 51.7 & 83.3 \\
\midrule
\multicolumn{4}{l}{\textit{Medium}} \\
& reachability & 50.5 & 91.1 \\
& tag\_reasoning & 37.8 & 81.1 \\
& topology & 49.0 & 66.3 \\
& type\_search & 42.5 & 77.0 \\
\midrule
\multicolumn{4}{l}{\textit{Hard}} \\
& comparative\_counting & 50.0 & 67.6 \\
& composition & 26.6 & 62.1 \\
& constrained\_routing & 50.0 & 68.6 \\
& counting\_bucketed & 23.5 & 71.4 \\
& isolation & 50.0 & 48.0 \\
& negation & 50.5 & 70.3 \\
& quantifier & 50.6 & 62.9 \\
& spatial\_topology & 51.5 & 57.4 \\
\bottomrule
\end{tabular}
\end{apptable}

\begin{apptable}
\centering
\caption{\topopid{} exact-match accuracy by category (\%) for Qwen3-VL-8B-Instruct.}
\label{tab:app-topopid-cat-qwen}
\small
\begin{tabular}{llrr}
\toprule
Level & Category & Image-only & \oursagent{} \\
\midrule
\multicolumn{4}{l}{\textit{Easy}} \\
& simple\_counting & 5.8 & 75.0 \\
& spatial\_connections & 54.7 & 91.3 \\
& structural & 53.0 & 85.7 \\
\midrule
\multicolumn{4}{l}{\textit{Medium}} \\
& reachability & 44.1 & 81.7 \\
& tag\_reasoning & 37.3 & 75.6 \\
& topology & 49.5 & 82.2 \\
& type\_search & 49.1 & 83.3 \\
\midrule
\multicolumn{4}{l}{\textit{Hard}} \\
& comparative\_counting & 46.1 & 95.1 \\
& composition & 32.0 & 67.5 \\
& constrained\_routing & 50.0 & 48.0 \\
& counting\_bucketed & 24.5 & 15.3 \\
& isolation & 50.0 & 65.7 \\
& negation & 50.5 & 83.2 \\
& quantifier & 49.4 & 62.9 \\
& spatial\_topology & 50.5 & 54.5 \\
\bottomrule
\end{tabular}
\end{apptable}

\begin{apptable}
\centering
\caption{\topopid{} exact-match accuracy by category (\%) for Gemma-4-E4B-it.}
\label{tab:app-topopid-cat-gemma}
\small
\begin{tabular}{llrr}
\toprule
Level & Category & Image-only & \oursagent{} \\
\midrule
\multicolumn{4}{l}{\textit{Easy}} \\
& simple\_counting & 5.9 & 80.5 \\
& spatial\_connections & 53.8 & 88.0 \\
& structural & 52.4 & 80.0 \\
\midrule
\multicolumn{4}{l}{\textit{Medium}} \\
& reachability & 49.5 & 95.0 \\
& tag\_reasoning & 22.9 & 62.2 \\
& topology & 24.8 & 70.8 \\
& type\_search & 47.1 & 72.4 \\
\midrule
\multicolumn{4}{l}{\textit{Hard}} \\
& comparative\_counting & 49.0 & 85.3 \\
& composition & 9.9 & 34.5 \\
& constrained\_routing & 52.5 & 67.2 \\
& counting\_bucketed & 0.0 & 71.4 \\
& isolation & 51.0 & 69.6 \\
& negation & 48.5 & 61.4 \\
& quantifier & 53.9 & 96.6 \\
& spatial\_topology & 50.5 & 74.3 \\
\bottomrule
\end{tabular}
\end{apptable}

\subsection{Qualitative overlays}
\label{sec:app-figures}

Test sheets 2 and 5 ($\sim$7168$\times$4561 native). Overlays use
class-colored symbol boxes; process edges are schematic center-to-center
process-connection links, not pipe routes.
Figures~\ref{fig:app-raw}--\ref{fig:app-dashed} follow the pipeline order:
raw raster, perception masks and boxes, vectorized geometry, the recovered
process graph, the dual-layer graph with signal edges, and a dashed-ink zoom.

\newcommand{\apppair}[2]{%
  \includegraphics[width=0.48\linewidth,height=0.22\textheight,keepaspectratio]{#1}\hfill
  \includegraphics[width=0.48\linewidth,height=0.22\textheight,keepaspectratio]{#2}%
}

\begin{appfig}
\centering
\apppair{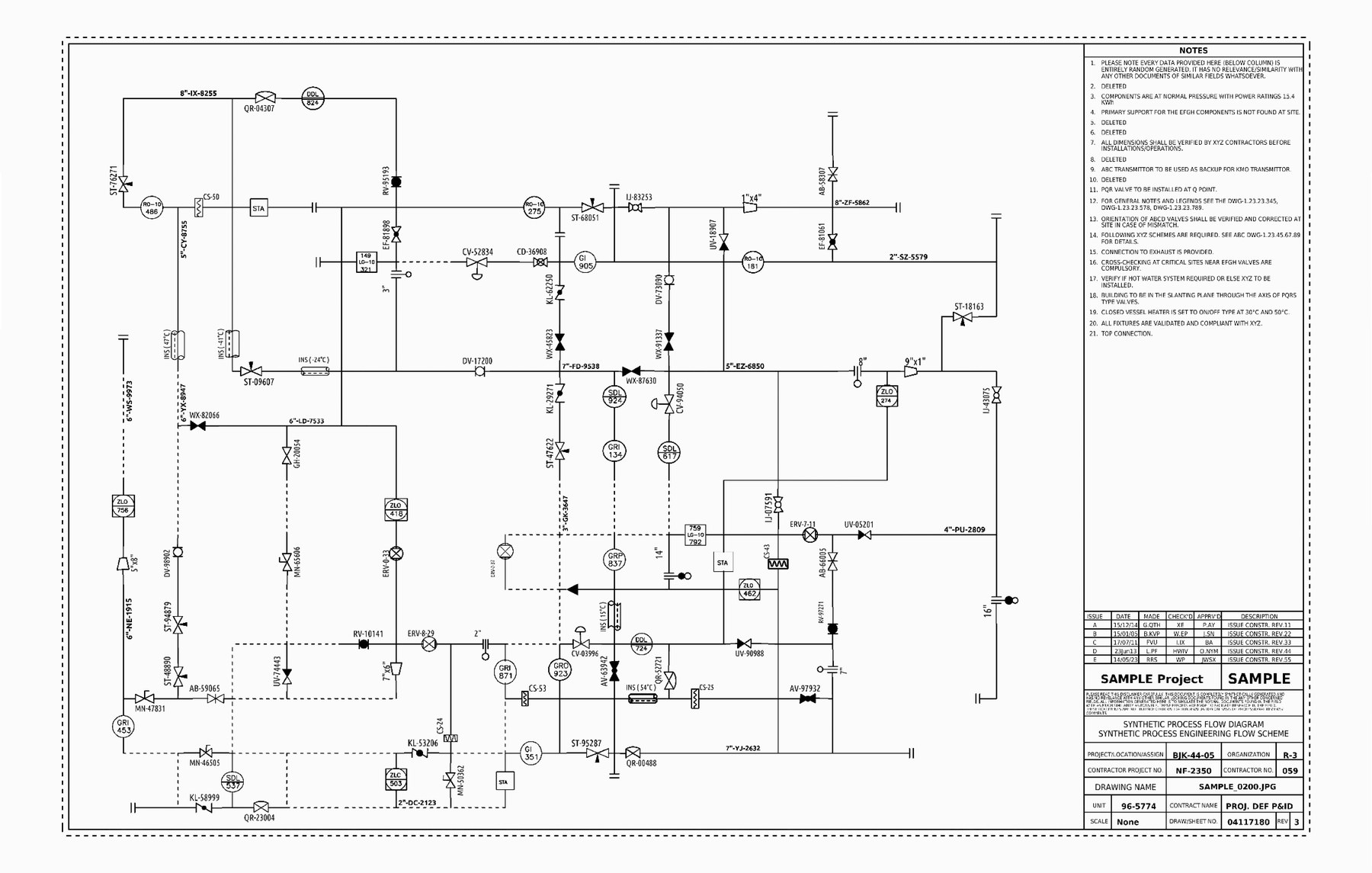}{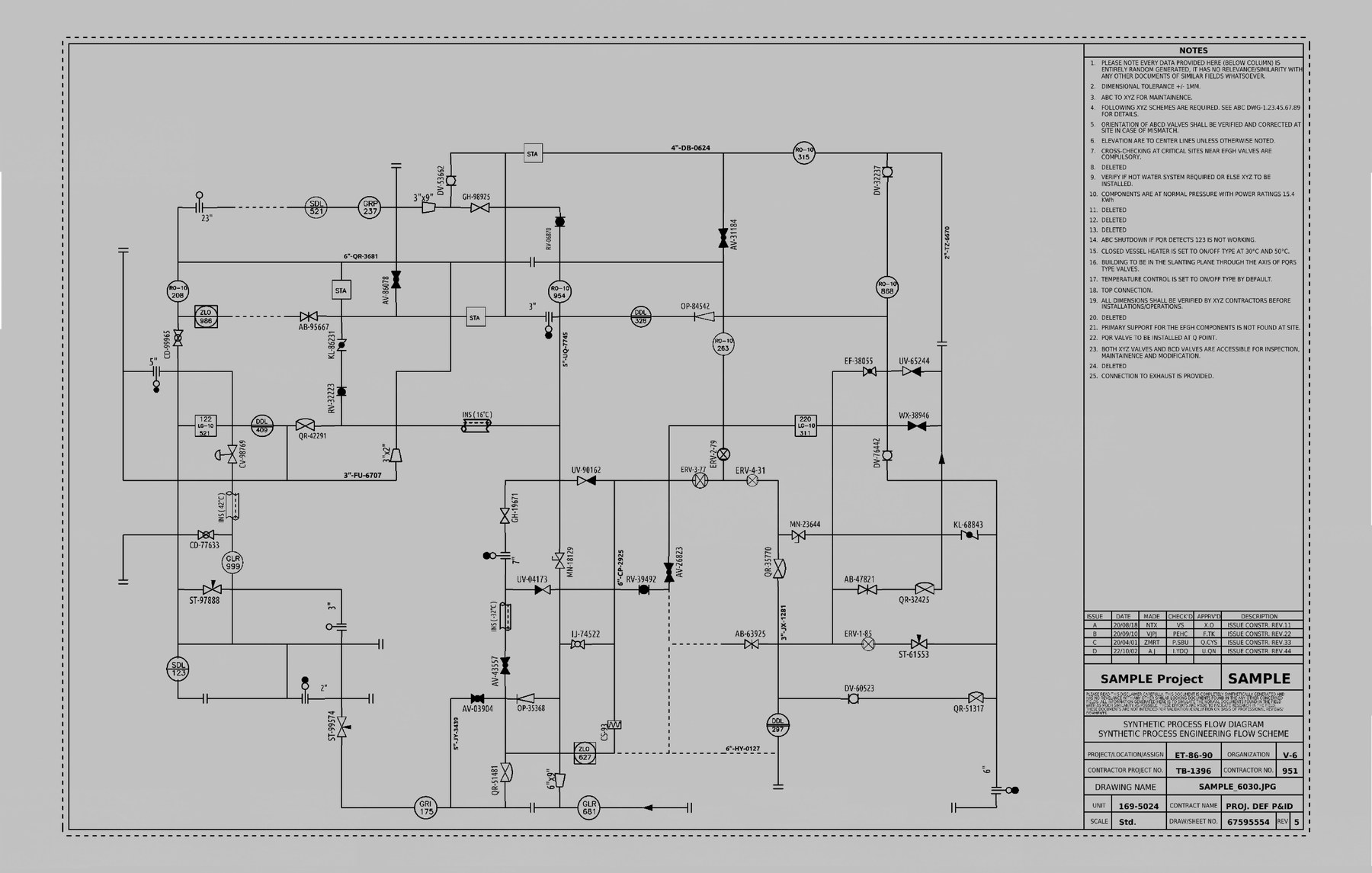}
\caption{Raw Digitize-PID sheets (downscaled for display).}
\label{fig:app-raw}
\end{appfig}

\begin{appfig}
\centering
\apppair{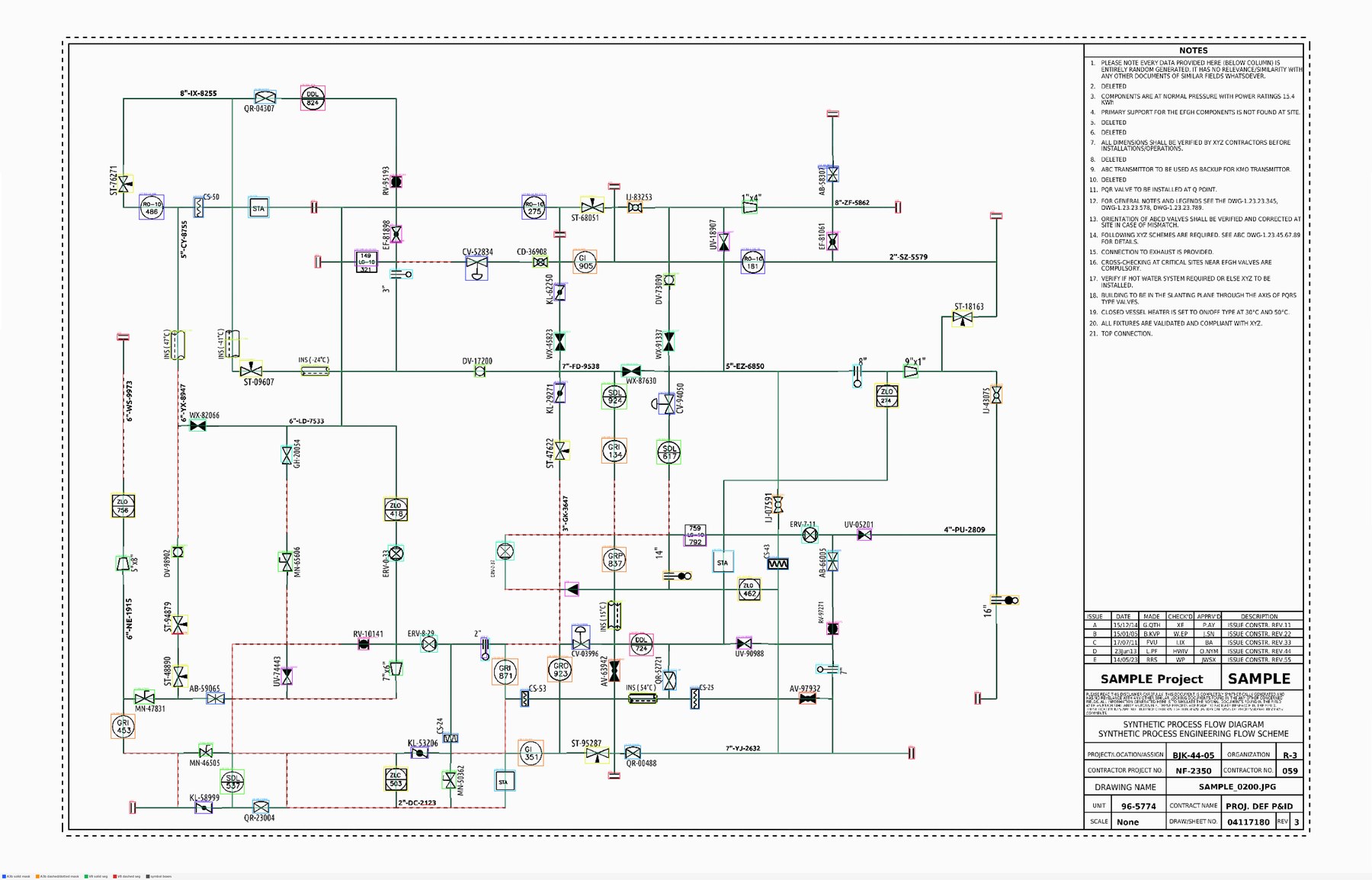}{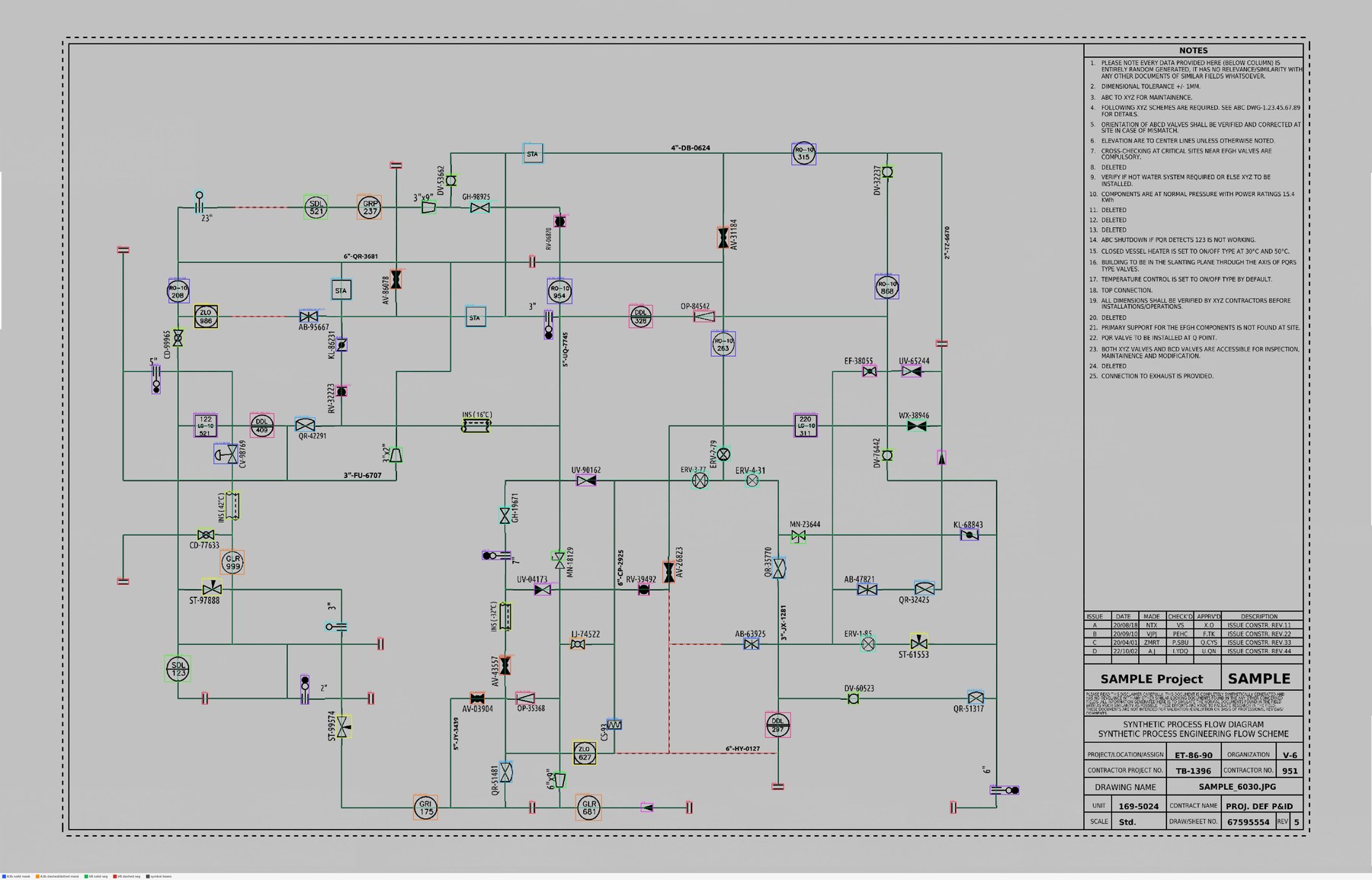}
\caption{Perception stack: solid/dashed masks, vectorized segments, symbol boxes.}
\label{fig:app-perception}
\end{appfig}

\begin{appfig}
\centering
\apppair{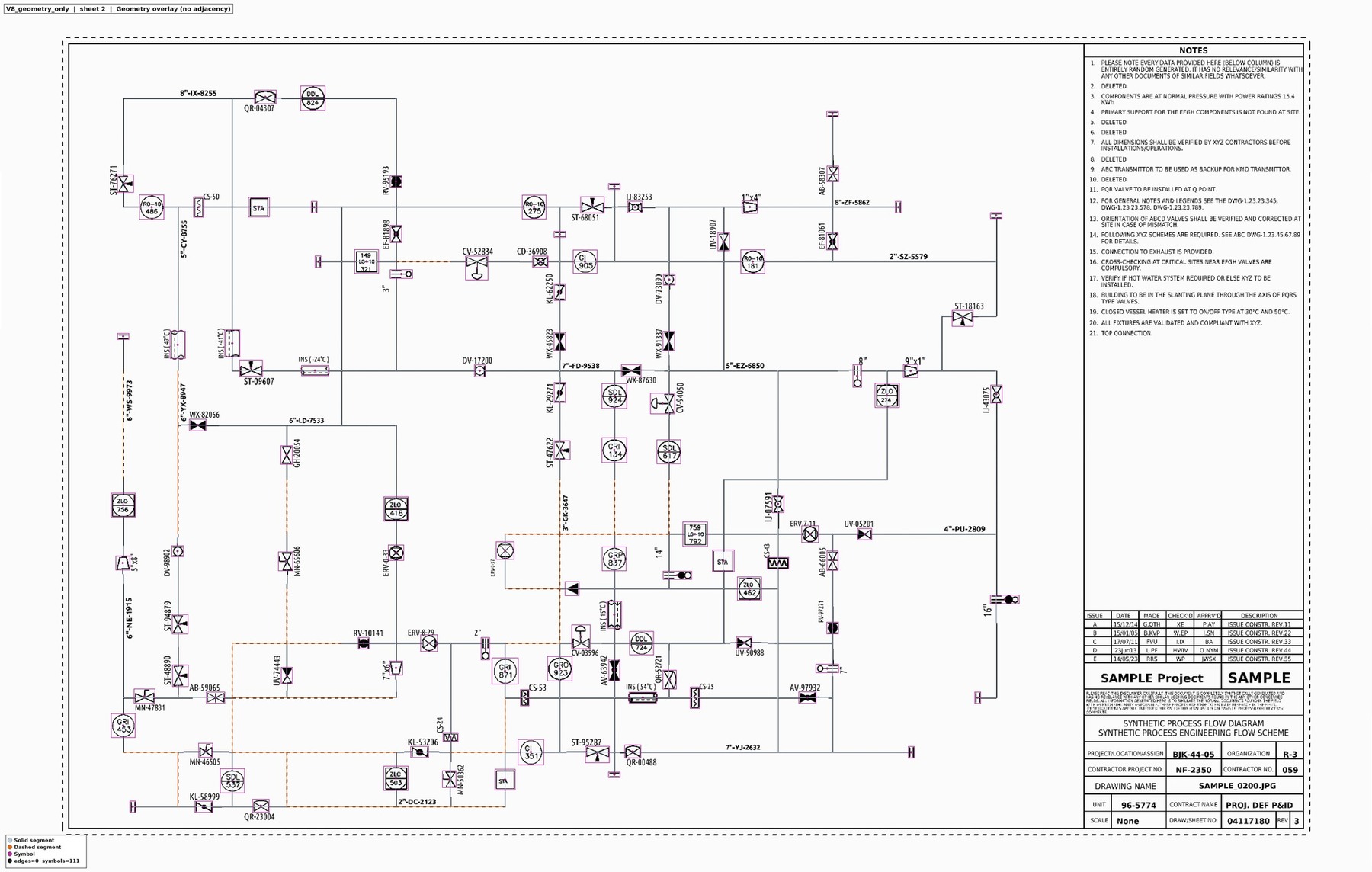}{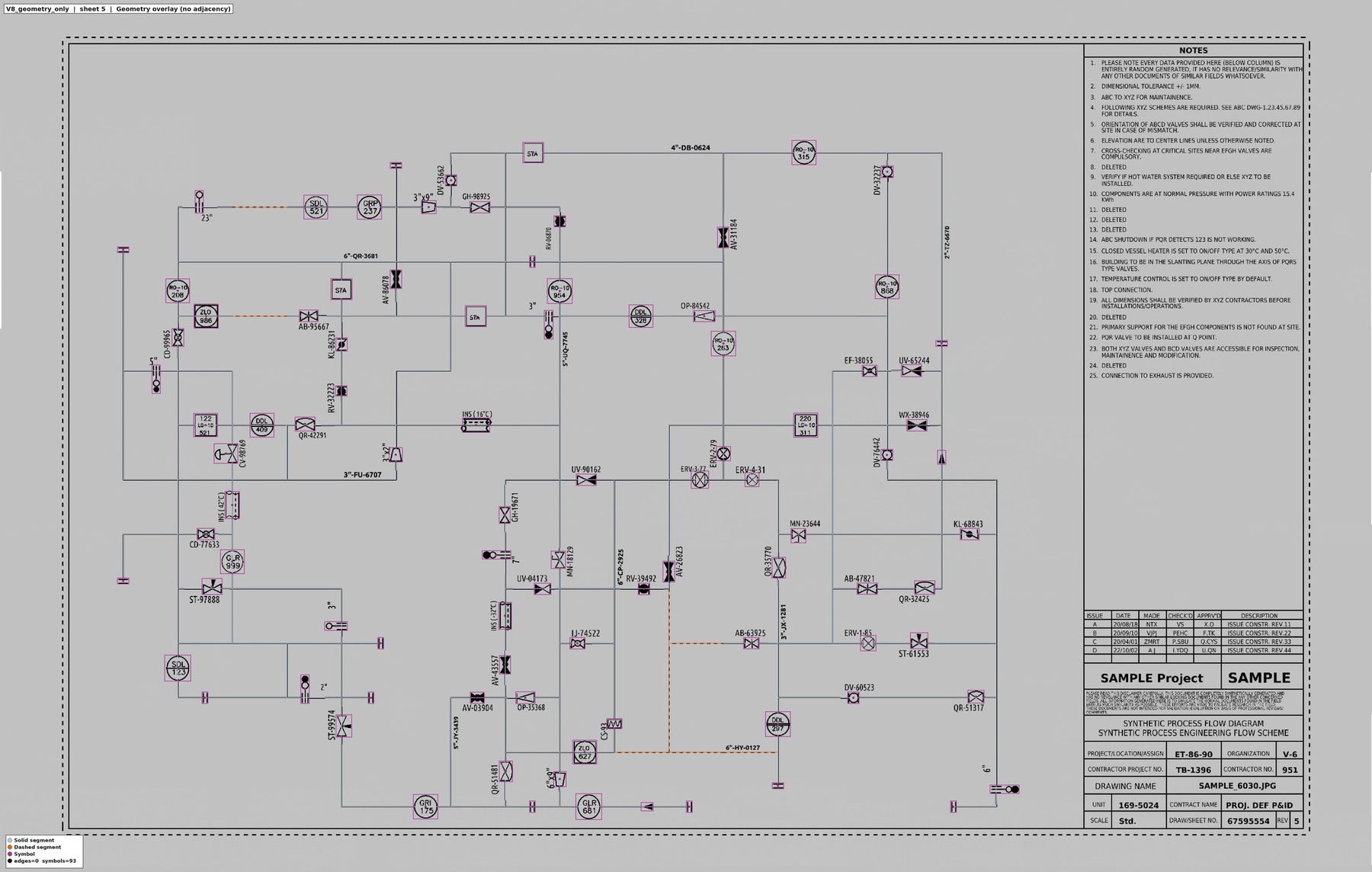}
\caption{Vectorized line geometry (solid and dashed polylines).}
\label{fig:app-geometry}
\end{appfig}

\begin{appfig}
\centering
\apppair{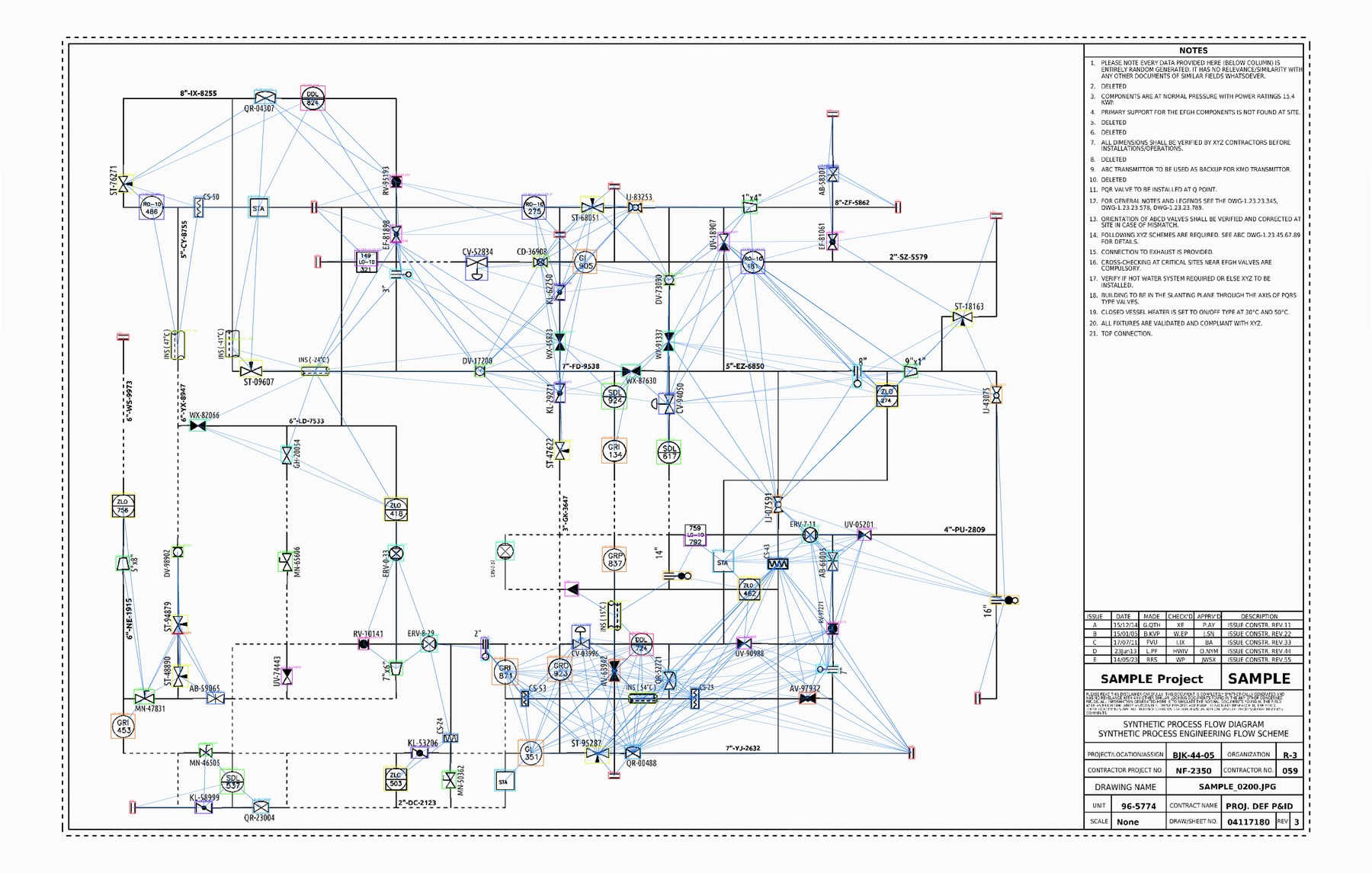}{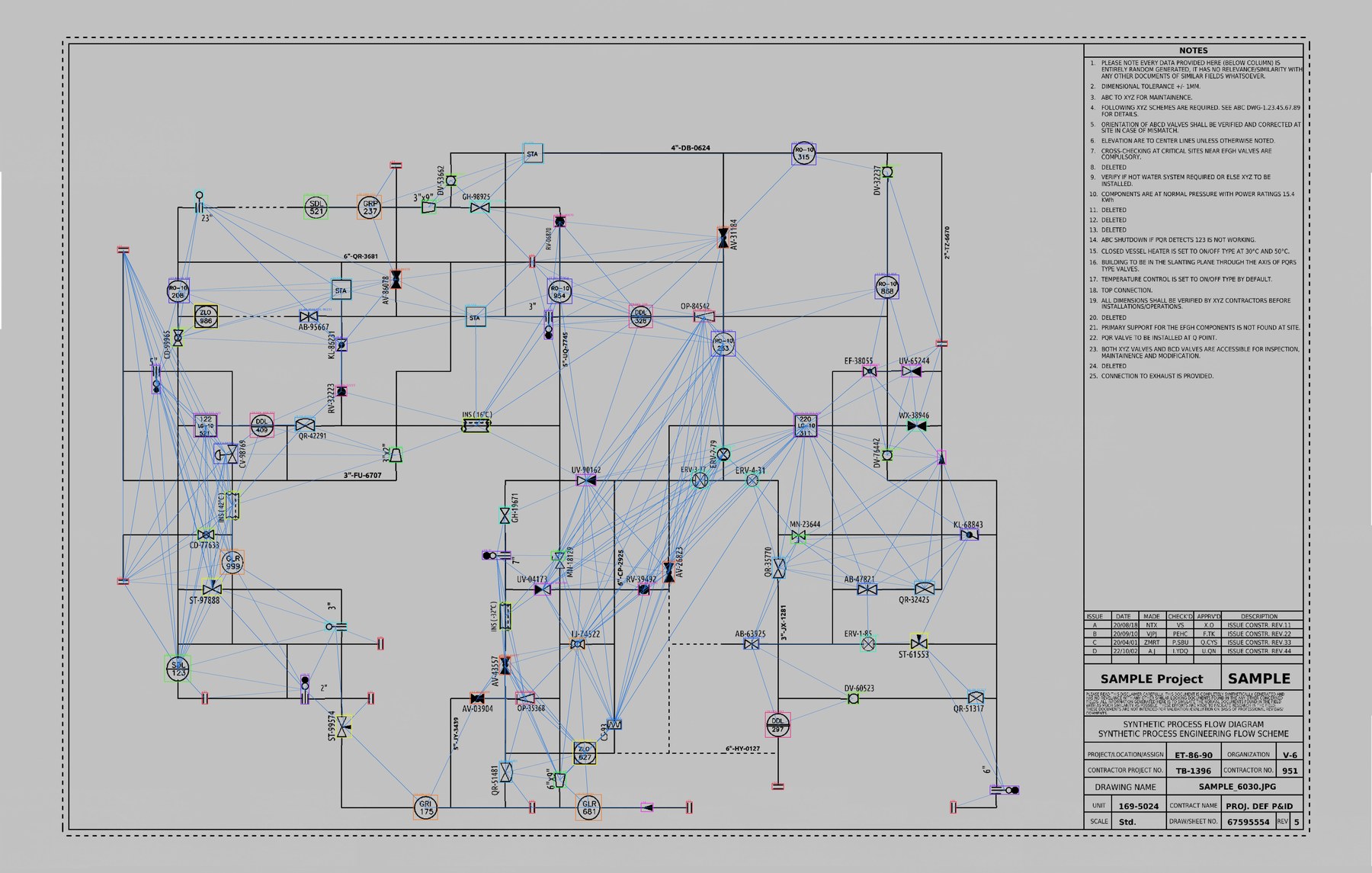}
\caption{Recovered process graph: process connections on $G_{\mathrm{auto}}$.}
\label{fig:app-process-graph}
\end{appfig}

\begin{appfig}
\centering
\apppair{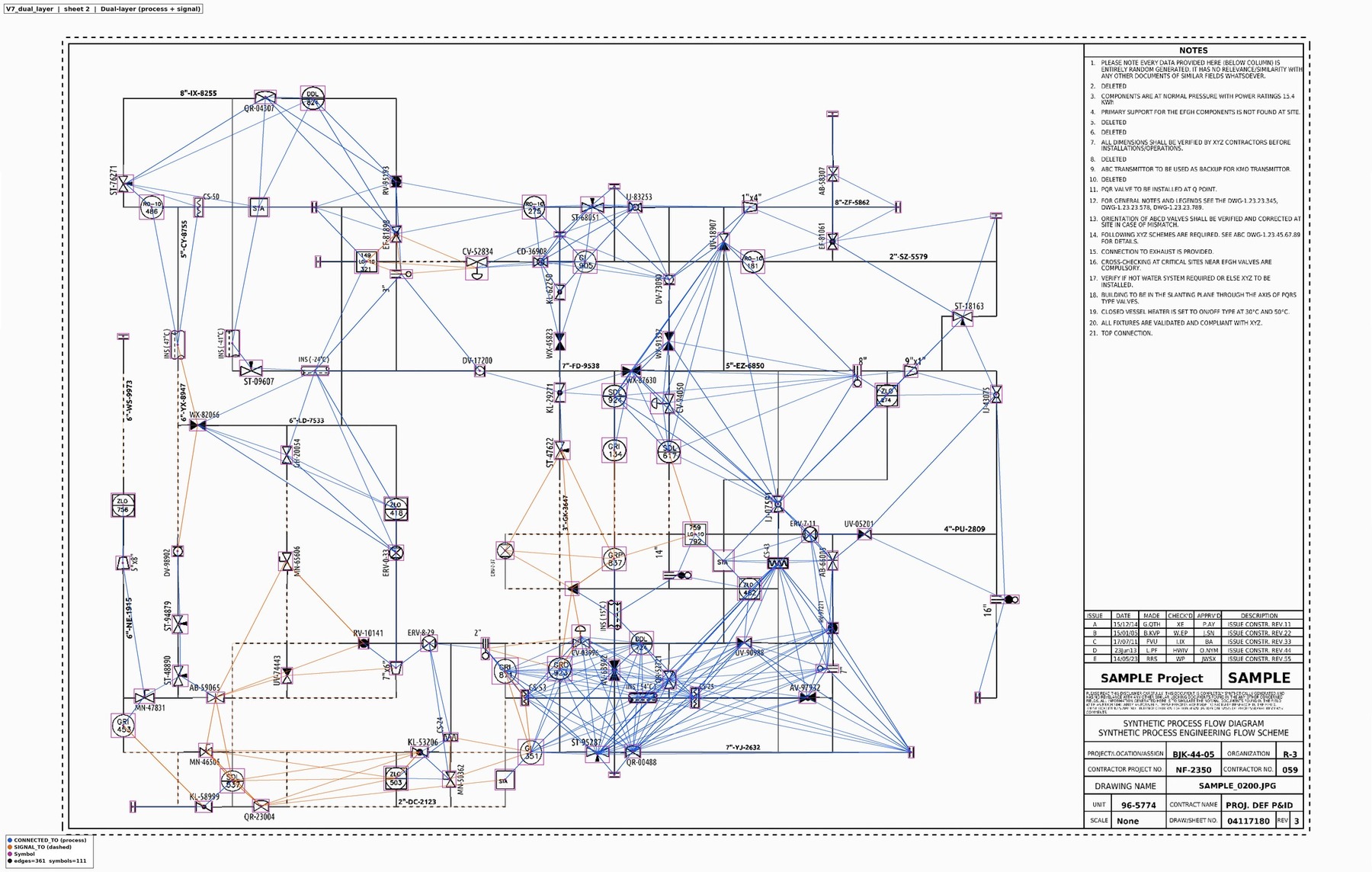}{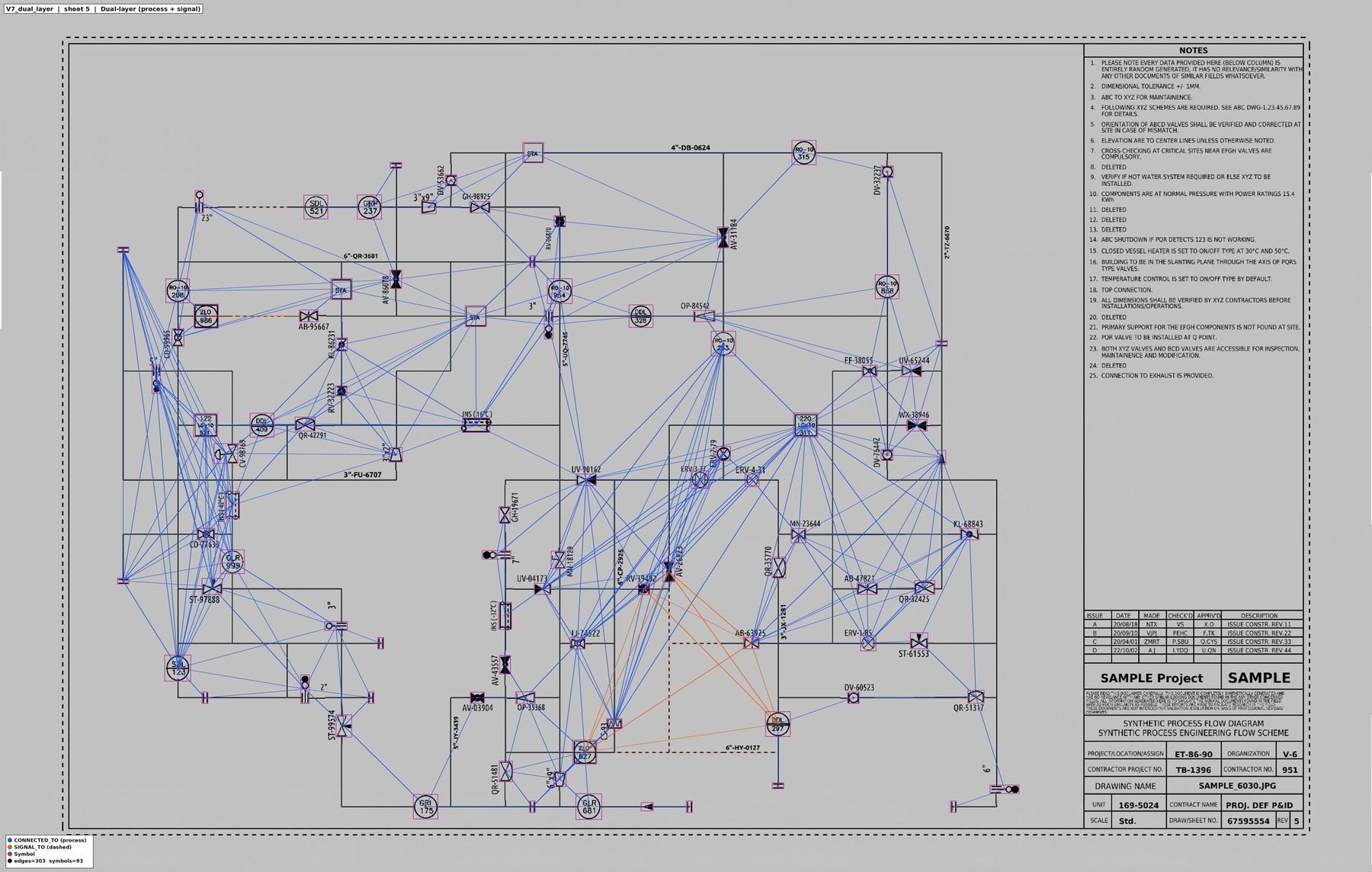}
\caption{Dual-layer graph: process connections (solid) and signal edges (dashed).}
\label{fig:app-signal-layer}
\end{appfig}

\begin{appfig}
\centering
\apppair{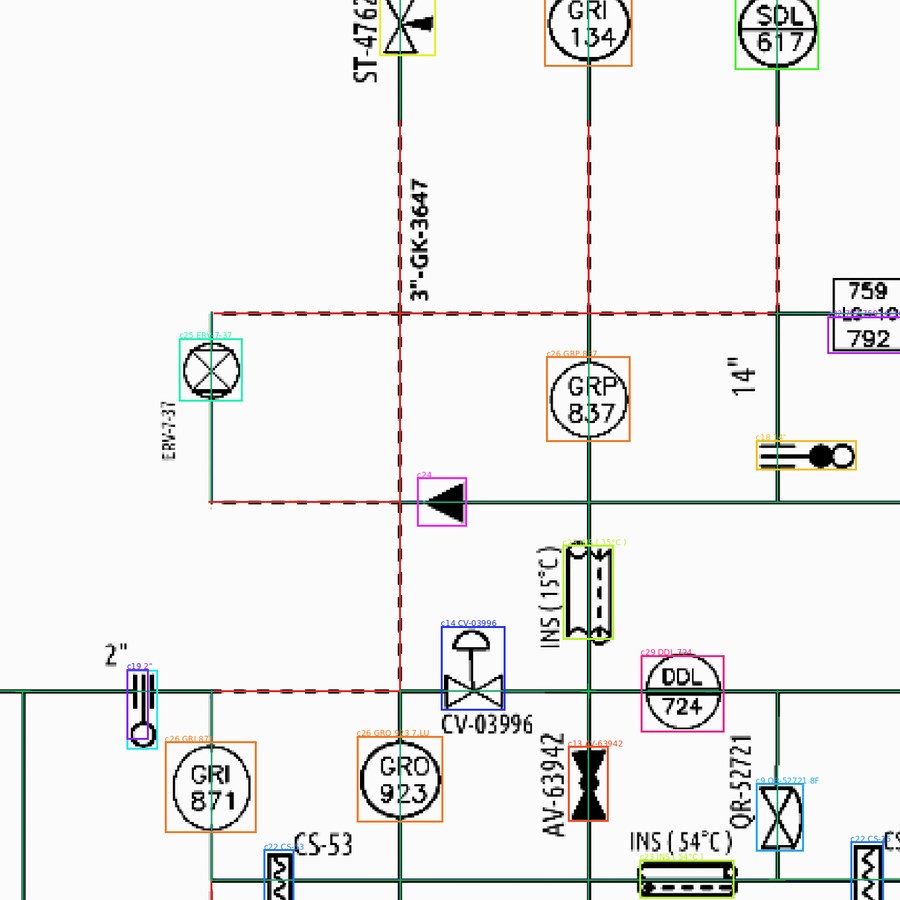}{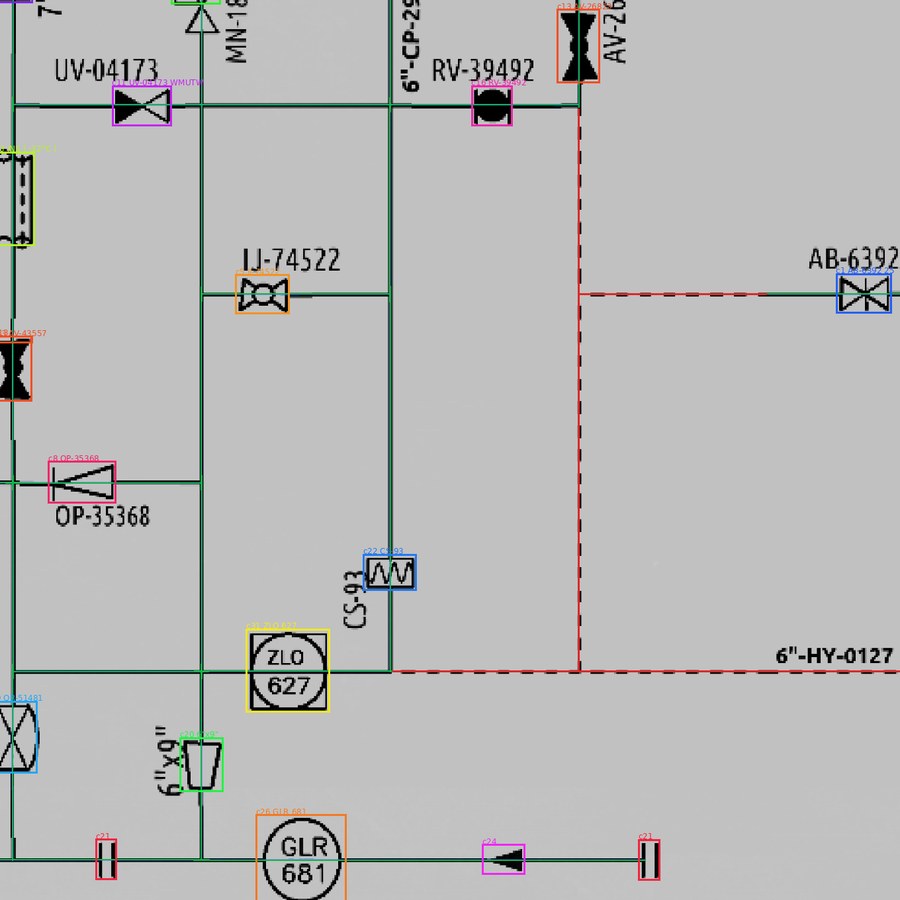}
\caption{Dashed-ink zoom: masks and vectorized dashed segments.}
\label{fig:app-dashed}
\end{appfig}

\FloatBarrier
\clearpage

\end{document}